\documentclass{cornell}

\usepackage{psfig}
\usepackage{epsfig}
\usepackage{tabularx}
\usepackage{graphicx}
\usepackage{setspace}
\usepackage{subfigure}
\usepackage{amsmath,amssymb,multicol}
\usepackage{algorithmic}
\usepackage{algorithm}
\usepackage{fancyhdr}
\usepackage{rotating}
\usepackage{bm}

\newtheorem{thm}{Theorem}[section]

\newtheorem{lem}{Definition}
\newtheorem{lem1}{Lemma}

\usepackage{tabularx}
\usepackage{hhline}
\usepackage{amsmath}
\usepackage{rotating}
\usepackage{multirow}
\newenvironment{proof}{{\it Proof. }}{$\triangleleft$}

\begin{document}

\title{TRUST-TECH based Methods for\\ Optimization and Learning}
\author{Chandankumar Reddy Karrem}
\maketitle
\begin{copyrightpage}
\vspace{3in}
\copyright ~2007 Chandankumar Reddy Karrem\\
ALL RIGHTS RESERVED
\end{copyrightpage}
\begin{abstract}

\noindent

Many problems that arise in machine learning domain deal with nonlinearity
and quite often demand users to obtain global optimal solutions rather
than local optimal ones. Optimization problems are inherent in machine learning algorithms and hence many methods in machine learning were inherited from the optimization literature. Popularly known as the {\it initialization problem}, the
ideal set of parameters required will significantly depend on the
given initialization values. The recently developed TRUST-TECH (TRansformation Under
STability-reTaining Equilibria CHaracterization) methodology systematically explores the subspace of the parameters
to obtain a complete set of local optimal solutions. In this thesis work, we propose TRUST-TECH based methods for solving several optimization and machine learning problems. TRUST-TECH explores the
dynamic and geometric characteristics of stability boundaries of a
nonlinear dynamical system corresponding to the nonlinear function
of interest. Basically, our method coalesces the advantages of the
traditional local optimizers with that of the dynamic and geometric
characteristics of the stability regions of the corresponding
nonlinear dynamical system. Two stages namely, the local stage and the neighborhood-search stage, are
repeated alternatively in the solution space to achieve
improvements in the quality of the solutions. The local stage
obtains the local maximum of the nonlinear function and the
neighborhood-search stage helps to escape out of the local maximum by
moving towards the neighboring stability regions. Our methods were tested on both synthetic and real
datasets and the advantages of using this novel
framework are clearly manifested. This framework not only reduces
the sensitivity to initialization, but also allows the flexibility
for the practitioners to use various global and local methods that
work well for a particular problem of interest. Other hierarchical stochastic algorithms like evolutionary algorithms and smoothing algorithms are also studied and frameworks for combining these methods with TRUST-TECH have been proposed and evaluated on several test systems.
%

\end{abstract}
\pagenumbering{roman} \setcounter{page}{4}
\begin{biosketch}
Chandan Reddy was born in Jammalamadugu, Andhra Pradesh, India on May 11, 1980. After obtaining Bachelor's degree from Pondicherry Engineering College in 2001, he moved to the United States to pursue his higher education. He obtained his master's degree from the Department of Computer Science and Engineering at Michigan State University in August 2003. Later, he joined the Department of Electrical and Computer Engineering at Cornell University to pursue his doctoral studies. His primary research interests are in the areas of Machine Learning, Data Mining, Optimization, Computational Statistics, Bioinformatics and Biomedical Imaging. 
\end{biosketch}
\begin{dedication}
{Dedicated to my parents}
\end{dedication}
\begin{acknowledgements}
\noindent
I would like to thank my thesis advisor, Prof. Hsiao-Dong Chiang, without whose guidance and involvement this work would not have been possible. His enthusiasm and inspiration were extremely helpful in successfully pursuing this research work. I would also like to thank my committee member Prof. John Hopcroft for his encouragement and useful discussions. I am grateful to Dr. Peter Doerschuk for serving on my thesis committee. Special thanks to Dr. Bala Rajaratnam for his help with a few statistical works related to my thesis.

I am deeply indebted to my father, who was my main inspiration for pursuing graduate studies and choosing to have an academic career. Needless to say I had plenty of great discussions and fun with many people in the department including Jeng-Heui, Warut, Choi, Seunghee, Siva, Bhavin, Wesley, Jay and Huachen. I was also fortunate to have a good company at Cornell who have kept my spirits up and my life interesting: Manpreet, Pallavi, Sudha, Raju, Kumar, Mahesh, Pankaj, Deepak, Pradeep, Ajay and Shobhit. A special thanks to the graduate secretaries for taking care of countless numbers of things without any hitch.

Last and the most important, my deepest gratitude goes to my dear mother for her love, and for making me the person I am. 
\end{acknowledgements}
\tableofcontents
\listoffigures
\listoftables
\newpage
\pagenumbering{arabic} \setcounter{page}{1}
\chapter{Introduction}
\label{ch:introduction} 

The problem of finding a global optimal solution arise in many disciplines ranging from science to engineering. In real world applications, multi-dimensional objective functions usually contain a large number of local optimal solutions. Obtaining a global optimal solution is of primary importance in these applications and is a very challenging problem. Some examples of these applications are : molecular confirmation prediction \cite{Byrd96}, VLSI design in microelectronics \cite{Litovski97}, resource allocation problems \cite{Dong94}, design of wireless networks \cite{He02}, financial decision making \cite{Kim04}, structural engineering \cite{Foley03} and parameter estimation problems \cite{Esposito00}. In this thesis, the primary focus is on the parameter estimation problems that arise in the field of machine learning.

\section{Machine Learning}

Machine learning algorithms can be broadly classified into two categories \cite{Duda01}: (i) Supervised learning and (ii) Unsupervised learning. The primary goal in {\it supervised learning} is to learn a mapping from $x$ to $y$ given a training dataset which consists of pairs ($x_i,y_i$), where $x_i \in \mathcal{X}$ are the data points and $y_i \in \mathcal{Y}$ are the labels (or targets). A standard assumption is that the pairs ($x_i,y_i$) are sampled i.i.d. from some distribution. If $y$ takes values in a finite set (discrete values) then it is a classification problem and if it takes values in a continuous space, then it is a regression problem. Support vector machines \cite{Burges98}, artificial neural networks \cite{Haykin99} and boosting \cite{Friedman00} are the most popular algorithms for supervised learning. All these algorithms will construct a classification (or regression) model based on certain training data available. Usually, the effectiveness of any algorithm is evaluated using testing data which is separate from the training data. In this thesis, we will primarily focus on {\it artificial neural networks} and estimating the parameters of its model. Constructing a model using artificial neural network involves estimating the parameters of the model that can effectively exploit the potential of the model. These parameters are usually obtained by finding the global minimum on the error surface. More details on training neural networks will be presented in Chapter~\ref{ch:training}.

{\it Unsupervised learning}, on the other hand, will train models using only the datapoints without the target values. In simple terms, only $x$ values are available without $y$ values. Problems like outlier detection, density estimation, data clustering and noise removal fall under this category. Data clustering is one of the widely studied unsupervised learning topics \cite{Jain88}. Density estimation is a more generalized notion of data clustering. It involves in estimating and understanding and the underlying distribution of the data. Applications of density estimation include trend analysis and data compression. Typically, one would like to estimate the parameters of a model that consists of multiple components of varying densities. More details on these mixture models and the use of expectation maximization (EM) algorithm for parameter estimation of these models will be presented in Chapter~\ref{ch:trust-tech-em}.

\begin{figure}
   \centering
   \subfigure[supervised Learning]{\includegraphics[width = 2.45 in]{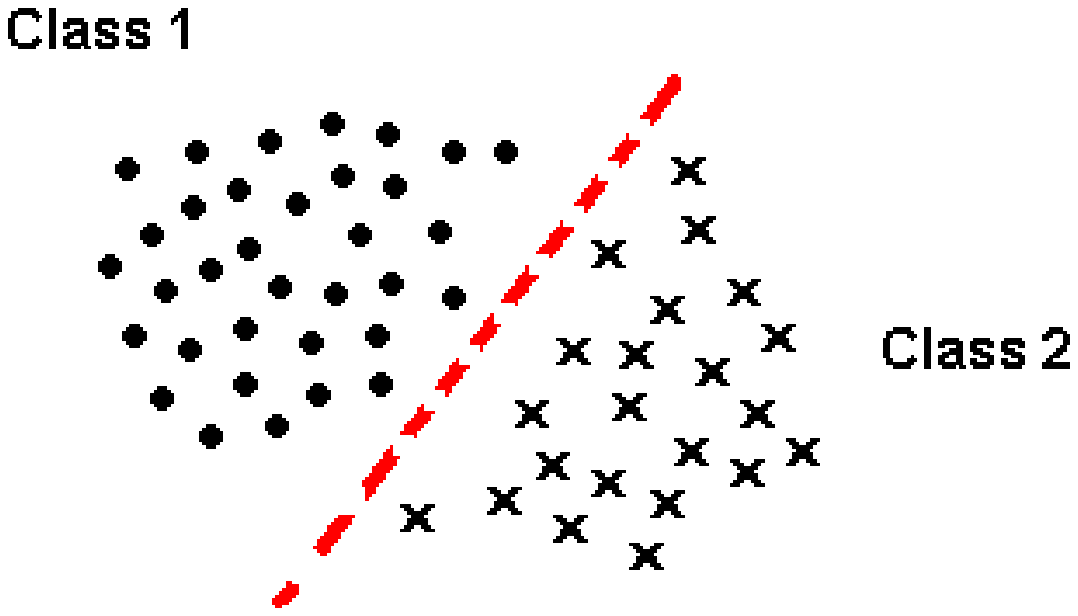}}\qquad
   \subfigure[Unsupervised Learning]{\includegraphics[width = 2.45 in]{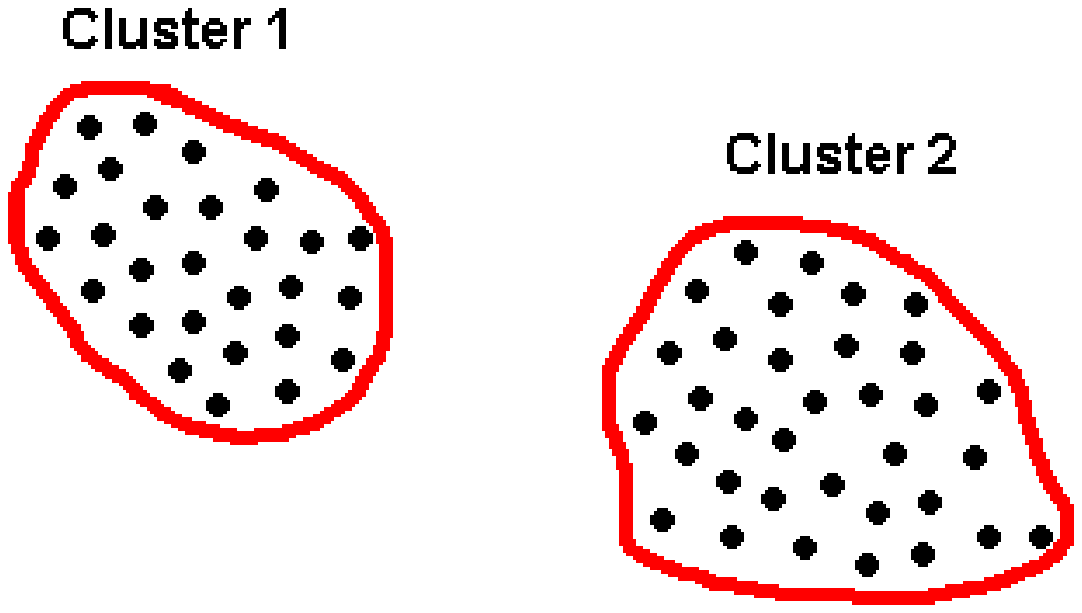}}\qquad
   \caption{\label{fig:algo} (a) Supervised learning with data points from two different classes separated by a hyperplane represented using a dashed line. (b) Unsupervised Learning or data clustering - two well separated clusters.
   }
   \label{fig:supunsup}
\end{figure}

Fig.~\ref{fig:supunsup} shows the supervised and unsupervised learning scenarios. In supervised learning, the main goal is to train a model such that a final target class can be estimated for a new (unseen) data point. In a simple binary classification problem, a hyperplane (indicated by a dashed line) separates both the classes. In clustering problems, the main goal is to form groupings of the data and obtain any interesting structure (or patterns) in the data (see Fig.~\ref{fig:supunsup}(b)). 

In both the models mentioned above (neural networks and mixture models), estimating the parameters correspond to obtaining a global optimal solution on a highly nonlinear surface. The surface can be generated based on a function that might represent the error in the training data or the likelihood of the data given the model. We will now introduce different categories of the nonlinear optimization methods studied in the literature.

\section{Optimization Methods}

 Optimization methods can be broadly classified into two categories: {\it global methods} and {\it local methods}. Global methods explore the entire search space and obtain promising regions that have a higher probability of finding a global optimal solution. They can be either deterministic or stochastic in nature depending on the usage of some random component in the algorithm \cite{Horst95}.  Deterministic global methods include branch and bound \cite{Horst96}, homotopy based \cite{Forster95}, interval analysis \cite{Hansen92} and trajectory-based \cite{Diener95}. These methods are also known as exact methods. Stochastic global methods include techniques such as evolutionary algorithms \cite{Back96}, simulated annealing \cite{Kirkpatrick83}, tabu search \cite{Glover93} and ant colony optimization \cite{Dorigo04} which can give asymptotic guarantees of finding the global optimal solution. Many heuristic search algorithms can be incorporated into these stochastic methods to improve the performance of the algorithm in terms of quality of the solution and the speed of convergence.

\begin{sidewaysfigure}[htp]
\centerline{
  \epsfig{figure=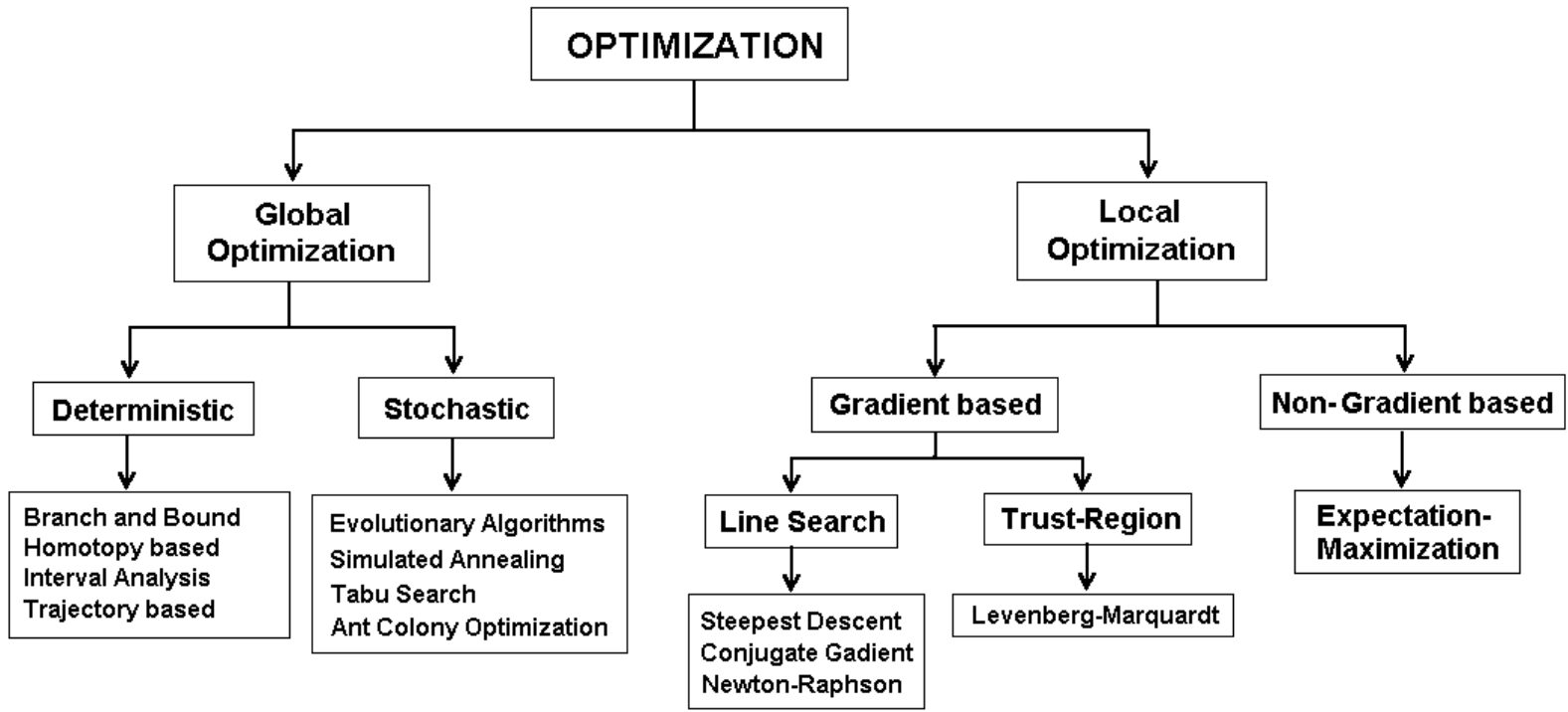, width=8.5in}
} \caption{Different optimization methods.}
\label{fig:opti}
\end{sidewaysfigure}

Usually, local methods are deterministic in nature and can be grouped into two categories: {\it gradient based} and {\it non-gradient based methods}.

\begin{enumerate}
\item{
{\it Gradient based methods} can be further classified into (i) Line search methods and (ii) Trust-region methods. Line search algorithms usually select some descent direction (based on the gradient information) and minimize the function value along the chosen direction. This process is repeated until a local minimum is reached. The most popular line search algorithms include steepest descent \cite{Press92}, conjugate gradient \cite{Avriel03} and Newton-Raphson method \cite{Nocedal99}. Trust region methods \cite{Celis85,Marquardt63}, on the other hand, make an assumption about the nonlinear surface locally and do not use any form of line search. Typically, they assume the surface to be a simple quadratic model such that the minimum can be located directly if the model assumption is good which usually happens when the initial guess is close to the local minimum. If the model assumption is not accurate, then gradient information is used to guide the initial guess and after a certain period the model assumption is made again.}

\item{In contrast to the above mentioned gradient based methods, {\it non-gradient based methods} do not use any gradient information. Usually, these methods rely on a particular form of the nonlinear function and ensure that a chosen iterative parameter updating scheme results in the decrease of the function value \cite{Cherkassky98}. For density estimation problems using maximum likelihood function, a popular class of iterative parameter estimation methods is the Expectation Maximization (EM) algorithm which converges to the maximum likelihood estimate of the mixture parameters locally \cite{Demspter77,Redner84}.}
    
\end{enumerate}

\section{Motivation for this Thesis}
Finding the global optimal solution of a function is a lot more tedious and challenging for many problems. The task of finding such a solution is quite complex and increases rapidly with the dimensionality of the problem. Typically, the problem of finding the global optimal solution is solved in a hierarchical manner. Identifying some promising regions in a search space is relatively easier using certain global methods (such as genetic algorithms and simulated annealing) available in the literature. However, the fine tuning capability of these global methods is very poor and sometimes yield a comparatively less accurate solution even though the neighborhood region appears to be promising. Hence, there is an absolute necessity for exploring this surrounding to obtain a better solution.

\begin{figure}
   \centering
   \subfigure[Promising regions in the search space]{\includegraphics[width = 4.05 in]{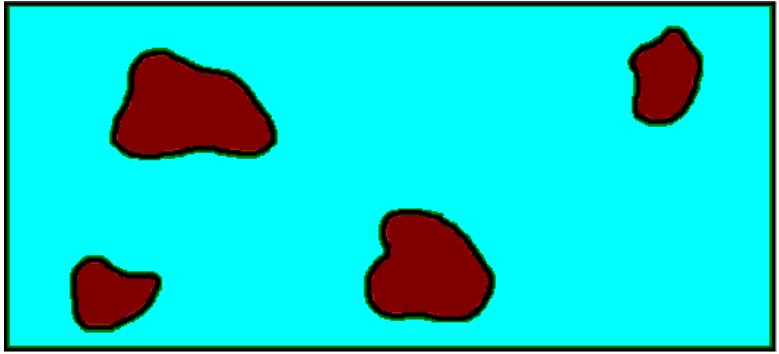}}\qquad
   \subfigure[Promising Subspaces with multiple local optimal solutions]{\includegraphics[width = 4.05 in]{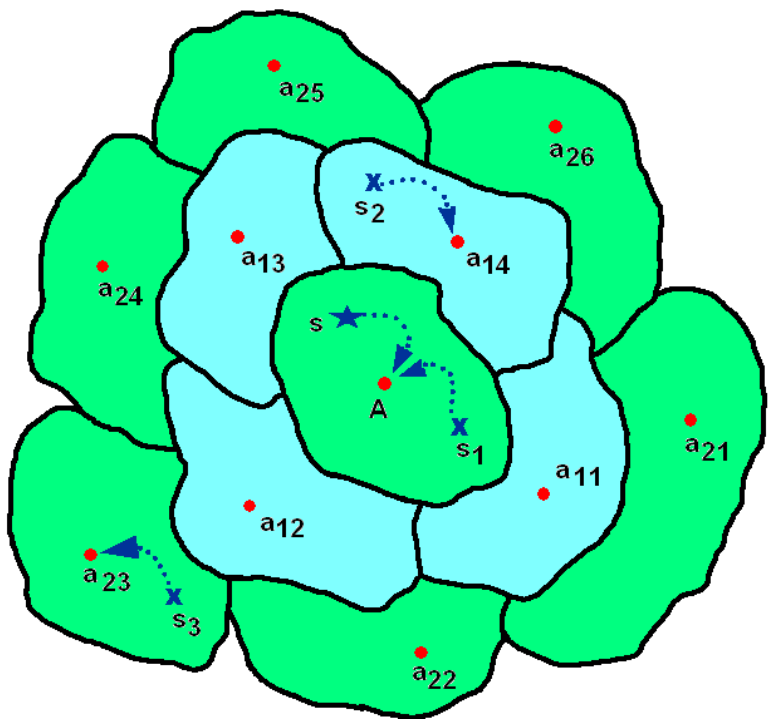}}\qquad
   \caption{\label{fig:algo}Various stages of TRUST-TECH (a) The dark regions indicate the promising subspaces. (b) The dots indicate the local optimal solutions and the dotted arrows indicate the convergence of local optimization method from a given initial point.
   }
   \label{fig:subspaces}
\end{figure}

Fig. \ref{fig:subspaces} clearly shows the difficulties in dealing with nonlinear surfaces. Global methods can be used to obtain promising subspaces in the parameter space. These are indicated by dark shaded regions in Fig. \ref{fig:subspaces}(a). However, these promising regions are not convex in nature. i.e. they will have multiple local optimal solutions. Fig. \ref{fig:subspaces}(b) gives the top view of the nonlinear surface in the promising region. The dots indicate the local optimal solutions. `S' is the initial point obtained from the global methods. Applying local method at `S' will converge to `A'. There are other stochastic methods that can search the neighborhood regions e.g. mutations in genetic algorithms, low temperature annealing in simulated annealing.

We will now discuss the problems with these optimization methods mentioned above and motivate the necessity of TRUST-TECH (TRansformation Under STability-reTaining Equilibria CHaracterization) based methods. Finding a local optimum is relatively easier and straightforward using a local method. The stochastic methods randomly perturb a given point without much topological understanding of the nonlinear surface. By transforming the nonlinear function into its corresponding dynamical system, TRUST-TECH  can obtain neighboring local optimal solutions deterministically \cite{Chiang03,Chiang07}. TRUST-TECH not only guarantees that a new local maximum obtained is different from the original solution but also confirms that any solution in a particular direction will not be missed. As shown in Fig. \ref{fig:subspaces}(b), the given local optimal solution `A' is randomly perturbed to obtain new initial points ($s_1,s_2,s_3$). Applying local method using these initial points again, one can obtain the local optimal solutions $A,a_{14}$ and $a_{23}$ respectively. It can be observed that the solutions might appear again or might sometimes even miss some of the neighborhood solutions.

In this thesis, we develop TRUST-TECH based methods for systematically finding neighborhood solutions for problems that arise in the fields of optimization and machine learning. This method is more reliable and deterministic when compared to other stochastic approaches which merely use random moves to obtain new solutions. To begin with, the original nonlinear function is transformed into its corresponding dynamical system. There will be a one-to-one correspondence of all the critical points under this transformation. Also, this will allow us to define the concepts like stability boundaries which can be used to obtain the neighborhood solutions effectively.

\section{Contributions of this Thesis}
The main contributions of this thesis work are :
\begin{itemize}
\item{Transform the problem of finding saddle points on potential energy surfaces to the problem of finding decomposition points on its corresponding nonlinear dynamical system. Apply a novel stability boundary based algorithm for tracing the stability boundary and obtaining saddle points on potential energy surfaces that arise in the fields of computational chemistry and computational biology. }

\item{Develop TRUST-TECH based Expectation Maximization (EM) algorithm for the learning finite mixture models.}

\item{Apply the above mentioned TRUST-TECH based EM algorithm for the challenging motif finding problem in bioinformatics.}

\item{Develop a component-wise kernel smoothing algorithm for learning Gaussian mixture models more efficiently. Demonstrate empirically that the number of unique local maxima on the likelihood surface can be reduced.}

\item{Implement TRUST-TECH based training algorithm for artificial neural networks that can explore the topology of the error surface and obtain optimal set of parameters for the network model.}

\item{Develop evolutionary TRUST-TECH methods that combines the advantages of the popular stochastic global optimization methods (like Genetic algorithms) with deterministic TRUST-TECH based search strategies. Demonstrate that the promising subspaces in the search space can be achieved at a much faster rate using evolutionary TRUST-TECH.}

\end{itemize}
\begin{figure}[htp]
\centerline{
  \epsfig{figure=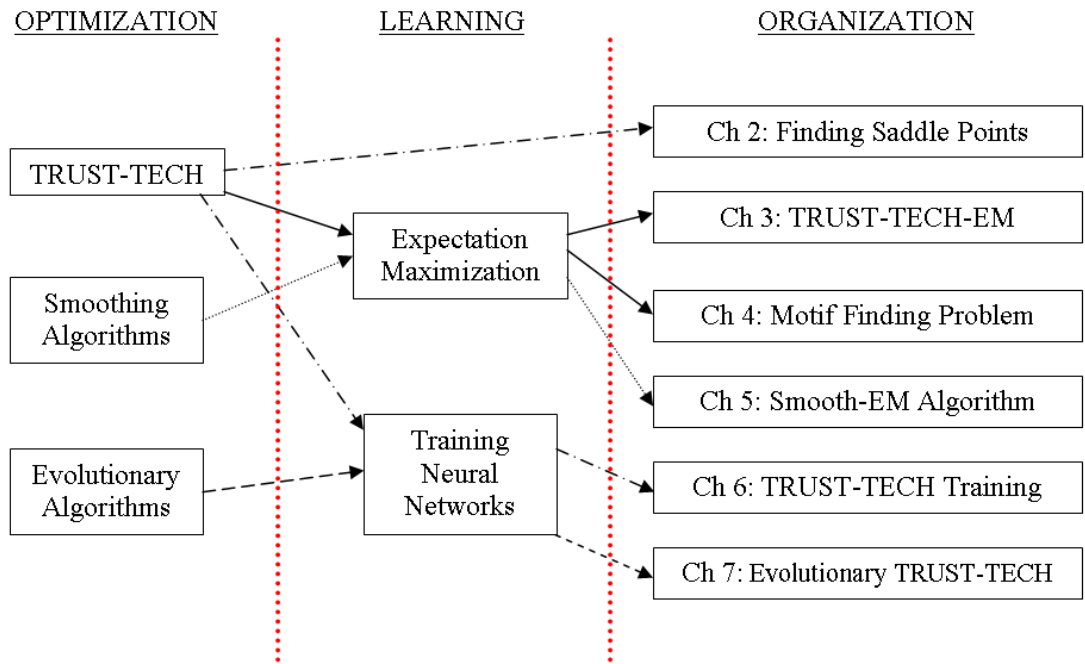, width=6.0in}
} \caption{Organization chart of this thesis. The main contributions are in the areas of optimization and learning.}
\label{fig:org5}
\end{figure}

\section{Organization of this Thesis}
Fig.~\ref{fig:org5} shows the organization chart of this thesis. The main contributions are in the areas of nonlinear optimization relevant to machine learning. TRUST-TECH, smoothing methods and evolutionary algorithms are the optimization methods studied and developed in this thesis. In the context of machine learning, improvements have been proposed for the two most widely studied algorithms, namely Expectation Maximization (in unsupervised learning) and training neural networks (in supervised learning). The remainder of this thesis is organized as follows:  Chapter~\ref{ch:saddle} describes a novel stability boundary based method to find saddle points on potential energy surfaces. A novel TRUST-TECH based Expectation Maximization algorithm for learning mixture models is proposed in Chapter~\ref{ch:trust-tech-em} and this algorithm is applied to the motif finding problem in bioinformatics in Chapter~\ref{ch:motif}. A Component-wise kernel smoothing algorithm for learning Gaussian mixture models is proposed in Chapter~\ref{ch:smooth}. Application of the TRUST-TECH method for efficient training of neural networks is discussed in Chapter~\ref{ch:training}. Chapter~\ref{ch:evoltrust} proposes evolutionary TRUST-TECH model with some preliminary yet promising results. Finally, Chapter~\ref{ch:conclusion} concludes the discussion and proposes a few future research directions for the algorithms proposed in this thesis.

\chapter{Finding Saddle Points on Potential Energy Surfaces}
\label{ch:saddle}

In this chapter, we will use the concepts of stability regions and stability boundaries to obtain saddle points on potential energy surfaces. The task of finding saddle points on potential energy surfaces
plays a crucial role in understanding the dynamics of a
micro-molecule as well as in studying the folding pathways of
macro-molecules like proteins. It is proposed that the problem of finding the saddle
points on a high dimensional potential energy surface be
transformed into the problem of finding dynamic decomposition points (DDP) of
its corresponding nonlinear dynamical system. A novel stability boundary following procedure is used to trace the stability boundary to compute the DDP; hence the saddle points. The proposed method was successful in finding the saddle points on different
potential energy surfaces of various dimensions. A simplified
version of the algorithm has also been used to find the saddle
points of symmetric systems with the help of some analytical
knowledge. The main advantages and effectiveness of the method are
clearly illustrated with some examples. Promising results of our
method are shown on various problems with varied degrees of
freedom.

\section{Introduction}

Recently, there has been a lot of interest across various
disciplines to understand a wide variety of problems related to
bioinformatics and computational biology. One of the most challenging problems in the field
of computational biology is de-novo protein structure prediction where
the structure of a protein is estimated from some complex energy
functions. Scientists have related the native structure of a
protein structurally to the global minimum of the potential energy
surface of its energy function \cite{Dill97}. If the global
minimum could be found reliably from the primary amino acid
sequence, it would provide us with new insights into the nature of
protein folding. However, understanding the process of protein
folding involves more than just predicting the folded structures
of foldable sequences. The folding pathways in which the proteins
attain their native structure can deliver some important
information about the properties of the protein structure
\cite{Merlo05}.

Proteins usually have multiple stable macrostates \cite{Erman97}.
The conformations associated with one macrostate correspond to a
certain biological function. Understanding the transition between
these macrostates is important to comprehend the interactions of that protein with its environment and to understand the kinetics
of the folding process, we need the structure of the transition
state. Since, it is difficult to characterize these structures by
manual experiments, simulations are an ideal tool for the
characterization of the transition structures. Recently,
biophysicists started exploring the computational methods that can
be used to analyze conformational changes and identify possible
reaction pathways \cite{ Bokinsky03}. In particular, the analysis
of complex transitions in macromolecules has been widely studied
\cite{Henkelman00b}.\\

\begin{figure}[htp]
\centerline{
  \epsfig{figure=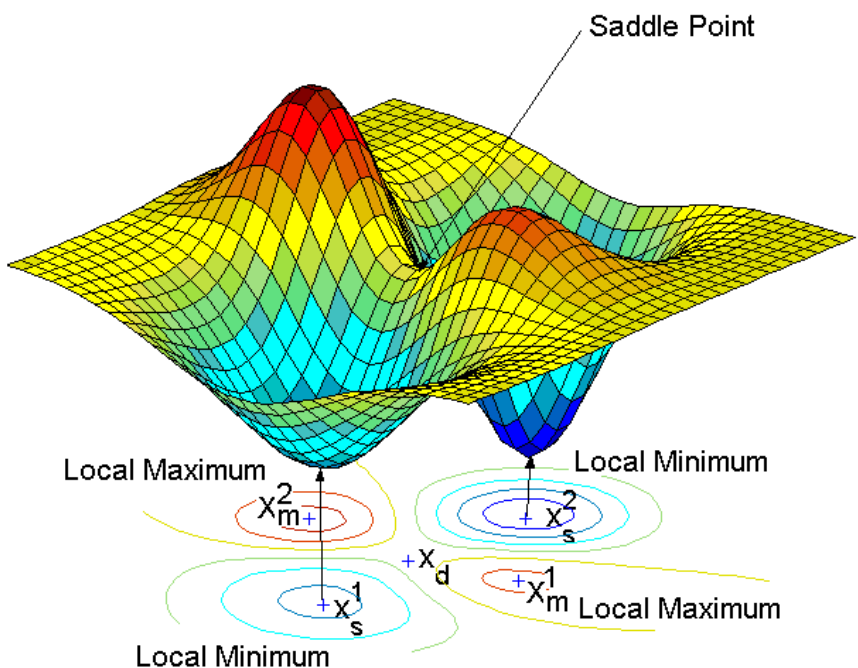, width=4.5in}
} \caption{The surface and contour plots of a two-dimensional
energy function. A saddle point ($x_d$) is located between two
local minima ($x_s^1$ and $x_s^2$). $x_m^1$ and $x_m^2$ are two
local maxima located in the orthogonal direction.}
\label{fig:saddlepoint}
\end{figure}
From a computational viewpoint, transition state conformations correspond to saddle points. {\it Saddle points} are the points on a potential energy surface where the gradient is zero and where the Hessian of
the potential energy function has only one negative eigenvalue
\cite {Heidrich91}. Intuitively, this means that a saddle point is
a maximum along one direction but a minimum along all other
orthogonal directions. Fig. 1 shows a saddle point ($x_d$) located
between two local minima ($x_s^1$ and $x_s^2$) and two local
maxima ($x_m^1$ and $x_m^2$). As shown in the figure, the saddle
point is a maximum along the direction of the vector joining the
two local minima and a minimum along its orthogonal direction (or
the direction of the vector joining the two local maxima). The
direction in which the saddle point is the maximum is usually
unknown in most of the practical problems and is the direction of
interest. This makes the problem of finding the saddle points more
challenging than the problem of finding local minima on a
potential energy surface. In terms of transition states, saddle
points are local maxima with respect to the reaction coordinates
for folding and local minima with respect to all other
coordinates. The search for the optimal transition state becomes a
search for the saddle points on the edge of the potential energy
basin corresponding to the initial state. Finding these
saddle points on potential energy surfaces can provide new insights about the folding mechanism of
proteins. The primary focus of this chapter is to find the saddle points on
different potential energy surfaces with varied degrees of freedom
using TRUST-TECH based method.




\section{Relevant Background}
\label{sec:background}

The task of finding saddle points has been a topic of active
research in the field of computational chemistry for almost two
decades. Recently, there has also been some interest in finding
the saddle points of the Lennard-Jones clusters since it will give
some idea about the dynamics of the system \cite {Doye02}. The
properties of higher-index saddle points have been invoked in
recent theories of the dynamics of supercooled liquids. Since the eigenvalues of the Hessian
matrix can provide some information about the saddle points, several methods based on the idea of diagonalization of the Hessian matrix \cite {Khait95, Baker88, Helgaker91} were proposed in the literature. Some improved methods
dealing with the updates of Hessian matrix have also been proposed
\cite {Quapp98}. Even though these methods appear to find saddle
points accurately, they work mainly for low dimensional systems.
These methods are not practical for higher dimensional problems
because of the tremendous increase in the computational cost.

However, some methods that work without the necessity for computing
the second derivatives have been developed. Because of the
scalability issues, much more importance is given to algorithms
that use only the first derivatives to compute the saddle points. A
detailed description of the methods that work only based on first
derivatives along with their advantages and disadvantages is given
in a recent review paper \cite{Henkelman00b}. The various methods
that are used to find saddle points are drag method
\cite{Henkelman00b}, dimer method \cite{Henkelman99}, self penalty
walk \cite{Czerminski90}, activation relaxation technique
\cite{Barkema96}, ridge method \cite{Ionova93}, conjugate peak
refinement \cite{Fischer92}, DHS method \cite{Dewar84}, Nudged
elastic band \cite{Jonsson98, Henkelman00}, Step and slide
\cite{Miron01}. Continuation methods for finding the saddle points
are described in \cite {Lastras98}. Almost all these methods except
the dimer method are used to identify the saddle point between two
given neighboring local minima. Though the dimer method successfully
finds the saddle points in those cases where only one minimum is
given, it does not have a good control over which saddle point it intends to find.

All these methods start searching for saddle points from the local
minimum itself and hence they need to compute the first
derivative. However, our approach doesn't require the gradient
information starting from the local minima. It will find the
stability boundary in a given direction and then trace the
stability boundary till the saddle point is reached
\cite{Reddy06s}. This tracing of the stability boundary is more
efficient than looking for saddle points in the entire search
space. This work presents a completely novel {\it stability
boundary} based approach to compute the saddle point between two
given local minima. Our method is based on some of the fundamental
results on stability regions of nonlinear dynamical systems
\cite{Chiang88,Chiang96a,Lee04}.

\section {Theoretical Background}
\label{sec:theory} 
Before presenting the details of the TRUST-TECH based methods, we
review some fundamental concepts of nonlinear dynamical systems. The notations, definitions and theorems introduced in this section will hold for the rest of the thesis without any changes unless otherwise explicitly stated. Let us consider an unconstrained search problem on a nonlinear surface defined by the objective function

\begin{equation}
 f(x)\label{eq:problem}
\end{equation}

where $f(x)$ is assumed to be in $C^2(\Re^n,\Re)$.

\begin{lem}\label{def:criticalpoint}
$\bar{x}$ is said to be a {\it critical point} of
(\ref{eq:problem}) if it satisfies the following condition
\begin{equation}
 \nabla f(x)=0\label{eq:criticalpoint}
\end{equation}
\end{lem}
A critical point is said to be {\it nondegenerate} if at the
critical point $\bar{x}\in \Re^n$, \\

\centerline {$d^T\nabla_{xx}^{2}f(\bar{x})d
\not= 0$\hspace{0.05in} ($\forall d \not= 0$).}

We construct the following {\it gradient system} in order
to locate critical points of the objective function
(\ref{eq:problem}):


\begin{equation}
 \frac{dx}{dt}=F(x)=-\nabla f(x)\label{eq:gradientsystem}
\end{equation}

where the state vector $x$ belongs to the Euclidean space $\Re^n$,
and the vector field $F : \Re^n \rightarrow \Re^n$ satisfies the
sufficient condition for the existence and uniqueness of the
solutions. The solution curve of Eq. (\ref{eq:gradientsystem})
starting from x at time $t=0$ is called a {\it trajectory} and it
is denoted by $\Phi(x,\cdot) : \Re \rightarrow \Re^n$. A state
vector $x$ is called an {\it equilibrium point} of Eq.
(\ref{eq:gradientsystem}) if $F(x)=0$.

\begin{lem}\label{def:stableequilibriumpoint}
An equilibrium point is said to be {\it hyperbolic} if the
Jacobian of $F$ at point $x$ has no eigenvalues with zero real
part. A hyperbolic equilibrium point is called a (asymptotically)
{\it stable equilibrium point} (SEP) if all the eigenvalues of its
corresponding Jacobian have negative real part. Conversely, it is
an {\it unstable equilibrium point} if some eigenvalues have a
positive real part.
\end{lem}

An equilibrium point is called a {\it type-k equilibrium point} if
its corresponding Jacobian has exact $k$ eigenvalues with positive
real part. When $k=0$, the equilibrium point is (asymptotically)
stable and it is called a {\it sink} (or {\it attractor}). If
$k=n$, then the equilibrium point is called a {\it source} (or
{\it repeller}).

A dynamical system is completely {\it stable} if every trajectory
of the system leads to one of its stable equilibrium points. The
{\it stable} ($W^s(\tilde{x})$) and {\it unstable}
($W^u(\tilde{x})$) manifolds of an equilibrium point, say
$\tilde{x}$, is defined as:
\begin{center}
\begin{eqnarray}
W^s(\tilde{x}) = \{ x \in \Re^n ~:~
\lim_{t\to\infty}\Phi(x,t)=\tilde{x}\} \\
W^u(\tilde{x}) = \{ x \in \Re^n ~:~
\lim_{t\to-\infty}\Phi(x,t)=\tilde{x}\}
\label{eq:unstablemanifold}
\end{eqnarray}
\end{center}

The {\it stability region} (also called {\it region of
attraction}) of a stable equilibrium point $x_s$ of a dynamical
system (\ref{eq:gradientsystem}) is denoted by $A(x_s)$ and is

\begin{equation}
A(x_s) = \{ x \in \Re^n ~:~ \lim_{t\to\infty}\Phi(x,t)=x_s\}
\label{eq:stabilityregion}
\end{equation}\vspace{0.05in}

The boundary of stability region is called the {\it stability
boundary} of $x_s$ and will be denoted by $\partial A(x_s)$. It
has been shown that the stability region is an open, invariant and
connected set \cite{Chiang88}. From the topological viewpoint, the
stability boundary is a ($n-1$) dimensional closed and invariant
set. A new concept related to the stability regions namely the
{\it quasi-stability region} (or {\it practical stability
region}), was developed in \cite{Chiang96a}.

The {\it practical stability
region} of a stable equilibrium point $x_s$ of a nonlinear
dynamical system (\ref{eq:gradientsystem}), denoted by $A_p(x_s)$
and is

\begin{equation}
A_p(x_s) =int~\overline{A(x_s)} \label{eq:stablemanifold}
\end{equation}

where $\bar{A}$ denotes the closure of A and $int~\bar{A}$ denotes
the interior of $\bar{A}$. $int~\overline{A(x_s)}$ is an open set.
The boundary of practical stability region is called the {\it
practical stability boundary} of $x_s$ and will be denoted by
$\partial A_p(x_s)$.

\begin{figure}[htp]
\centerline{
  \epsfig{figure=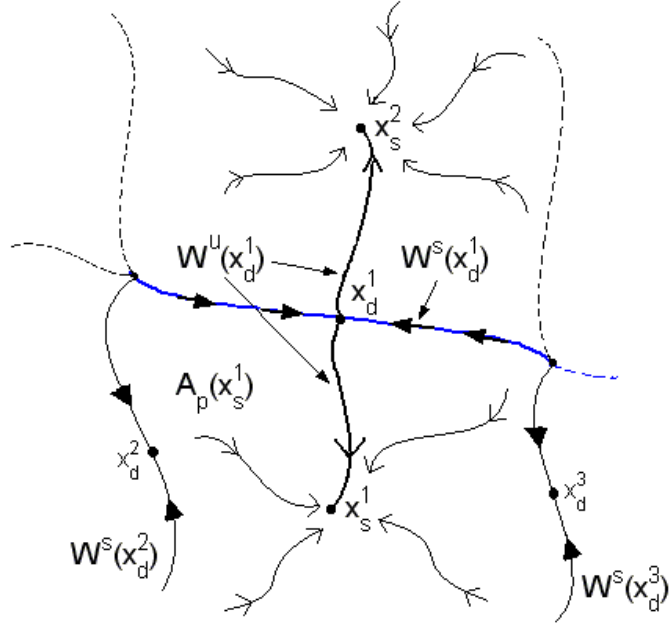, width=3.5in}
} \caption{Phase potrait of a gradient system. The solid lines with solid arrows represent the basin
boundary. $\partial A_p(x_s^1) = \bigcup_{i=1}^{3}
\overline{W^s(x_d^i)}$. The local minima $x_s^1$ and $x_s^2$
correspond to the stable equilibrium points of the gradient
system. The saddle point ($x_d^1$) corresponds to the
dynamic decomposition point that connects the two stable equilibrium points.}\label{fig:saddlepoint1}
\end{figure}

It has been shown that the practical stability boundary $\partial
A_p(x_s)$ is equal to $\partial \bar{A}(x_s)$ \cite{Chiang88}. The practical
stability boundary is a subset of its stability boundary. It
eliminates the complex portion of the stability boundary which has
no ``contact" with the complement of the closure of the stability
region. A complete characterization of the practical stability
boundary for a large class of nonlinear dynamical systems can be
found.

\begin{lem}\label{def:decompositionpoint}
A type-1 equilibrium point $x_d$ (k=1) on the practical stability
boundary of a stable equilibrium point $x_s$ is called a {\it dynamic 
decomposition point}.
\end{lem}
%
%
%


%
To comprehend the transformation, we need to define {\it Lyapunov
function}. A smooth function $V(\cdot) : \Re^n \rightarrow \Re$
satisfying $\dot{V}(\Phi(x,t))~<~0~,~\forall ~x \notin $ \{set of
equilibrium points (E)\} and t $\in \Re^+ $ is termed as Lyapunov
function.

\begin{thm}\label{th:lyapunov}\cite{Chiang96}: $F(x)$ is a
Lyapunov function for the negative quasi-gradient system
(\ref{eq:gradientsystem}).
\end{thm}

%
%
%
%
%
%
%
%


\begin{thm}\label{th:characterization}{\it (Characterization of stability
boundary)\cite{Chiang96a}:} Consider a nonlinear dynamical system
described by (\ref {eq:gradientsystem}). Let $\sigma_i$, i=1,2,... be the equilibrium points on
the stability boundary $\partial A(x_s)$ of a stable equilibrium
point, say $x_s$. Then
\begin{equation}
\partial A(x_s) \subseteq \bigcup_{\sigma_i\in\partial A} W^s(\sigma_i).
\label{eq:charstabilityregion}
\end{equation}
\end{thm}
Theorem \ref{th:characterization} completely characterizes the
stability boundary for nonlinear dynamical systems by asserting that the stability boundary is
the union of the stable manifolds of all critical elements on the
stability boundary. This theorem gives an explicit description of
the geometrical and dynamical structure of the stability boundary.
This theorem can be extended to the characterization of the
practical stability boundary in terms of the stable manifold of the
dynamic decomposition point.

\begin{thm}{\it (Characterization of practical stability
boundary)\cite{Chiang96a}:} Consider a nonlinear dynamical system
described by (\ref {eq:gradientsystem}). Let $\sigma_i$ , i=1,2,... be the dynamic decomposition points
on the practical stability boundary $\partial A_p(x_s)$ of a
stable equilibrium point, say $x_s$. Then

\begin{equation}
\partial A_p(x_s) \subseteq \bigcup_{\sigma_i\in\partial A_p} \overline{W^s(\sigma_i)}.
\label{eq:charpracstabilityregion}
\end{equation}
\label{th:characterization1}
\end{thm}

Theorem \ref{th:characterization1} asserts that the practical
stability boundary is contained in the union of the closure of the
stable manifolds of all the dynamic decomposition points on the practical
stability boundary. Hence, if the dynamic decomposition points can be
identified, then an explicit characterization of the practical
stability boundary can be established using
(\ref{eq:charpracstabilityregion}).

\begin{thm}\label{th:um}{\it (Unstable manifold of type-1 equilibrium
point)\cite{Lastras98}:} Let $x_s^1$ be a stable equilibrium point
of the gradient system (\ref{eq:gradientsystem}) and $x_d$ be a
type-1 equilibrium point on the practical stability boundary
$\partial A_p(x_s)$. Assume that there exist $\epsilon$ and
$\delta$ such that $\| \nabla f(x) \| > \epsilon$ unless $x\in
B_\delta(\hat{x}), \hat{x} \in \{x: \nabla f(x)=0\}$. There exists another
stable equilibrium point $x_s^2$ to which the one dimensional
unstable manifold of $x_d$ converges. Conversely, if
$\overline{A_p(x_s^1)}\bigcap\overline{A_p(x_s^2)}\not=\emptyset$,
then there exists a dynamic decomposition point $x_d$ on $\partial
A_p(x_s^1)$.
\end{thm}

Theorem \ref{th:um} is imperative to understand some of the
underlying concepts behind the development of TRUST-TECH. It
associates the notion of stable equilibrium points, practical
stability regions ($A_p(x_s)$), practical stability boundaries
($\partial A_p(x_s)$) and type-1 equilibrium points. As shown in
fig.~\ref {fig:saddlepoint1}, The unstable manifold ($W^u$) of the
dynamic decomposition point $x_d^1$ converges to the two stable
equilibrium points $x_s^1$ and $x_s^2$. Also, it should be noted
that $x_d^1$ is present on the stability boundary of $x_s^1$ and
$x_s^2$.

We also need to show that under the transformation from
(\ref{eq:problem}) to (\ref{eq:gradientsystem}), the properties of
the critical points remain unchanged. Theorem \ref{th:cpc}
illustrates the correspondence of the critical points of the
original system.



\begin{thm}\label{th:cpc}{\it (Critical Points and their correspondence)\cite{Chiang96}: }
An equilibrium point of (\ref{eq:gradientsystem}) is hyperbolic
if, and only if, the corresponding critical point is
nondegenerate. Moreover, if $\bar{x}$ is a hyperbolic equilibrium
point of (\ref{eq:gradientsystem}), then
\begin {enumerate}
\item {$\bar{x}$ is a stable equilibrium point of
(\ref{eq:gradientsystem}) if and only if $\bar{x}$ is an isolated
local minimum for (\ref{eq:problem})}

\item {$\bar{x}$ is a source of (\ref{eq:gradientsystem}) if and
only if $\bar{x}$ is an isolated local maximum for
(\ref{eq:problem})}

\item {$\bar{x}$ is a dynamic decomposition point of
(\ref{eq:gradientsystem}) if and only if $\bar{x}$ is a saddle
point for (\ref{eq:problem})}
\end {enumerate}
\end{thm}



\section{A Stability Boundary based Method}
\label{sec:algorithm} Our TRUST-TECH based boundary tracing method uses the
theoretical concepts of dynamical systems presented in the
previous section. The method described in this section finds the
DDP when the two neighborhood local minima are
given. Our method is illustrated on a two-dimensional LEPS
potential energy surface \cite{Polanyi69}. The equations
corresponding to the LEPS potential are given in the appendix-A. The
two local minima are $A$ and
$B$ and the dynamic decomposition point is $DDP$.\\
\\{\bf Given :} Two neighborhood local minima (A, B) \\
{\bf Goal :} To obtain the corresponding DDP\\
{\bf Algorithm :}\\
Step1: {\it Initializing the Search direction} : Since the
location of the neighborhood local minima is already given, the
initial search direction becomes explicit. The vector that joins
the two given local minima ($A$ and $B$) is chosen to be the initial search direction.

\begin{figure}[htp]
\centerline{ \epsfig{figure=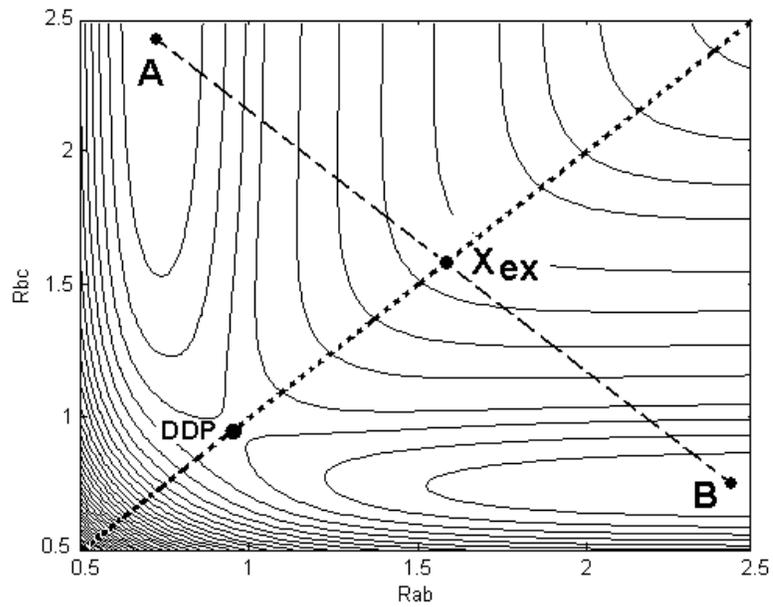, width=4.0in}
}\caption {Contour plot of a 2-D LEPS potential (described in
appendix-A). Each line represents the values of a constant
potential. $A$ and $B$ are the two local minima. $DDP$ is the
dynamic decomposition point to be computed. The search direction is the
direction of the vector joining AB. The exit point ($X_{ex}$) is
obtained by finding the peak of the function value along this
vector. The dotted line indicates the stability boundary. The
dashed line indicates the search direction.} \label{fig:myleps2}
\end{figure}

Step 2: {\it Locating the exit point ($X_{ex}$)} : (see fig.
\ref{fig:myleps2}) Along the direction $AB$, starting from $A$,
the function value is evaluated at different step intervals. Since
the vector is between two given local minima, the function
value will monotonically increase and then decrease till it reaches the
other local minimum ($B$). The
point where the energy value attains its peak is called the {\it exit point}.

Step 3: {\it Moving along the stability boundary to locate the
Minimum Gradient Point} : We used a novel {\it stability boundary following procedure} to move along the practical stability
boundary. Once the exit point is identified, the consecutive
points on the stability boundary can be identified by this
stability boundary following procedure. The exit point ($X_{ex}$) is
integrated for a predefined number of times. Let $m'_1$ be the new
point obtained after integration. The function value between
$m'_1$ and the local minimum is evaluated and the peak value is
obtained. Let the new boundary point along the vector $m'_1B$
starting from the point $m'_1$ and where the value attains the
peak be $m_2$. This process is repeated and several points on the
stability boundary are obtained. During this traversal, the value
of the gradient along the boundary points is noted and the process of
moving along the boundary is terminated when the minimum gradient
point (MGP) is obtained. In summary, the trajectory of integration
is being modified so that it moves towards the MGP and will not
converge to one of the local minima. This is an intelligent
TRUST-TECH based scheme for following the stability boundary which
is the {\it heart} of the proposed method. This step is named as
the stability boundary following procedure.

Step 4: { \it Locating the Dynamic Decomposition point (DDP)} : The Minimum
Gradient Point ($m_n$) obtained from the previous step will be
located in the neighborhood of the dynamic decomposition point. A local
minimizer to solve the system of nonlinear equations is applied
with $m_n$ as initial guess and this will yield the DDP. A detailed survey about different Local minimizations
applied to a wide variety of areas is given in \cite{Schlick02}.

\begin{figure}[htp]
\centerline{ \epsfig{figure=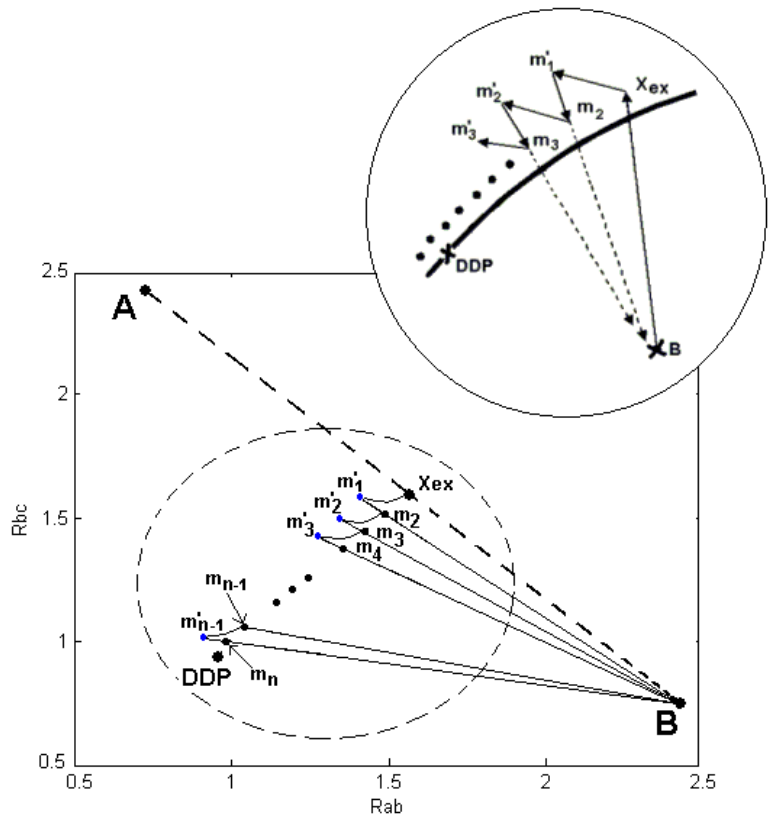, width=4.0in} }
\caption {Illustration of the step 3 of our algorithm. $m_1$ is
integrated till $m'_1$ is reached and traced back along the vector
$m'_1B$ to get $m_2$, and so on. $m_n$ are the $n$ exit points on the
stability boundary. The $n^{th}$ point is the Minimum Gradient
Point where the magnitude of the gradient ($G_M$) is the minimum amongst all the computed gradient values along the stability boundary ($X_{ex}$,$m_2$,$m_3$..$m_n$). DDP is the dynamic decomposition point.} \label{fig:energy}
\end{figure}

{\bf Remarks:}
\begin{itemize}
{\item The step size to be chosen during the step 2 of our
algorithm is very critical for faster computation and accuracy of
the exit points.} {\item The number of integrations to be performed
from a point on the stability boundary ($m_k$) till the new point
($m'_k$) is reached is problem specific and it depends on the
behaviour of the stability boundary near that point.}

{\item The minimum gradient point is usually in the neighborhood of
the DDP. Newton's method is a powerful local solver that can be used to obtain the DDP.}
\end{itemize}

\section{Implementation Issues}
\label{sec:implementation}

For our illustration, we consider a N dimensional function $F$
with variables $X_i$, where $i=1...N$. From the algorithmic
viewpoint, our method consists of three stages: (i) Finding the exit
point, (ii) Following the stability boundary and (iii) Computing the DDP. Let $A$ and $B$ are the given local minima, the pseudocode for finding the DDP is as follows: \\

$point${\bf
\hspace{0.1in}procedure\hspace{0.1in}}$locate\_DDP(A,B)$
\begin{algorithmic}
\STATE Initialize $stepsize=10~$  // initial evaluation step size
\STATE $EP~\leftarrow ~ find\_ExitPt(A,B, stepsize)~$ // Exit
point \STATE $MGP~\leftarrow ~ Boundary\_Following(EP)~$ // Trace
the stability boundary\STATE $DDP~\leftarrow ~ local\_minimizer(MGP)$  // Compute the DDP

\STATE
\bf return $DDP$
\end{algorithmic}

\begin{figure}[htp]
\centerline{
  \epsfig{figure=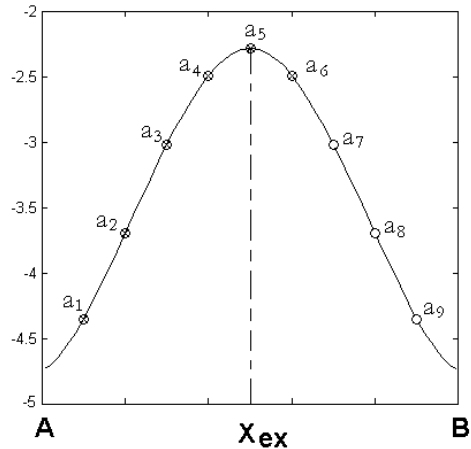, width=2.5in} } \caption{The plot of the function value along
the vector $AB$. The curve monotonically increases starting from
$A$ and then decreases till it reaches $B$. $X_{ex}$ is the exit
point where the function attains its peak value. $a_1$, $a_2$, ...
$a_9$ are the interval points. The marked circles indicate that
the function has been evaluated at these points. The empty circles
indicate the points where the function is yet to be evaluated.}
\label{fig:enecurve}
\end{figure}

The procedure $find\_ExitPt$ will find the point on the stability
boundary between two given points $A$ and $B$ starting from $A$
({\it step 2} of our algorithm). If the function value is
monotonically decreasing from the first step, then it indicates
that there will not be any boundary in that direction, and the
search is changed to a new direction. The new direction can be
obtained by making $B=2A-B$. Finding the exit point can be done
more efficiently by first evaluating the function at comparatively
large step intervals (See Fig.~\ref{fig:enecurve}). The function
evaluation is started from $a_1$ , $a_2$ and so on. Once $a_6$ is
reached, the energy value starts to reduce indicating that the
peak value has been reached. Golden section search algorithm (see
appendix-B) is used to efficiently find the exit point within a
small interval range where the peak value is present. The golden
section search is applied to obtain the exit point $X_{ex}$ within
the intervals $a_4$ and $a_6$. In fact, golden section search could
have been used from the two given local minima. We prefer to
evaluate the function value at certain intervals because using
this method the stability boundary can be identified without
knowing the other local minimum as well.

$point${\bf
\hspace{0.05in}procedure\hspace{0.05in}}$find\_ExitPt(A,B,steps )$
\begin{algorithmic}[1]
\STATE Initialize $eps$  // accuracy \STATE
$interval~=~(B-A)/steps$ \STATE $cur~=~eval(A)$ \STATE
$tmp~=~A~+~interval$
    \IF{$eval(tmp)~<~cur$}
    \STATE $B~=~2*A~-~B$
    \ENDIF
\FOR{$i=1~to~steps$} \STATE  $tmp~=~A~+~i*interval$ \STATE
$prev~=~cur$ \STATE $cur~=~eval(tmp)$
    \IF{$prev~>~cur$}
    \STATE $newPt~=~A~+~(i-2)*interval$
    \STATE $BdPt~=~GoldenSectionSearch(tmp,newPt,eps)$
    \STATE \bf return $BdPt$
    \ENDIF
\ENDFOR \STATE \bf return $NULL$
\end{algorithmic}

\begin{figure}
\begin{center}
 \epsfig{figure=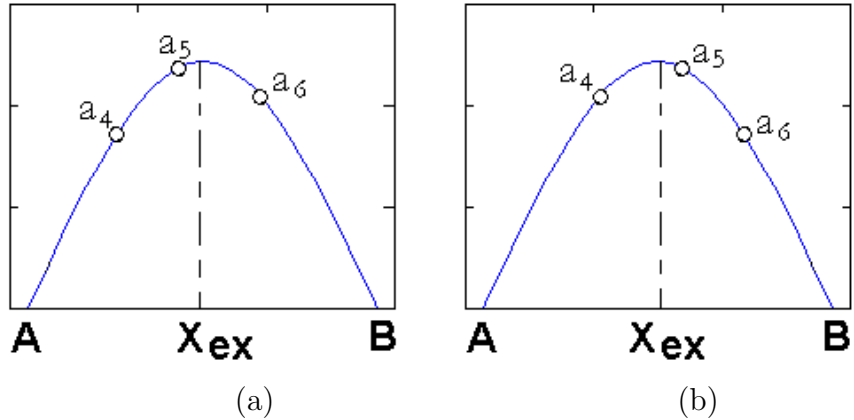, width=4.5in}\\
 \hspace{0.4in} (a) \hspace{2.0in} (b)\\
\end{center}\caption {The interval point $a_5$ can be present
either to the left or to the right of the peak. When $a_6$ is
reached, the golden section search method is invoked with $a_6$
and $a_4$ as the interval.} \label{fig:energycurve}
\end{figure}

Fig. \ref{fig:energycurve} shows the other two possibilities of
having the interval points. It must be noted that the golden
section search procedure should not be invoked with two
consecutive intervals between which the value starts to reduce. It
should be applied to two intervals $a_4$ and $a_6$ because the point $a_5$ can be present on either side of the peak. From Fig.
\ref{fig:energycurve}b, we can see that the value at $a_5$ is
greater than $a_4$ and less than $a_6$ but still the peak is not
present between $a_5$ and $a_6$. This is the reason why the golden
section search method should have $a_4$ and $a_6$ as its
arguments.

In an ideal case, integration from the exit point will lead to the
DDP. However, due to the numerical approximations
made, the integration starting from the exit point will eventually converge to
its corresponding local minimum. Hence, we need to adjust the path
of integration so that it can effectively trace the stability boundary.

$point${\bf
\hspace{0.1in}procedure\hspace{0.1in}}$Boundary\_Following( ExPt)$

\begin{algorithmic}[1]
\STATE Initialize $dt$  // Integral step size

\STATE Initialize $intsteps$  // No. of integration steps

\STATE Initialize $smallstep$  //small step size

\STATE $reduce\_flag~=~OFF$  // To check if gradient reduced

\STATE Store $BdPt~=~ExPt$

\STATE $GM\_next~=~GM(BdPt)$

\WHILE{(1)}

\STATE integrate $BdPt$ for $intsteps$ number of times and with $dt$ step size to obtain $NewPt$ \STATE
$prevPt~=~BdPt$

\STATE Obtain another exit point $BdPt~=~find\_ExitPt(NewPt,~A,~smallstep)$

\STATE $GM\_prev~=~GM\_next$

\STATE Magnitude of the gradient for the new exit point $GM\_next~=~GM(BdPt)$

    \IF{$GM\_next~<~GM\_prev$}
    \STATE $reduce\_flag~=~ON$ // compare the magnitude of the gradient
    \ENDIF

\IF{$GM\_next>GM\_prev$ \& $reduce\_flag=ON$}
    \STATE MGP has been reached $MGPt=prevPt$
    \ENDIF

\STATE \bf return $MGPt$

\ENDWHILE

\end{algorithmic}

%
%
%
%
%
%
%
%
%
%
%
%
%
%
%
%
%

The procedure $Boundary\_Following$ takes in the exit point as the
argument and returns the computed MGP obtained by tracing the
stability boundary starting from the exit point ({\it step 3} of
our algorithm). This function implements the stability boundary following procedure in our algorithm and is responsible for carefully tracing the practical stability boundary. The point on the stability boundary is integrated
for a predefined number of times. From the exit point, it is
integrated using the equation shown below.

\begin{equation}
X_{(i+1)} = X_{(i)} - \left(\frac{\partial F}{\partial
X_{(i)}}\right)~.~ \Delta t
 \label{th:integrate}
\end{equation}

The exit point ($X_{ex}$) is integrated for a predefined number of times. Let $m'_1$ be the new
point obtained after integration. The function value between
$m'_1$ and the local minimum is evaluated and the peak value is
obtained. Let the new boundary point along the vector $m'_1B$
starting from the point $m'_1$ and where the value attains the
peak be $m_2$. This process is repeated and several points on the
stability boundary are obtained. During this traversal, the value
of the gradient along the boundary is noted and the process of
moving along the boundary is terminated when the minimum gradient
point (MGP) is obtained. In summary, the trajectory of integration
is being modified so that it moves towards the MGP and will not
converge to one of the local minima.

The procedure $integrate$ computes a point ($newPt$) by
integrating a certain number of integral steps ($intstep$) with a
predefined integration step size ($\Delta t$) from  the boundary
point ($bdPt$). Once again a new exit point on the practical stability boundary is obtained from
$newPt$ using the procedure $find\_ExitPt$. This process of integrating and tracing back is repeated for a certain number of steps. Another important issue is the stopping criterion for this tracing procedure. For this, we will have to compute the magnitude of the
gradient at all points obtained on the stability boundary. The
magnitude of the gradient ($G_M$) is calculated using Eq.~(\ref{th:GM}).
\begin{equation}
G_M =  \sqrt {\sum_{i=1}^{N}\left(\frac{\partial F}{\partial
X_i}\right)^2 } \label{th:GM}
\end{equation}

\noindent where $\Delta t$ is the integral step size. The $G_M$ value can
either start increasing and then reduce or it might start reducing
from the exit point. The $reduce\_flag$ indicates that the $G_M$
value started to reduce before. $GM\_next$ and $GM\_prev$ are the
two variables used to store the values of the current and previous
$G_M$ values respectively. The $MGP$ is obtained when $GM\_next
> GM\_prev$ and $reduce\_flag=ON$.



\section{Experimental Results}
\label{sec:results}

\subsection {Test Case 1: Two-dimensional Potential Energy Surface}

\begin{figure}[htp]
\centerline{
  \epsfig{figure=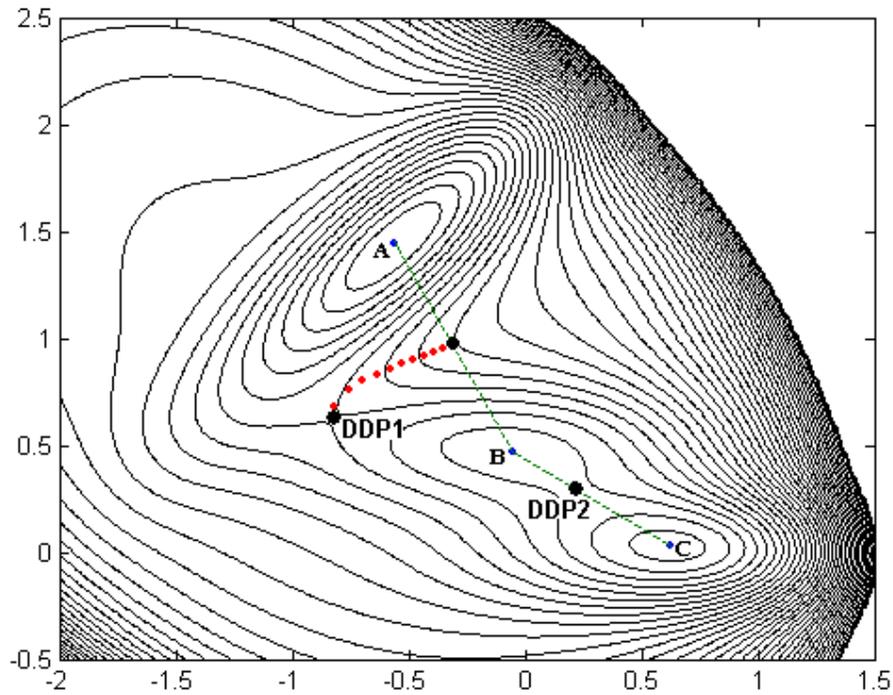, width=4.7in}
} \caption{Two dimensional contour plot of the potential energy
surface corresponding to the muller-brown function described in
Eq. (\ref{eq:mullerbrown}). A,B and C are stable equilibrium points and
DDP1,DDP2 are two dynamic decomposition points. The dashed lines indicate the
initial search direction. The dots indicate the results of the
stability boundary following procedure. These points are the points on
the stability boundary that reach the MGP.}
\label{fig:mullerbrownresult}
\end{figure}

The {\it Muller-Brown surface} is a standard two-dimensional example of
a potential energy function in theoretical chemistry \cite
{Muller79}. This surface was designed for testing the algorithms
that find saddle points. Eq. (\ref {eq:mullerbrown}) gives the
Muller-Brown energy function. Fig. \ref{fig:mullerbrownresult}
represents the two-dimensional contour plot of the potential energy
surface of the muller-brown function.
\begin{equation} \label{eq:mullerbrown}
C(x,y) = \sum_{i=1}^4 A_i~
exp\Bigl[a_i(x-x_i^o)^2 + b_i(x-x_i^o)(y-y_i^o) + c_i(y-y_i^o)^2\Bigr].
\end{equation}
where
\begin{table}[htp]
\begin{center}
\begin{tabular}{l l}
 A  = (-200.0, -100.0, -170.0, -15.0) &   a = ( -1.0, -1.0, -6.5, -0.7) \\
 $x^o$ = (1.0, 0.0, -0.5, -1.0)& b = (0.0, 0.0, 11.0, 0.6)\\
$y^o$ = ( 0.0, 0.5, 1.5, 1.0)   & c = (-10.0, -10.0, -6.5, 0.7)
\end{tabular}
\end{center}
\label{TB:values}
\end{table}

As shown in Fig.~\ref{fig:mullerbrownresult}, there are three
stable equilibrium points (A,B and C) and two dynamic decomposition points
(DDP1,DDP2) on the muller-brown potential energy surface.
DDP 1 is present between A and B and is more
challenging to find, compared to DDP 2 which is
present between B and C. Table~\ref {TB:mullerbrownresults} shows
the exact locations and energy values of the local minima and the
dynamic decomposition points.

\begin{table}[htp]
\centering \caption{\protect Stable equilibrium points and
DDPs for the Muller-Brown Surface described in Eq.
(\ref {eq:mullerbrown}).}
\begin{center}
\begin{tabular}{|c|c|c|}
\hline
Equilibrium Points &  Location & Energy value c($\cdot$) \\
\hline SEP A & (-0.558,1.442) & -146.7 \\
\hline DDP 1 & (-0.822,0.624) & -40.67 \\
\hline SEP B & (-0.05,0.467) & -80.77 \\
\hline DDP 2 & (0.212,0.293) & -72.25 \\
\hline SEP C & (0.623,0.028) & -108.7 \\ \hline
\end{tabular}
\end{center}
\label{TB:mullerbrownresults}
\end{table}

\begin{figure}[htp]
\centerline{
  \epsfig{figure=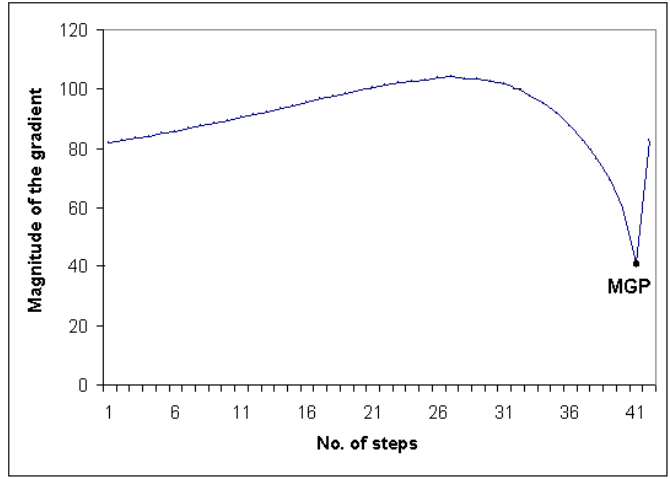, width=3.5in}
} \caption{The gradient curve corresponding to the various points
obtained from the stability boundary following procedure. The graph shows
that the magnitude of the gradient slowly increases in the initial
phases and then starts to reduce. The highlighted point
corresponds to the gradient at the MGP. }
\label{fig:gradientcurve}
\end{figure}

\newpage
We construct the dynamical system corresponding to
(\ref{eq:mullerbrown}) as follows:
\begin{center}
 $~~\left[ \begin{array}{c} \dot{x}(t) \vspace{0.05in}\\
 \dot{y}(t)  \\
\end{array} \right]~~=
 ~-~\left[ \begin{array}{c} \frac{\partial
C}{\partial x} \vspace{0.05in}\\
\frac{\partial C}{\partial y} \\
\end{array} \right]$
%
\begin{eqnarray*}
\frac{\partial C}{\partial x} = \sum_{i=1}^4
A_i~.~P~.~[~2a_i(x-x_i^o)+ b_i(y-y_i^o)~]\\
\frac{\partial C}{\partial y}= \sum_{i=1}^4
A_i~.~P~.~[~b_i(x-x_i^o)+2c_i(y-y_i^o)~]
\label{eq:dcdxy}
\end{eqnarray*}
\end{center}
where
\begin{equation*}
P=exp~[~a_i(x-x_i^o)^2+ b_i(x-x_i^o)(y-y_i^o) +
c_i(y-y_i^o)^2~]\label{eq:derivative}
\end{equation*}
The exit point obtained between the local minima A and B is
(-0.313, 0.971). Fig. \ref{fig:mullerbrownresult} shows the
results of our algorithm on Muller-Brown surface. The dashed lines
indicate the initial search vector which is used to compute the
exit point. The dots indicate the points along the stability
boundary obtained during the stability boundary following procedure. These
dots move from the initial exit point towards the MGP. The gradient curve
corresponding to the points along this stability boundary is shown
in Fig. \ref{fig:gradientcurve}. From the MGP, the local minimizer
is applied to obtain the dynamic decomposition point (DDP 1). The exit point
obtained between the local minima B and C is (0.218, 0.292). It
converges to the dynamic decomposition point (DDP 2) directly when the
local minimizer is applied. Hence, given the three local minima,
we are able to find the to saddle points present between them
using our method.

\begin{table}[htp]
\centering \caption{\protect Stable equilibrium points and
dynamic decomposition points for the Muller-Brown Surface described in Eq.
(\ref {eq:mullerbrown}).}
\begin{center}
\begin{tabular}{|c|c|c|}
\hline
Equilibrium Points &  Location & Energy value c($\cdot$) \\
\hline SEP A & (-0.558,1.442) & -146.7 \\
\hline DDP 1 & (-0.822,0.624) & -40.67 \\
\hline SEP B & (-0.05,0.467) & -80.77 \\
\hline DDP 2 & (0.212,0.293) & -72.25 \\
\hline SEP C & (0.623,0.028) & -108.7 \\ \hline
\end{tabular}
\end{center}
\label{TB:mullerbrownresults}
\end{table}

\subsection {Test Case 2: Three-dimensional symmetric systems}
\label{sec:lennardjones} 
\begin{figure}[htp]
\centerline{ \epsfig{figure=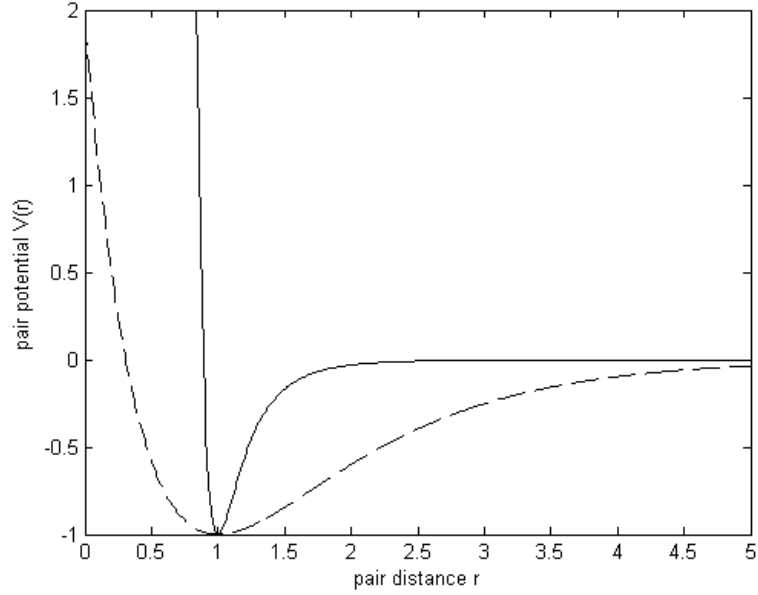, width=4.5in}}
 \caption{Characteristic curves of Lennard-Jones potential and the Morse potential with all parameters set to unity.
 Solid line represents Lennard-Jones potential (\ref{eq:lennjones}) and the dashed line represents the Morse potential (\ref{eq:morsepotential}).} \label{fig:lennardjones}
\end{figure}

{\it Three atom Lennard Jones clusters :}
This system is mainly used to demonstrate a simplified version of
our algorithm. Our method has some advantages when applied to
energy surfaces that are symmetric in nature. To demonstrate this,
we used the Lennard Jones pair potential which is a simple and
commonly used model for interaction between atoms. The Lennard
Jones potential is given by the Eq. (\ref{eq:lennjones}). For
simplicity, we applied reduced units, i.e. the values of
$\epsilon$ and $r_0$ are taken to be unity. Plot of Lennard-Jones
Potential of interaction between two atoms generated using Eq.
(\ref{eq:lennjones}) is shown in Fig. \ref{fig:lennardjones}. The
original problem is to find the global minimum of the potential
energy surface obtained from the interaction between N atoms with
two-body central forces. In this example, we consider the
potential energy surface corresponding to the three-atom cluster
which exhibits symmetric behaviour along the x-axis.
\begin{center}
\begin{eqnarray}
V = \sum_{i=1}^{N-1} \sum_{j=i+1}^N v(r_{ij}) \nonumber \\
v(r_{ij}) = \epsilon\left[\left(\frac{r_0}{r_{ij}}\right)^{12} -
2\left(\frac{r_0}{r_{ij}}\right)^6\right] \label{eq:lennjones}
\end{eqnarray}
\end{center}
where
\centerline {$ r_{ij} = \sqrt {  (x_i - x_j)^2 + (y_i - y_j)^2 +
(z_i - z_j)^2 }$}
The total potential energy ($V$) of the microcluster is the
summation of of all two-body interaction terms, $v(r_{ij})$ is the
potential energy term corresponding to the interaction of atom $i$
with atom $j$, and $r_{ij}$ is the Euclidean distance between $i$
and $j$. $\epsilon$ describes the strength of the interaction and
$r_0$ is the distance at which the potential is zero.

For a three atom cluster, let the coordinates be ($x_1,y_1,z_1$),
($x_2,y_2,z_2$) and ($x_3,y_3,z_3$). Though, there are nine
variables in this system, due to the translational and rotational
variants, the effective dimension is reduced to three. This
reduction can be done by setting the other six variables to zero.
Hence, the effective variables are ($x_2,x_3,y_3$).


We construct the dynamical system corresponding to
(\ref{eq:lennjones}) as follows:

\begin{equation}
 ~~\left[ \begin{array}{c} \dot{x_2}(t) \vspace{0.05in}\\
 \dot{x_3}(t) \vspace{0.05in}\\ \dot{y_3}(t) \vspace{0.05in}\\
\end{array} \right]~~=
 ~-~\left[ \begin{array}{c} \frac{\partial V}{\partial x_2} \vspace{0.05in}\\
\frac{\partial V}{\partial x_3} \vspace{0.05in}\\
\frac{\partial V}{\partial y_3} \vspace{0.05in}\\
\end{array} \right] \label{eq:finalderivative} \end{equation}

 where

%
%
\[ \frac{\partial V}{\partial x_i} =\sum_{\substack{j=1\\j\neq i}}^3
~\frac{12}{(r_{ij})^8}
 ~ \left[ 1 - \left( \frac{1}{r_{ij}} \right)^6 \right]~ . ~
 (x_i-x_j)\]
and\\
\[ \frac{\partial V}{\partial y_i} =\sum_{\substack{j=1\\j\neq i}}^3
~\frac{12}{(r_{ij})^8}
 ~ \left[ 1 - \left( \frac{1}{r_{ij}} \right)^6 \right]~ . ~
 (y_i-y_j)\]


Based on the two given local minima, one can compute the exit
point analytically (not numerically) since the system is
symmetric. Since the exit point is an exact (not an approximate)
value, one can eventually reach the dynamic decomposition point by
integrating the exit point. The exit point is (0.0,0.0,0.0),
(2.0,0.0,0.0), (1.0,0.0,0.0).


\begin{table}[htp]
\centering \caption{\protect Stable equilibrium points and
dynamic decomposition points for the three atom Lennard Jones Cluster
described in Eq. (\ref {eq:lennjones}).}
\begin{center}
\begin{tabular}{|c|c|c|}
\hline
{\bf \scriptsize Equilibrium Point} &  {\bf \scriptsize Location} & {\bf \scriptsize Energy value c(.)} \\
\hline

SEP A &$\begin{tabular}{c}(1.0,0.5,0.866) \\ \end{tabular}$&-3.000\\

\hline DDP 1 & $\begin{tabular}{c} (2.0,1.0,0.0)  \\ \end{tabular}$ & -2.031\\
 \hline SEP B & $\begin{tabular}{c}
(1.0,0.5,-0.866)  \\ \end{tabular}$ &-3.000 \\

\hline
\end{tabular}
\end{center}
\label{TB:results}
\end{table}
In this case, the stability boundary following procedure is not needed.
Integrating from the exit point will eventually find the
dynamic decomposition point. The last two steps of our method,
stability boundary following procedure and the local minimizer are not
required to obtain the dynamic decomposition point. It is clear from the
example above that in all cases where we compute the exit point
numerically, we get an approximate of the exit point which will
eventually converge to one of the two local minima after
integration. In such cases, the stability boundary following procedure
will guide us to maintain the path along the stability boundary
and prevent us from being trapped in one of the local minima.
Hence, a simplified version of our algorithm is developed for
finding saddle points on symmetric surfaces.

\begin{figure}[htp]
\centerline{
  \epsfig{figure=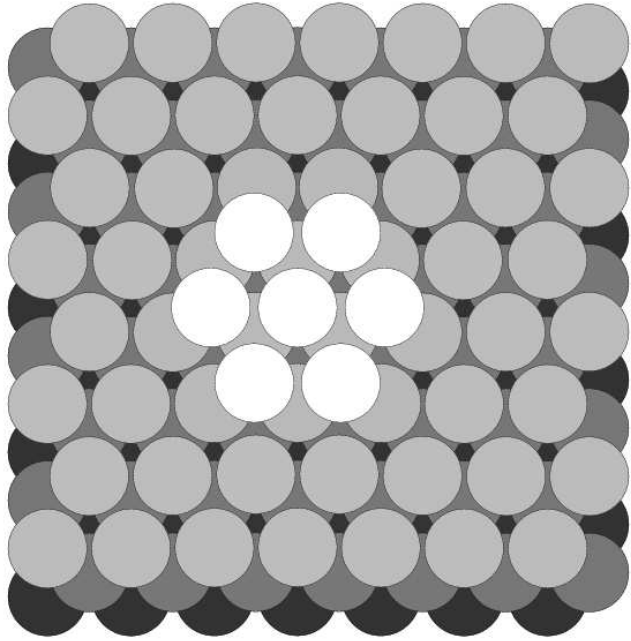, width=3.5in}
} \caption{Top view of the surface and the seven atom island on
the surface of an FCC crystal. The shading indicates the height of
the atoms.} \label{fig:minimum1}
\end{figure}
\subsection {Test Case 3: Higher Dimensional Systems}
\label{sec:pes} 
\begin{figure}[htp]
\begin{center}
\epsfig{figure=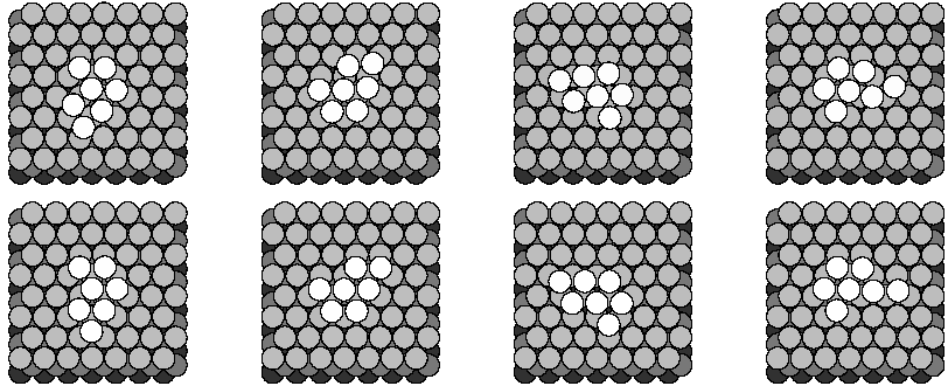, width=5.0in}
\end{center}
\caption{Some sample configurations of the heptamer island on the
surface of the FCC crystal. First row- saddle points
configurations. Second row- other local minimum energy
configurations corresponding to the above saddle point
configurations.} \label{fig:atoms}
\end{figure}

{\it Heptamer island on a crystal:} To test our
method on a higher dimensional system, we have chosen a heptamer
island on the surface of an Face-Centered Cubic (FCC) crystal. This
system will not only illustrate the atomic scale mechanism of island
diffusion on surfaces but also will help us to understand the
kinetics of a process. The atoms interact via a pairwise additive
Morse potential described by Eq. (\ref {eq:morsepotential}).
\begin{equation}
V(r)\ = \ A~\left( e^{-2\alpha (r-r_0)}-2 e^{-\alpha
(r-r_0)}\right) \label{eq:morsepotential}
\end{equation}
where $A=0.71$~eV, $\alpha=1.61$~\AA$^{-1}$,\
$r_0=2.9$~\AA.
These parameters were chosen in such a way that it will reproduce
diffusion barriers on real surfaces. The potential was cut and
shifted at $9.5$~\AA. The surface is simulated with a 6 layer
slab, each layer containing 56 atoms. The minimum energy lattice
constant for the FCC solid is  $2.74$~\AA. The bottom three layers
in the slab are held fixed.  A total of 7 + 168 = 175 atoms are
allowed to move during the search for dynamic decomposition points. Hence,
this is an example of 525 (175 X 3) dimensional search problem.
This is the same system used in the review paper by Henkelman et
al \cite{Henkelman00b}. Fig. \ref{fig:minimum1} shows the top view
of the initial configuration of the island with a compact heptamer
sitting on top of the surface. The shading indicates the height of
the atoms. The white ones are the heptamer island on the surface
of the crystal and the black ones are at the bottom most part of
the crystal. Fig.~\ref{fig:atoms} shows some sample configurations corresponding to saddle points and local minima on the potential energy surface.

Let $N$ be the number of atoms that can move (175) and $N'$ be the
total number of atoms (343).
\begin{equation}
V(r)\ = \sum_{i=1}^N ~\sum_{j=1}^{N'} ~\ A~\left( e^{-2\alpha
(r_{ij}-r_0)}-2 e^{-\alpha (r_{ij}-r_0)}\right) \label{eq:energy}
\end{equation}

where $r_{ij}$ is the Euclidean distance between $i$ and $j$. The
\emph{Dynamical System} is a $3N$~ column matrix given by :
\[ \textstyle{ \left[ \dot{x_1}(t) ~ \dot{x_2}(t)~..~\dot{x_n}(t)~\dot{y_1}(t) ~ \dot{y_2}(t)~..~\dot{y_n}(t)~\dot{z_1}(t) ~ \dot{z_2}(t)~..~\dot{z_n}(t)\right]^T}
\]\[=~ -~  \left[ \begin{array}{c} \frac{\partial V}{\partial
x_1}~\frac{\partial V}{\partial x_2}~..~\frac{\partial V}{\partial
x_n}~\frac{\partial V}{\partial y_1}~\frac{\partial V}{\partial
y_2}~..~\frac{\partial V}{\partial y_n}~\frac{\partial V}{\partial
z_1}~\frac{\partial V}{\partial z_2}~..~\frac{\partial V}{\partial
z_n}
 \end{array} \right]^T \]
where
\begin{equation}
\frac{\partial V}{\partial x_i} =\sum_{\substack{j=1\\j\neq i}}^n
\ 2\alpha A~\left( e^{-\alpha (r-r_0)}- e^{-2\alpha
(r-r_0)}\right)  . \frac{(x_i-x_j)}{r_{ij}} \label{eq:derivative}
\end{equation}

and derivatives are computed with respect to $y_i$ and $z_i$ in a
similar manner.

%
%

\begin{figure}[htp]
\centerline{
  \epsfig{figure=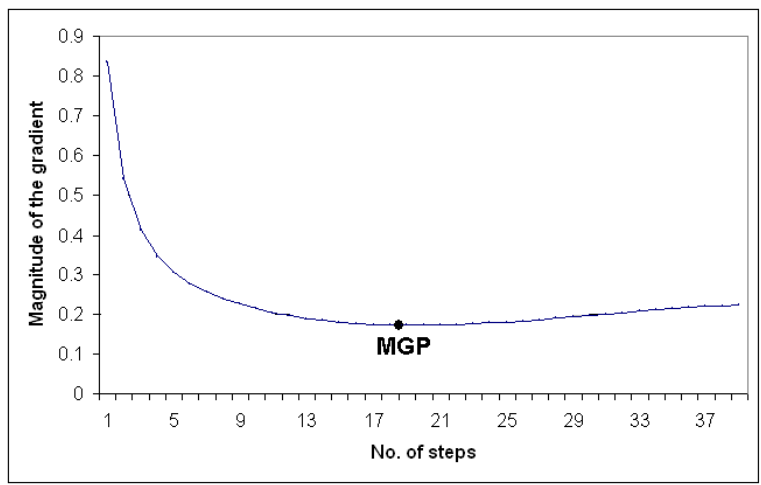, width=4.0in}
} \caption{The gradient curve obtained in one of the higher
dimensional test cases. MGP indicates the minimum gradient point where the magnitude of the gradient reaches the minimum value. This point is used as a initial guess for a local optimization method to obtain DDP.} \label{fig:gradientcurve2}
\end{figure}

\begin{table}
\centering \caption{\protect Results of our algorithm on a
heptamer island over the FCC crystal. The number of force
evaluations made to compute the MGP is given in the last column. }
\begin{center}
\begin{small}
\begin{tabular*}{1\textwidth}{@{\extracolsep{\fill}}|c|c|c|c|c|c|c|c|}
\hline
No. & $E_{min}$ & $E_{saddle}$ & $E_{MGP}$ &$RMSD$& $\Delta$Energy & $\Delta$ Force & fevals \\

\hline 1 & -1775.7787   & -1775.19 & -1775.2139 & 0.00917 &
0.02376 & 0.01833 & 49 \\
\hline 2 & -1775.7787  & -1775.1716
 & -1775.1906& 0.00802 &  0.01903 & 0.01671
& 43\\
\hline 3 & -1775.0079 &-1774.8055  & -1774.8213 & 0.0121 & 0.01578 & 0.01664& 97\\
\hline 4 & -1775.006  & -1774.8041  & -1774.9228
 & 0.03208 & 0.11868
& 0.02654& 67\\
\hline 5 & -1775.0058  & -1774.8024  & -1774.8149 & 0.01229 & 0.01256 & 0.01704 & 91\\
\hline 6 & -1775.0942  &-1774.5956  & -1774.3819 & 0.04285 &
-0.21362
 & 0.04343
& 37\\
\hline 7 & -1775.0931 &-1774.5841   & -1774.3916 &0.03296  &
-0.19252
 & 0.03905& 43\\
\hline 8 & -1775.01 &-1774.3106  & -1775.0789
 & 0.05287  & 0.76832
 & 0.04031& 97\\
\hline 9 & -1775.0097 &-1774.3082
  & -1775.0848
 & 0.05297 & 0.77662 & 0.04113
& 97\\
\hline 10 & -1774.3896
 &-1774.2979
 & -1774.9551 & 0.05623  & 0.65718
 & 0.03103& 79\\
\hline 11 & -1774.3928  &-1774.2997   & -1774.9541
 & 0.05615  & 0.65439
 & 0.03086
& 79\\
\hline 12 & -1774.3933 &-1774.2792   & -1774.2938
 & 0.02262 & 0.01450& 0.07092& 19\\
\hline
\end{tabular*}
\end{small}
\end{center}
\label{TB:highdimension}
\end{table}

To illustrate the importance of the minimum gradient point and the
effectiveness of the stability boundary tracing, we compared our results
with other methods reported in \cite{Henkelman00b}. Energy value
at the given local minimum is -1775.7911. $E_{min}$ is the energy
value at the new local minimum. $E_{saddle}$ is the energy value
at the saddle point. $E_{MGP}$ is the energy value at the Minimum
Gradient Point. $RMSD$ is the root mean square distance of all the
atoms at the MGP and the saddle point. Fig.~\ref{fig:gradientcurve2} shows the gradient curve for one of the higher dimensional test cases. $\Delta$Energy is the
energy difference between $E_{MGP}$ and $E_{saddle}$. $\Delta$
Force is the magnitude of the gradient at the MGP. The results of
our method is shown in Table \ref{TB:highdimension}. The last
column indicates the number of gradient computations that were
made to reach the minimum gradient point. As seen from the table,
our method finds the saddle points with fewer number of gradient
computations when compared to other methods. Typically, even the
best available method takes at least 200-300 evaluations of the
gradient. For detailed results about the performance of other
methods refer to \cite{Henkelman00b}. The MGP that was computed in
our case varied from 0.01-0.05. The MGP can be treated as a saddle
point for most of the practical applications. The RMSD (root mean
square distance) value between the MGP and the saddle point is
very low.

\subsection{Special Cases : Eckhardt surface}

\begin{figure}[htp]
\centerline{
  \epsfig{figure=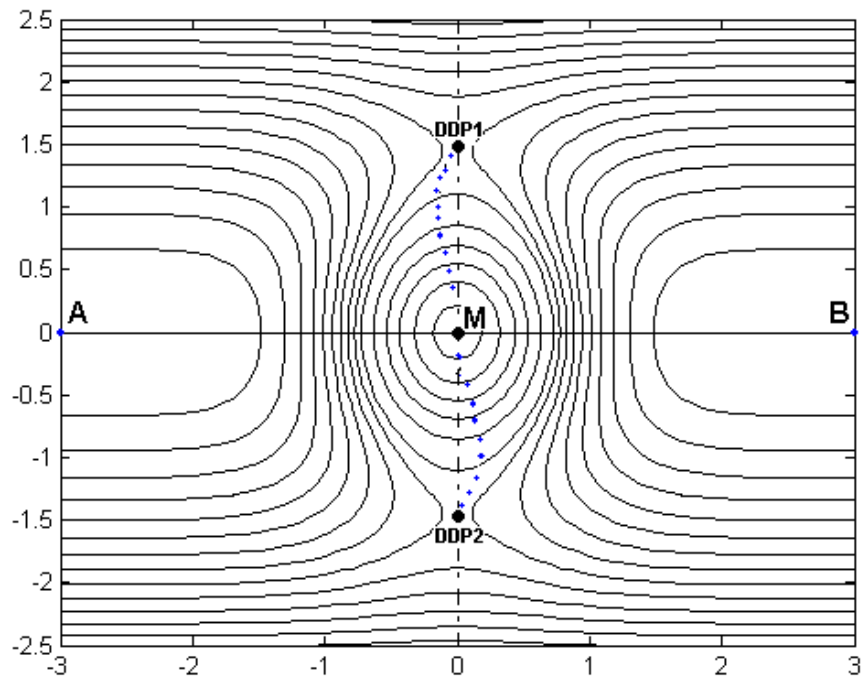, width=4.5in}
} \caption{Two dimensional contour plot of the potential energy
surface of the Eckhardt energy function described on Eq.
(\ref{eq:eckhardt}). A and B are stable equilibrium points and DDP1,DDP2
are two dynamic decomposition points. M is a source. The dots correspond to the points on
the stability boundary during the stability boundary following
procedure.} \label{fig:eckhardt}
\end{figure}

The Eckhardt surface \cite {Eckhardt88} is an exceptional case where
we need to perturb the exit point in order to follow the stability
boundary. Such cases almost never occur in practice and hence
dealing with such surfaces will not be given much importance. Eq.
(\ref {eq:eckhardt}) gives the Eckhardt energy function. Fig.
\ref{fig:eckhardt} represents the two-dimensional contour plot of
the potential energy surface of the Eckhardt function.

\begin{equation} \label{eq:eckhardt}
C(x,y) = e^{ -[x^2 +(y+1)^2]} + e^{-[x^2+(y-1)^2]}+ 4~e^{-[3(x^2+y^2)/2]}+y^2/2
\end{equation}

As shown in Fig. \ref {fig:eckhardt}, there are two local minima
(A,B), two dynamic decomposition points (1,2) on the Eckhardt potential
energy surface. A local maximum is present exactly at the center
of the vector joining the two local minima. The two dynamic decomposition points are on
either side of the maximum. Table~\ref{TB:Eckhardtresults} shows
the energy values at the local minima and the dynamic decomposition
points.

\newpage
We construct the dynamical system corresponding to
(\ref{eq:eckhardt}) as follows:
\begin{center}
 $~~\left[ \begin{array}{c} \dot{x}(t) \vspace{0.05in}\\
 \dot{y}(t)  \\
\end{array} \right]~~=
~-~\left[ \begin{array}{c} \frac{\partial
C}{\partial x} \vspace{0.05in}\\
\frac{\partial C}{\partial y} \\
\end{array} \right]$
\end{center}

\begin{equation*}
\frac{\partial C}{\partial x} =
-2x~e^{-[x^2+(y+1)^2]}-2x~e^{-[x^2+(y-1)^2]}
-12x~e^{-[3(x^2+y^2)/2]} \label{eq:dcdx}
\end{equation*}
\begin{equation*}
\frac{\partial C}{\partial y} =
-2(y+1)~e^{-[x^2+(y+1)^2]}-2(y-1)~e^{-[x^2+(y-1)^2]}-12y~e^{-[3(x^2+y^2)/2]}+y
\label{eq:dcdy}
\end{equation*}

\begin{figure}[htp]
\centerline{
  \epsfig{figure=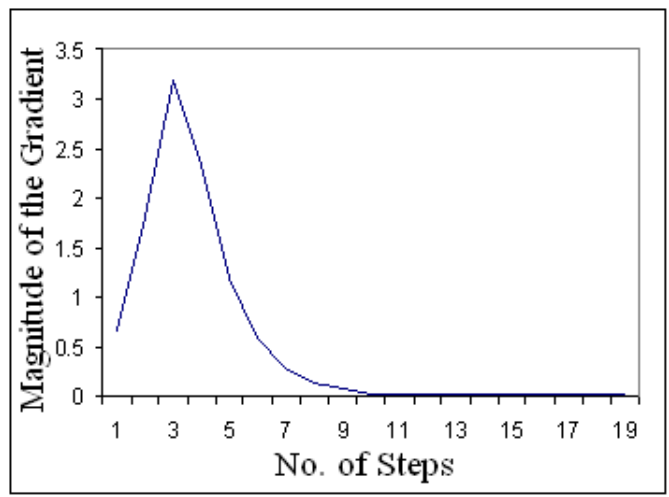, width=3.5in}
} \caption{Gradient curve corresponding to the various points
along the stability boundary on the Eckhardt surface.}
\label{fig:eckhardtgradientcurve}
\end{figure}
\begin{table}
\centering \caption{\protect Stable equilibrium points and
dynamic decomposition points for the Eckhardt Surface.}
\begin{center}
\begin{tabular}{|c|c|c|}
\hline
Equilibrium Points &  Location & Energy value c(.) \\
\hline SEP A & (-3.0,0.0) & 0.0 \\
\hline DDP 1 & (0.0,1.4644) & 2.0409 \\
\hline SEP B & (3.0,0.0) & 0.0 \\
\hline DDP 2 & (0.0,-1.4644) & 2.0409 \\
\hline SEP M & (0.0,0.0) & 4.7358 \\ \hline
\end{tabular}
\end{center}
\label{TB:Eckhardtresults}
\end{table}
This surface can be treated as a special case where there are
different critical points lying on the vector joining the two
local minima. As seen from the Fig.~\ref{fig:eckhardt}, there is a local maximum
at (0,0), which is the exit point obtained. It should be noted
that this system is also symmetric and hence one can obtain the
exit point analytically. Since the exit point is also a critical
point, the exit point is first perturbed and then the stability boundary following procedure is used to compute the MGP. The two
dynamic decomposition points are at (0.0,1.4644) and (0.0,-1.4644). The
gradient curve is shown in Fig. \ref{fig:eckhardtgradientcurve}.
From this MGP, the local minimizer is applied to obtain the
DDP1. The other dynamic decomposition point (DDP2) is
similarly obtained. Our method was also successful in finding
saddle points on other two dimensional test surfaces like Minyaev
Quapp \cite{Minyaev97}, NFK (Neria-Fischer-Karplus)
 \cite{Neria96} etc.

\section{Discussion}
\label{sec:conclusion}

Saddle points play a vital role in realizing the folding pathways
of a protein as well as in understanding the transition state
structures during chemical reactions. This chapter primarily focuses
on a new TRUST-TECH based method for finding saddle points on
potential energy surfaces using stability boundaries. Our approach
is based on some fundamental results of nonlinear dynamical
systems. A novel {\it stability boundary following procedure} has been
used to move along the stability boundary. Our method was able to
find the saddle points on a wide range of surfaces with varying
dimensions. The primary advantage of our method comes from the
fact that the stability boundary following is computationally more
efficient than directly searching for a saddle point from the
given local minimum. Deterministic nature of the algorithm ensures
that the same saddle point is obtained every run. Very few
user-specific parameters makes it easy for a new user to
implement. We have also explored the symmetric behaviour of some
energy surfaces to obtain the exit point analytically and
developed a simplified version of our algorithm to compute the
saddle points. The algorithm has also been tested successfully on
a heptamer island over the surface of an FCC crystal.

Finding saddle points can be of importance for other problems related to global optimization. This can be done by using our method presented here as a tool for {\it escaping from a given local minimum
to another local minimum in the neighborhood}. Though the current
work assumes that the two local minima between which the saddle
point is computed are given, it can be easily extended to find
saddle points from a single local minimum.
\newpage
\section*{APPENDIX-A:  LEPS Potential} \label{sec:appendix-A}
The model of LEPS potential simulates a reaction involving three
atoms confined to motion along a line. Only one bond can be formed
either between atoms A and B or between atoms B and C.

\begin{eqnarray*}
 \label{eq:lepspotential}
 C(x,y)=
\frac{Q_{AB}}{1+a}+\frac{Q_{BC}}{1+b}+\frac{Q_{AC}}{1+c}-\Bigl[\frac{J_{AB}^2}{(1+a)^2}+\frac{J_{BC}^2}{(1+b)^2}+\frac{J_{AC}^2}{(1+c)^2}+
\frac{J_{AB}J_{BC}}{(1+a)(1+b)}\\
 +\frac{J_{BC}J_{AC}}{(1+b)(1+c)}+\frac{J_{AB}J_{AC}}{(1+a)(1+c)}\Bigr]^{\frac{1}{2}}
\end{eqnarray*}

where the $Q$ functions represent the Coulomb interactions between
electron clouds and the nuclei and the $J$ functions represent the
quantum mechanical exchange interactions. The form of the Q and J
functions is given below:
\begin{equation*}
Q(r)\ =  \frac{d}{2}~\left( \frac{3}{2}e^{-2\alpha (r-r_0)}-
e^{-\alpha (r-r_0)}\right) \label{eq:lepspotential1}
\end{equation*}

\begin{equation*}
J(r)\ = \ \frac{d}{4}~\left( e^{-2\alpha (r-r_0)}-6 e^{-\alpha
(r-r_0)}\right) \label{eq:lepspotential2}
\end{equation*}

The parameters were chosen to be $a=b=c=0.05$, $d_{AB} = d_{AC} =
d_{BC} = 4.746$, $\alpha=1.942$ and $r_0=0.742$. The details about
the LEPS potential are given in \cite{Polanyi69}.

\newpage
\section*{APPENDIX-B: Golden section search}\label{sec:appendix-B}
The following procedure describes the golden section search
method. Let $a$ and $b$ be the two intervals between which the
{\it exit point} is located. Golden section search method computes
the exit point with an accuracy of $\pm \epsilon$.
$r$ is the {\it golden mean} ($\frac {3- \sqrt{5}}{2}$) \cite {Press92}. $f(x)$ returns the function value at point $x$. \\

{\bf procedure\hspace{0.1in}}$GoldenSectionSearch(a,b,\epsilon)$
\begin{algorithmic}[1]
\STATE Initialize $r=0.38197$   (golden mean) \STATE $c=a+r(b-a)$
\STATE $d=b-r(b-a)$
 \WHILE{$|b-a|>\epsilon$}
    \IF{$f(c)>f(d)$}
    \STATE $b=d,~d=c,~c=a+r(b-a)$
    \ELSE
    \STATE $a=c,~c=d,~d=b-r(b-a)$
\ENDIF \ENDWHILE \STATE \bf return $b$
\end{algorithmic}

\chapter{TRUST-TECH based Expectation Maximization for Learning Mixture Models}
\label{ch:trust-tech-em}

In this chapter, we develop a TRUST-TECH based algorithm for solving the problem of mixture modeling. In the field of statistical pattern recognition, finite mixtures
allow a probabilistic model-based approach to unsupervised learning
\cite{McLachlan88}. One of the most popular methods used for fitting
mixture models to the observed data is the {\it
Expectation-Maximization} (EM) algorithm which converges to the
maximum likelihood estimate of the mixture parameters locally
\cite{Demspter77,Redner84}. The usual steepest descent, conjugate
gradient, or Newton-Raphson methods are too complicated for use in
solving this problem \cite{Xu96}. EM has become a popular method
since it takes advantage of problem specific properties. EM based
methods have been successfully applied to solve a wide range of
problems that arise in pattern recognition \cite {Baum70,Bilmes98},
clustering \cite{Banfield93}, information retrieval \cite {Nigam00},
computer vision \cite {Carson02}, data mining \cite{Shumway82}~etc.

\begin{figure}[htp]
\centerline{
  \epsfig{figure=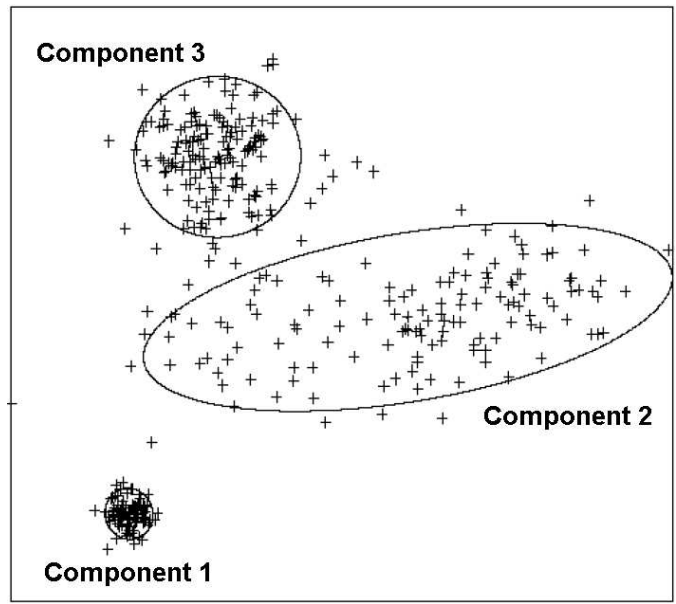, width=3.5in}
} \caption{Data consisting of three Gaussian components with
different mean and variance values. Note that each data point doesn't have
a hard membership that it belongs to only one component. Most of the
points in the first component will have high probability with which
they belong to it. In this case, the other components do not have
much influence. Components 2 and 3 data points are not clearly separated. The
problem of learning mixture models involves estimating the
parameters of the Gaussian components and finding the
probabilities with which each data sample belongs to the component.}
\label{fig:gmm}
\end{figure}

Without loss of generality, we will consider the problem of learning parameters of
Gaussian Mixture Models (GMM). Fig \ref{fig:gmm} shows data
generated by three Gaussian components with different mean and
variance. Note that every data point has a probabilistic (or soft)
membership that gives the probability with which it belongs to each
of the components. Points that belong to component 1 will have high
probability of membership for component 1. On the other hand, data
points belonging to components 2 and 3 are not well separated. The
problem of learning mixture models involves estimating the
parameters of these components and finding the probabilities
with which each data point belongs to these components. Given the
number of components and an initial set of parameters, EM algorithm
computes the optimal estimates of the parameters
that maximize the likelihood of the data given the estimates of
these components. However, the main problem with the EM algorithm is
that it is a `{\it greedy}' method which is very sensitive to the
given initial set of parameters. To overcome this problem, a novel
three-stage algorithm is proposed \cite{Reddy07}. The main research concerns that motivated the new algorithm
presented in this chapter are :
\begin {itemize}
\item{EM algorithm converges to a local maximum
of the likelihood function very quickly.}

\item{There are several other promising local optimal solutions in
the vicinity of the solutions obtained from the methods that
provide good initial guesses of the solution.}

\item{Model selection criteria usually assumes that the global
optimal solution of the log-likelihood function can be obtained.
However, achieving this is computationally intractable.}

\item{Some regions in the search space do not contain any promising solutions.
The promising and non-promising regions coexist and it becomes
challenging to avoid wasting computational resources to search in
non-promising regions.}

\end{itemize}

Of all the concerns mentioned above, the fact that most of the local
maxima are not distributed uniformly \cite{Ueda98} makes it
important to develop algorithms that can avoid searching in the low-likelihood regions and focus on exploring promising subspaces more thoroughly. This
subspace search will also be useful for making the solution less
sensitive to the initial set of parameters. Here, we
propose a novel three-stage algorithm for estimating the parameters of
mixture models. Using TRUST-TECH method and EM algorithm
simultaneously to exploit the problem specific features of the
mixture models, the proposed three-stage algorithm obtains the optimal set of parameters
by searching for the global maximum in a
systematic manner.

\section{Relevant Background}
\label{sec:background}  Although EM and its variants have been extensively used for learning
mixture models, several researchers have approached the problem by
identifying new techniques that give good initial points. More
generic techniques like deterministic annealing \cite
{Rose98,Ueda98}, genetic algorithms \cite{Pernkopf05,Martínez00}
have been applied to obtain a good set of parameters. Though, these
techniques have asymptotic guarantees, they are very time consuming
and hence may not be used for most of the practical applications.
Some problem specific algorithms like split and merge EM
\cite{Ueda00}, component-wise EM \cite{Figueiredo02}, greedy
learning \cite{Verbeek03}, incremental version for sparse
representations \cite{neal98}, parameter space grid \cite{Li99} are
also proposed in the literature. Some of these algorithms are either
computationally very expensive or infeasible when learning mixtures
in high dimensional spaces \cite{Li99}. Inspite of all the expense
in these methods, very little effort has been taken to explore
promising subspaces within the larger parameter space. Most of these
algorithms eventually apply the EM algorithm to move to a locally
maximal set of parameters on the likelihood surface. Simple approaches like running EM from several random
initializations, and then choosing the final estimate that leads to
the local maximum with higher value of the likelihood can be successful to certain extent \cite{hastie96,Roberts98}.

Though some of these methods apply other additional mechanisms (like
perturbations \cite{Elidan02}) to escape out of the local optimal
solutions, systematic methods are yet to be developed for searching
the subspace. The dynamical system of the log-likelihood function
reveals more information about the topology of the nonlinear log-likelihood surface \cite{Chiang96}. Hence, the
difficulties of finding good solutions when the error surface is
very rugged can be overcome by understanding the geometric and dynamic characteristics of the log-likelihood surface. Though this method might introduce some additional cost, one
has to realize that existing approaches are much more expensive due
to their stochastic nature. Specifically, for a problem in this
context, where there is a non-uniform distribution of local maxima,
it is difficult for most of the methods to search neighboring
regions \cite{Zhang04}. For this reason, it is more desirable to
apply TRUST-TECH based Expectation Maximization (TT-EM) algorithm
after obtaining some point in a promising region. The main
advantages of the proposed algorithm are that it :

\begin{itemize}
\item{Explores most of the neighborhood local optimal solutions unlike the traditional stochastic algorithms.}
\item{Acts as a flexible interface between the EM algorithm and other global method. Sometimes, a global method will optimize an approximation of the original function. Hence, it is important to provide an interface between the EM algorithm and the global method.}
\item{Allows the user to work with existing clusters obtained from the traditional approaches and improves the quality of the solutions based on the maximum likelihood criteria.}
\item{Helps the expensive global methods to truncate early.}
\item{Exploits the heuristics that the EM algorithm that it converges at a faster rate if the solutions are promising.}
\end{itemize}

\noindent While trying to obtain multiple optimal solutions, TRUST-TECH can dynamically change the threshold for the number of iterations. For e.g. while computing Tier-1 solutions, if a promising solution has been obtained with a few iterations, then all the rest of the tier-1 solutions will use this value as their threshold.

\section{Preliminaries}
\label{sec:problem} We will now introduce some necessary preliminaries on
mixture models, EM algorithm and nonlinear transformation. Table~\ref{TB:datanot} gives the notations used in this chapter :

\begin{table}[h]
\centering \caption{\protect Description of the Notations
used}
\begin{center}
\begin{tabular}{cl}
\hline  Notation &   Description \\
 \hline

 d & number of features\\
 n & number of data points\\
 k & number of components\\
 s & total number of parameters\\
 $\Theta$ & parameter set\\
 $\theta_i$ & parameters of $i^{th}$ component\\
 $\alpha_i$ & mixing weights for $i^{th}$ component\\
 $\mathcal{X}$ & observed data\\
 $\mathcal{Z}$ & missing data\\
 $\mathcal{Y}$ & complete data\\
 t &  timestep for the estimates\\
 \hline
\end{tabular}
\end{center}
\label{TB:datanot}
\end{table}

\subsection{Mixture Models}

Lets assume that there are $k$ Gaussians in the mixture model. The
form of the probability density function is as follows :

\begin{equation}\label{eq:gaussian1}
    p(x|\Theta) =\sum_{i=1}^{k}{\alpha_i p(x|\theta_i)}
\end{equation}

\noindent where $x=[x_1,x_2,...,x_d]^T$ is the feature vector of $d$
dimensions. The $\alpha_k$'s represent the {\it mixing weights}. $\Theta$ represents the parameter set ($\alpha_1, \alpha_2,...
\alpha_k,\theta_1,\theta_2,...\theta_k$) and $p$ is a univariate
Gaussian density parameterized by $\theta_i$(i.e. $\mu_i$ and
$\sigma_i$):

\begin{equation}\label{eq:gaussiandensity}
    p(x|\theta_i) =\frac{1}{\sqrt{(2\pi)}\sigma_i}e^{-\frac{(x-\mu_i)^2}{2\sigma_i^2}}
\end{equation}

Also, it should be noticed that being probabilities $\alpha_i$ must
satisfy

\begin{equation}\label{eq:probabi}
    0 \leq \alpha_i\leq 1 ~,~ \forall i=1,..,k,~ and ~~ \sum_{i=1}^k
    \alpha_i=1
\end{equation}

Given a set of n i.i.d samples
$\mathcal{X}=\{x^{(1)},x^{(2)},..,x^{(n)}\}$, the log-likelihood
corresponding to a mixture is

\begin{equation}\label{eq:log}
    log ~p(\mathcal{X}|\Theta)=log \prod_{j=1}^n
    ~p(x^{(j)}|\Theta)\\
     =\sum_{j=1}^n log \sum_{i=1}^k \alpha_i
    ~p(x^{(j)}|\theta_i)
\end{equation}

The goal of learning mixture models is to obtain the parameters
$\widehat{\Theta}$ from a set of n data points which are the samples
of a distribution with density given by (\ref{eq:gaussian1}). The
{\it Maximum Likelihood Estimate } (MLE) is given by :
\begin{equation}\label{eq:MLE}
\widehat{\Theta}_{MLE} = arg \max_{\tilde{\Theta}} ~\{~log
~p(\mathcal{X}|\Theta)~\}
\end{equation}

where $\tilde{\Theta}$ indicates the entire parameter space. Since,
this MLE cannot be found analytically for mixture models, one has to
rely on iterative procedures that can find the global maximum of
$log~ p(\mathcal{X}|\Theta)$. The EM algorithm described in the next
section has been used successfully to find the local maximum of such
a function \cite{McLachlan97}.

\subsection{Expectation Maximization}

The EM algorithm assumes $\mathcal{X}$ to be $observed$ data. The
missing part, termed as $hidden$ data, is a set of {\it n} labels
$\mathcal{Z}=\{\footnotesize{\bf z}^{(1)},\footnotesize{\bf
z}^{(2)},..,\footnotesize{\bf z}^{(n)}\}$ associated with $n$
samples, indicating which component produced each sample
\cite{McLachlan97}. Each label $\footnotesize{\bf
z}^{(j)}=[z_1^{(j)},z_2^{(j)},..,z_k^{(j)}]$ is a binary vector
where $z_i^{(j)}=1$ and $z_m^{(j)}=0$  $\forall m \neq i$, means the
sample $x^{(j)}$ was produced by the $i^{th}$ component. Now, the
complete log-likelihood i.e. the one from which we would estimate
$\Theta$ if the {\it complete data}
$\mathcal{Y}=~\{~\mathcal{X},\mathcal{Z}~\}$ is

\begin{equation*}\label{eq:beforecomplete}
    log ~p(\mathcal{X},\mathcal{Z}|\Theta)=\sum_{j=1}^n ~log \prod_{i=1}^k
    ~ [~\alpha_i~p(x^{(j)}|\theta_i)~]^{z_i^{(j)}}
\end{equation*}

\begin{equation}\label{eq:complete}
    log ~p(\mathcal{Y}|\Theta)=\sum_{j=1}^n \sum_{i=1}^k
    z_i^{(j)}~log~ [~\alpha_i~p(x^{(j)}|\theta_i)~]
\end{equation}

The EM algorithm produces a sequence of estimates
$\{\widehat{\Theta}(t),t=0,1,2,...\}$ by alternately applying the
following two steps until convergence :

\begin {itemize}
\item{{\bf E-Step : } Compute the conditional expectation of the
hidden data, given $\mathcal{X}$ and the current estimate
$\widehat{\Theta}(t)$. Since $log~p(\mathcal{X,Z}|\Theta)$ is linear
with respect to the missing data $\mathcal{Z}$, we simply have to
compute the conditional expectation $\mathcal{W} \equiv
E[\mathcal{Z|X},\widehat{\Theta}(t)]$, and plug it into $log ~p
(\mathcal{X,Z}|\Theta)$. This gives the $Q$-function as follows :

 \begin{equation}\label{eq:qfunc}
    Q(\Theta|\widehat{\Theta}(t))\equiv
   E_Z[log~p(\mathcal{X},\mathcal{Z})|\mathcal{X},\widehat{\Theta}(t)]
\end{equation}

Since $\mathcal{Z}$ is a binary vector, its conditional expectation
is given by :

\begin{equation}\label{eq:conditionw}
    w_i^{(j)} \equiv E~[~z_i^{(j)}|\mathcal{X},\widehat{\Theta}(t)~] \\
 = Pr~[~z_i^{(j)}=1|x^{(j)},\widehat{\Theta}(t)~]\\
=
    \frac{\widehat{\alpha}_i(t) p(x^{(j)}|\widehat{\theta}_i(t))}{\sum_{i=1}^{k}{\widehat{\alpha}_i(t) p(x^{(j)}|\widehat{\theta}_i(t))}}
\end{equation}

where the last equality is simply the Bayes law ($\alpha_i$ is the a
priori probability that $z_i^{(j)}=1$), while $w_i^{(j)}$ is the a
posteriori probability that $z_i^{(j)}=1$ given the observation
$x^{(j)}$.}

\item{{\bf M-Step : } The estimates of the new parameters are
updated using the following equation :
\begin{equation}\label{eq:update}
    \widehat{\Theta} (t+1) = arg \max_{\Theta}\{Q(\Theta,\widehat{\Theta}(t))\}
\end{equation}
}
\end{itemize}
\subsection{EM for GMMs}

Several variants of the EM algorithm have been extensively used to
solve this problem. The convergence properties of the EM algorithm
for Gaussian mixtures are thoroughly discussed in \cite{Xu96}. The
$Q-function$ for GMM is given by :

 \begin{equation}\label{eq:qfuncgmm}
 Q(\Theta|\widehat{\Theta}(t))= \sum_{j=1}^{n}\sum_{i=1}^{k}  w_i^{(j)}[log\frac{1}{\sigma_i\sqrt{2\pi}} \\-\frac{(x^{(j)}-\mu_i)^2}{2\sigma_i^2}+log ~\alpha_i]
\end{equation}

where
 \begin{equation}\label{eq:expectz}
w_i^{(j)}=\frac{\frac{\alpha_i(t)}{\sigma_i(t)}e^{-\frac{1}{2\sigma_i(t)^2}(x^{(j)}-\mu_i(t))^2}}{\sum_{i=1}^k
\frac{\alpha_i(t)}{\sigma_i(t)}e^{-\frac{1}{2\sigma_i(t)^2}(x^{(j)}-\mu_i(t))^2}}
 \end{equation}

The maximization step is given by the following equation :
 \begin{equation}\label{eq:max}
    \frac{\partial }{\partial \Theta_k} Q(\Theta|\widehat{\Theta}(t))= 0
\end{equation}
where $\Theta_k$ is the parameters for the $k^{th}$ component.
Because of the assumption made that each data point comes from a
single component, solving the above equation becomes trivial. The
updates for the maximization step in the case of GMMs are given as
follows :
\begin{eqnarray}
\mu_i(t+1) = \frac{\sum_{j=1}^{n}w_i^{(j)}x^{(j)}}{\sum_{j=1}^{n}w_i^{(j)}}\\
  \sigma_i^2(t+1) = \frac{\sum_{j=1}^{n}w_i^{(j)} (x^{(j)}-\mu_i(t+1))^2}{\sum_{j=1}^{n}w_i^{(j)}}\\
\alpha_i(t+1)=\frac{1}{n}\sum_{j=1}^{n}w_i^{(j)} \label{eq:update}
\end{eqnarray}

\subsection{Nonlinear Transformation}
This section mainly deals with the transformation of the original
log-likelihood function into its corresponding nonlinear dynamical
system and introduces some terminology pertinent to comprehend our
algorithm. This transformation gives the correspondence between all
the critical points of the $s$-dimensional likelihood surface and
that of its dynamical system. For the case of spherical Gaussian
mixtures with $k$ components, we have the number of unknown
parameters $s=3k-1$. For convenience, the maximization problem is
transformed into a minimization problem defined by the following
objective function :
\begin{equation}
~\max_\Theta ~\{~log
~p(\mathcal{X}|\Theta)~\}=~\min_\Theta ~\{~-~log
~p(\mathcal{X}|\Theta)~\}\\= \min_\Theta f(\Theta)\label{eq:problem1} 
\end{equation}


\begin{lem1}\label{def:cont}
$f(\Theta)$ is $C^2(\Re^s,\Re)$.
\end{lem1}
\begin{proof}

Note from Eq.(\ref{eq:log}), we have
\begin{equation}\label{eq:otherlog1}
    f(\Theta)=-log ~p(\mathcal{X}|\Theta)
    =-\sum_{j=1}^n log \sum_{i=1}^k \alpha_i
    ~p(\bm{x}^{(j)}|\bm{\theta}_i)
\end{equation}
Each of the simple functions which appear in Eq. (\ref{eq:otherlog1}) are twice differentiable and continuous in the interior of the domain over which $f(\Theta)$ is defined. The function $f(\Theta)$ is composed of arithmetic operations of these simple functions and from basic results in analysis, we can conclude that $f(\Theta)$ is twice continuously differentiable.
\end{proof}

Lemma \ref{def:cont} and the preceeding arguments guarantee the existence of the gradient system associated with $f(\Theta)$ for the log-likelihood function in the case of spherical Gaussians and allows us to construct the following negative gradient system :
\begin{equation}
\begin{split}
\textstyle{ \left[ \dot{\mu}_1(t)~..~\dot{\mu}_k(t)~\dot{\sigma}_1(t) ~..~\dot{\sigma}_k(t)~\dot{\alpha}_1(t) ~..~\dot{\alpha}_{k-1}(t)\right]^T}\\ 
=~-~\left[ \frac{\partial f}{\partial
\mu_1}~..~\frac{\partial f}{\partial \mu_k}~\frac{\partial
f}{\partial \sigma_1}~..~\frac{\partial f}{\partial
\sigma_k}~\frac{\partial f}{\partial \alpha_1}~..~\frac{\partial
f}{\partial \alpha_{k-1}} \right]^T  \label{def:loggrad}
\end{split}
\end{equation}

\begin{thm}\label{th:stabgrad}{\it (Stabilitiy):}
The gradient system~\ref{def:loggrad} is completely stable.
\end{thm}
{\it Proof: See Appendix-A.}\\

Developing a gradient system is one of the simplest transformation possible. One can think of a more complicated nonlinear transformations as well. We will now describe three main guidelines that must be satisfied by the transformation :

\begin{itemize}
\item{The original log-likelihood function must be a Lyapunov function for the dynamical system.}
\item{The location of the critical points must be preserved under this transformation.}
\item{The system must be completely stable. In other words, every trajectory $\Phi(x,t)$ must be bounded.}
\end{itemize}

From the implementation point of view, it is not required to construct this gradient system. However, to understand the details of our method, it is necessary to obtain this gradient system. For simplicity, we show the construction of the gradient system for
the case of spherical Gaussians. It can be easily extended to the
full covariance Gaussian mixture case. It should be noted that only
(k-1) $\alpha$ values are considered in the gradient system because
of the unity constraint. The dependent variable $\alpha_k$ is
written as follows :

\begin{equation}
 \alpha_k=1-\sum_{j=1}^{k-1} \alpha_j\label{eq:alphak}
\end{equation}

This gradient system and the decomposition points on the practical stability boundary of the stable equilibrium points will enable us to define {\it Tier-1 stable equilibrium point}.

\begin{lem}\label{def:tier1}
For a given stable equilibrium point ($x_s$), a {\it Tier-1 stable equilibrium point} is defined as a stable equilibrium point whose stability boundary intersects with the stability boundary of $x_s$.
\end{lem}

\begin{figure}
   \centering
   \subfigure[Parameter Space]{\includegraphics[width = 2.75 in]{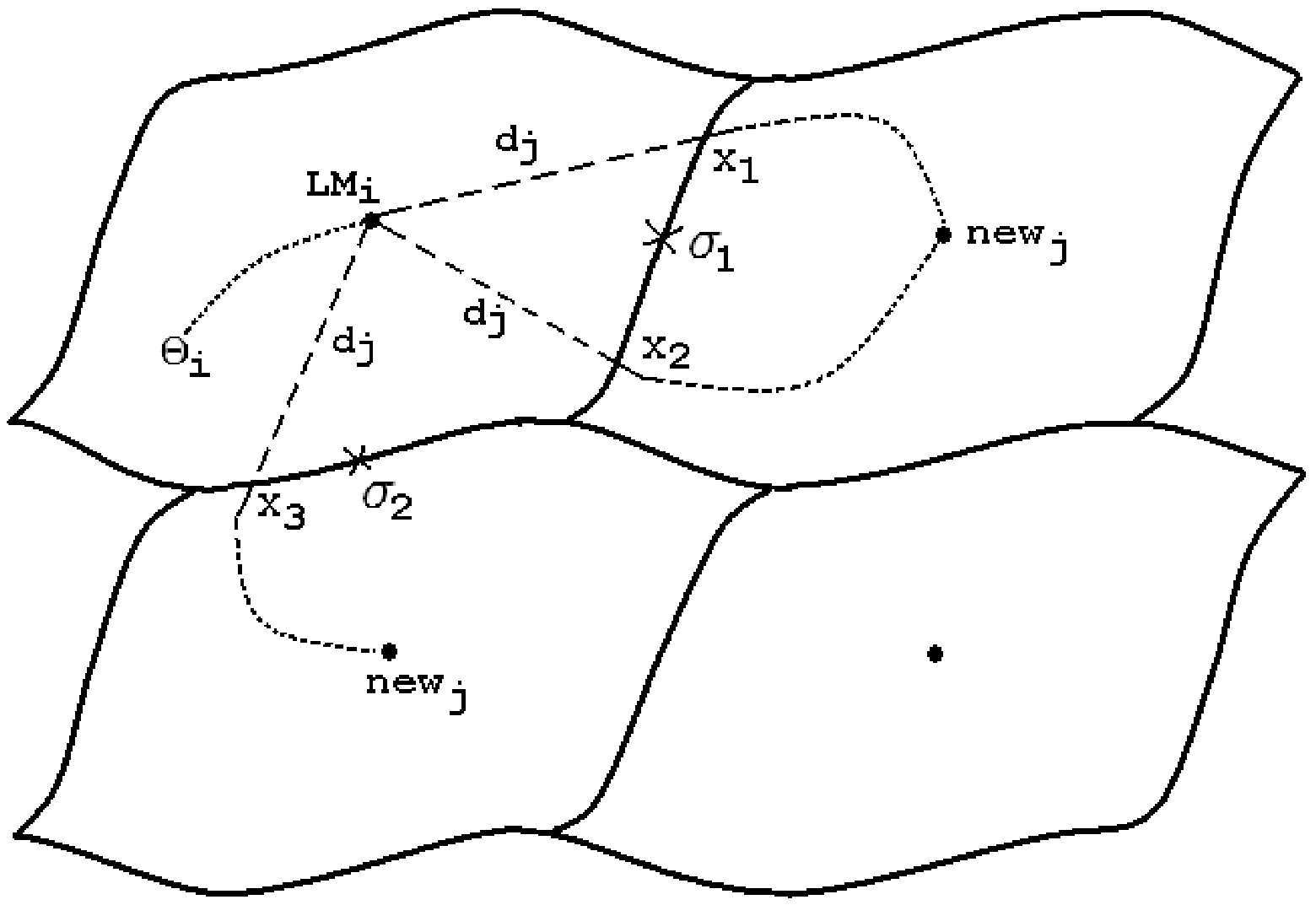}}\qquad
   \subfigure[Function Space]{\includegraphics[width = 2.75 in]{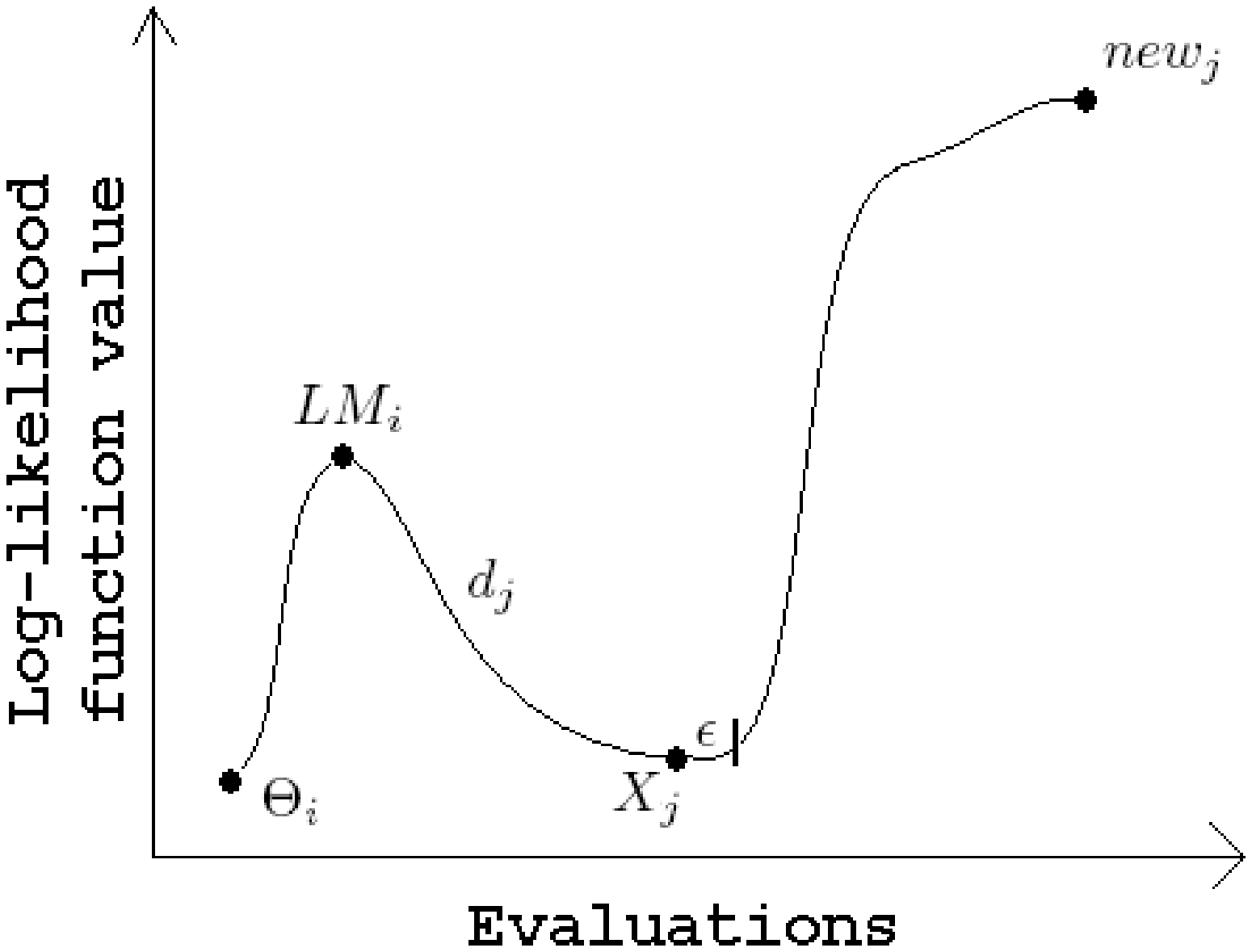}}\qquad
   \caption{\label{fig:algo_s}Various stages of our algorithm in (a) Parameter space - the solid lines indicate the practical stability boundary.
Points highlighted on the stability boundary are the decomposition
points. The dotted arrows indicate the convergence of the EM
algorithm. The dashed lines indicate the neighborhood-search stage.
$x_1$ and $x_2$ are the exit points on the practical stability
boundary (b) Different points in the function space and their corresponding log-likelihood function values.
   }
 \end{figure}

\section{TRUST-TECH based Expectation Maximization}
\label{sec:algorithm}

Our framework consists three stages namely: (i) global stage, (ii) local stage and (iii) neighborhood-search stage.
The last two stages are repeated in the solution space to obtain promising solutions. Global method obtains promising subspaces of the solution space. The next stage is the local stage (or the EM stage) where the results from the global methods are refined to the corresponding locally optimal parameter set. Then, during the neighborhood search stage, the exit points are computed
and the neighborhood solutions are systematically explored through these exit points. Fig. \ref{fig:algo_s} shows the different steps of our algorithm both in (a) the parameter space and (b) the function space.

It is beneficial to use the TRUST-TECH based algorithm at the promising subspaces. In this sense, the neighborhood-search stage can act as a interface between global methods for initialization and the EM algorithm which gives the local maxima.
This approach differs from
traditional local methods by computing multiple local maxima in
the neighborhood region. This also enhances user flexibility in choosing between different sets of good
clusterings. Though global methods can identify promising subsets, it is
important to explore this more thoroughly especially in
problems like parameter estimation.

\begin{algorithm}
\caption{TRUST-TECH based EM Algorithm} \label{nexttieralg}
\begin{algorithmic}
\STATE \textbf{Input:} Parameters $\Theta$, Data  $\mathcal{X}$,
tolerance $\tau$, Step $S_p$  \STATE \textbf{Output:}
$\widehat{\Theta}_{MLE} $  \STATE \textbf{Algorithm:}

\STATE Apply global method and store the q promising solutions $
\Theta_{init}=\{\Theta_1,\Theta_2,..,\Theta_q\}$ ~~~~~~~~ Initialize
E= $\phi$

\WHILE{$\Theta_{init} \neq \phi$} \STATE Choose $\Theta_i \in
\Theta_{init}$, set $\Theta_{init}= \Theta_{init} \backslash
\{\Theta_i\}$ \STATE $LM_i=EM(\Theta_i,\mathcal{X},\tau)$~~~~~~~~~
$E=E \cup \{LM_i\}$ \STATE Generate promising direction vectors
$d_j$ from $LM_i$

\FOR {each $d_j$} \STATE Compute Exit Point ($X_j$) along $d_j$
starting from $LM_i$ by evaluating the log-likelihood function given
by (\ref {eq:log})

\STATE $New_j=EM(X_j+\epsilon \cdot d_j,\mathcal{X},\tau)$
\IF{$new_j \notin E$} \STATE $E=E \cup New_j$ \ENDIF

\ENDFOR

   \ENDWHILE
\STATE $\widehat{\Theta}_{MLE} =max\{val(E_i)\}$

%
%
%
%
\end{algorithmic}
\end{algorithm}

In order to escape out of a found local maximum, our method needs to
compute certain promising directions based on the local behaviour of
the function. One can realize that generating these promising
directions is one of the important aspects of our algorithm.
Surprisingly, choosing random directions to move out of the local
maximum works well for this problem. One might also use other
directions like eigenvectors of the Hessian or incorporate some
domain-specific knowledge (like information about priors,
approximate location of cluster means, user preferences on the final
clusters) depending on the application that they are working on and
the level of computational expense that they can afford. We used
random directions in our work because they are very cheap to
compute. Once the promising directions are generated, exit points
are computed along these directions. {\it Exit points} are points of
intersection between any given direction and the practical stability
boundary of that local maximum along that particular direction. If
the stability boundary is not encountered along a given direction,
then there is a guarantee that one will not be able to find any new local maximum in
that direction. With a new initial guess in the vicinity of the exit
points, EM algorithm is applied again to obtain a new local maximum. Sometimes, this new point ($X_j+\epsilon \cdot d_j$) might have convergence problems. In such cases, TRUST-TECH can help the convergence by integrating the dynamical system and obtaining another point that is much closer to the local optimal solution. However, this is not done here because of the fact that the computation of gradient for log-likelihood function is expensive.

\begin{algorithm}
\caption{Params[~] $TT\_EM(Pset,Data,Tol,Step)$} \label{nexttier5}
\begin{algorithmic}
\STATE$Val=eval(Pset)$ \STATE $Dir[~]=Gen\_Dir (Pset)$ \STATE
$Eval\_MAX=500$

\FOR{$k=1$ to $size(Dir)$} \STATE $Params[k]= Pset~~~~~~ExtPt=OFF$
\STATE $Prev\_Val=Val~~~~~~~~~Cnt=0$

\WHILE{$(!~ExtPt)~ \&\& ~(Cnt<Eval\_MAX)$} \STATE
$Params[k]=update(Params[k],Dir[k],Step)$ \STATE $Cnt~=~Cnt~+~1$
\STATE $Next\_Val~=~eval(Params[k])$

\IF{$(Next\_Val~>~Prev\_Val)$} \STATE $ExtPt=ON$ \ENDIF

\STATE $Prev\_Val=Next\_Val$
    \ENDWHILE
\IF{$count<Eval\_MAX$}
    \STATE $Params[k]=update(Params[k],Dir[k],ASC)$
\STATE $Params[k]=EM(Params[k],Data,Tol)$ \ELSE \STATE
$Params[k]=NULL$ \ENDIF \ENDFOR \STATE \bf Return
$max(eval(Params[~]))$
\end{algorithmic}
\end{algorithm}

\section{Implementation Details}
\label{sec:implementation}

Our program is implemented in MATLAB and runs on Pentium IV 2.8 GHz
machine. The main procedure implemented is $TT\_EM$ described in
Algorithm ~\ref{nexttier5}. The algorithm takes the mixture data and
the initial set of parameters as input along with step size for
moving out and tolerance for convergence in the EM algorithm. It
returns the set of parameters that correspond to the Tier-1
neighboring local optimal solutions. The procedure $eval$ returns
the log-likelihood score given by Eq. (\ref{eq:log}). The $Gen\_Dir$
procedure generates promising directions from the local maximum. Exit
points are obtained along these generated directions. The procedure
$update$ moves the current parameter to the next parameter set along
a given $k^{th}$ direction $Dir[k]$. Some of the directions might
have one of the following two problems: (i) exit points might not be
obtained in these directions. (ii) even if the exit point is
obtained it might converge to a less promising solution. If the exit
points are not found along these directions, search will be
terminated after $Eval\_MAX$ number of evaluations. For all exit
points that are successfully found, $EM$ procedure is applied and
all the corresponding neighborhood set of parameters are stored in
the $Params[~]$. To ensure that the new initial points are
in a new convergence region of the EM algorithm, one should move (along that particular direction) `$\epsilon$' away from the exit points. Since,
different parameters will be of different ranges, care must be taken
while multiplying with the step sizes. It is important to use the
current estimates to get an approximation of the step size with
which one should move out along each parameter in the search space.
Finally, the solution with the highest likelihood score amongst the
original set of parameters and the Tier-1 solutions is returned.

\begin{figure*}[htp]
   \centering
   \subfigure[]{\includegraphics[width = 2.45 in]{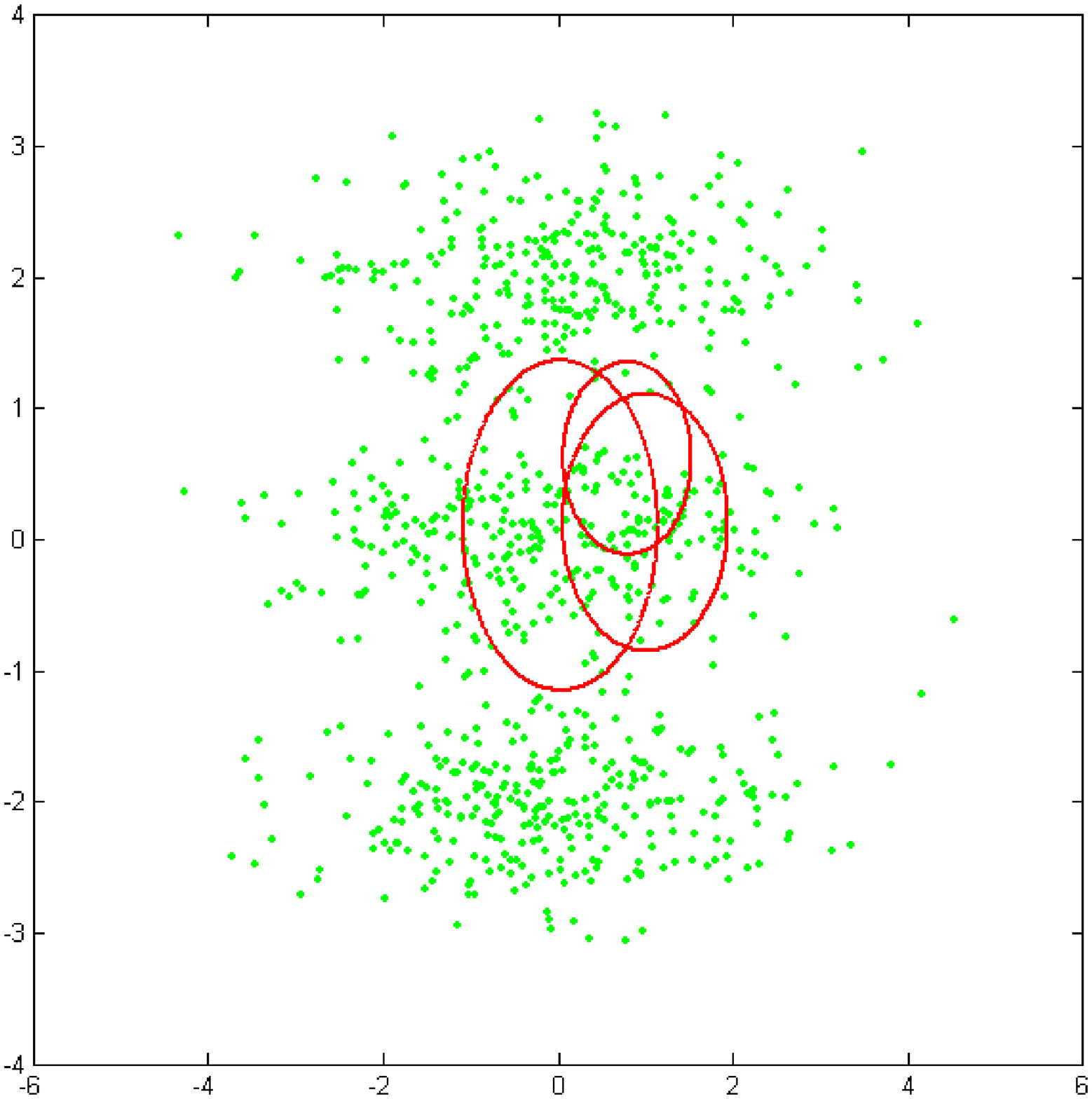}}\qquad
   \subfigure[]{\includegraphics[width = 2.45 in]{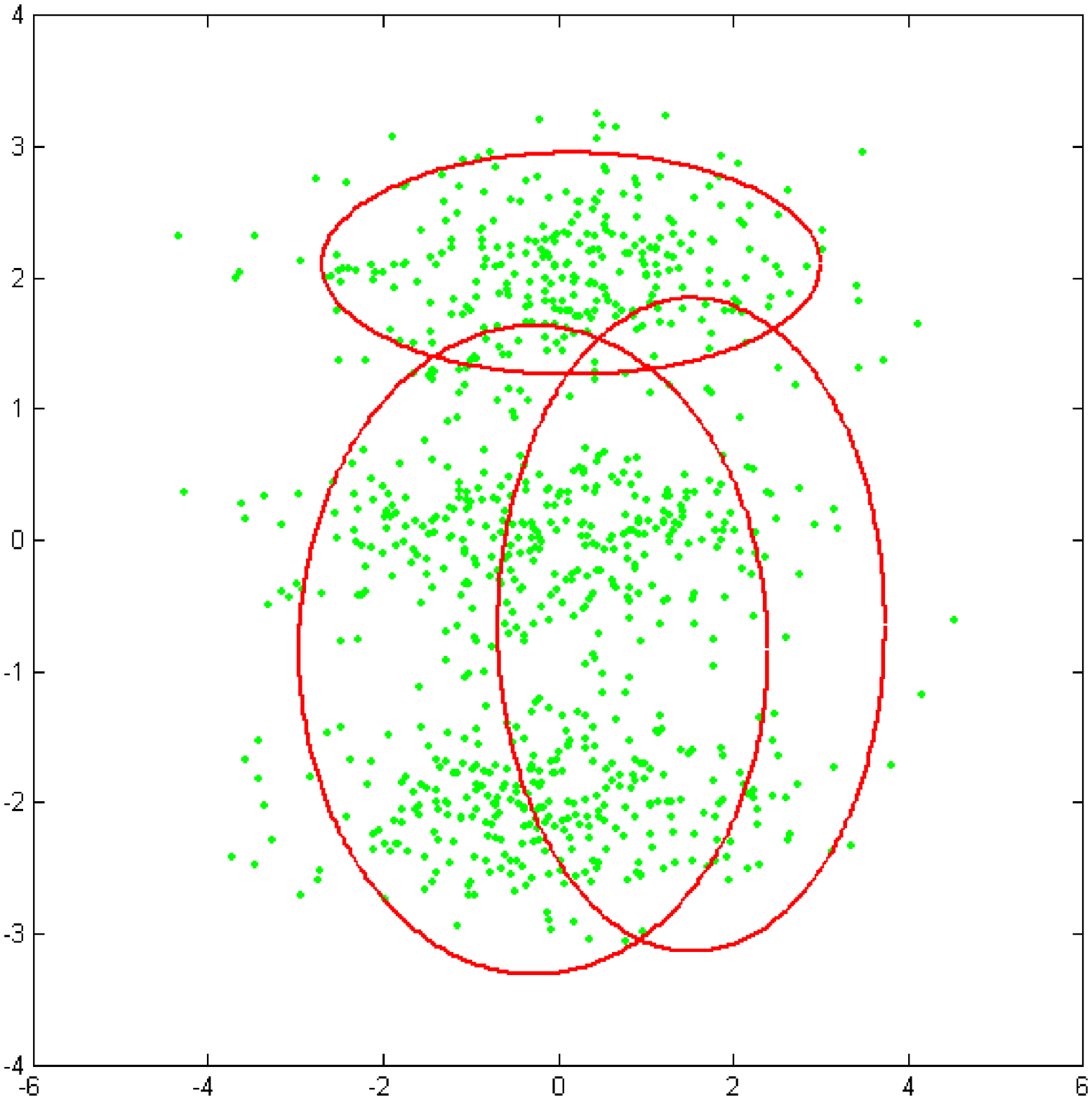}}\qquad
   \subfigure[]{\includegraphics[width = 2.45 in]{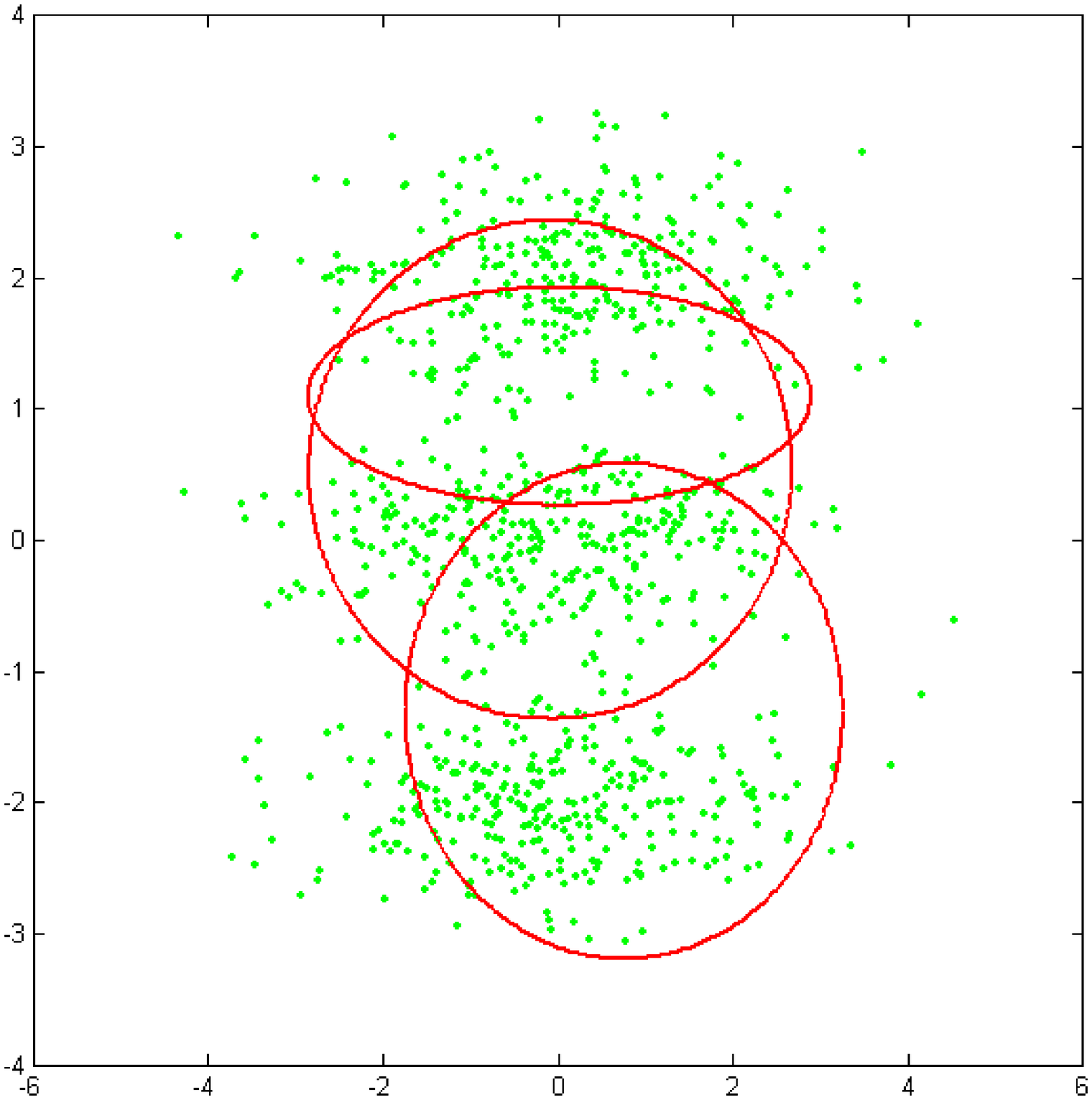}}\qquad
   \subfigure[]{\includegraphics[width = 2.45 in]{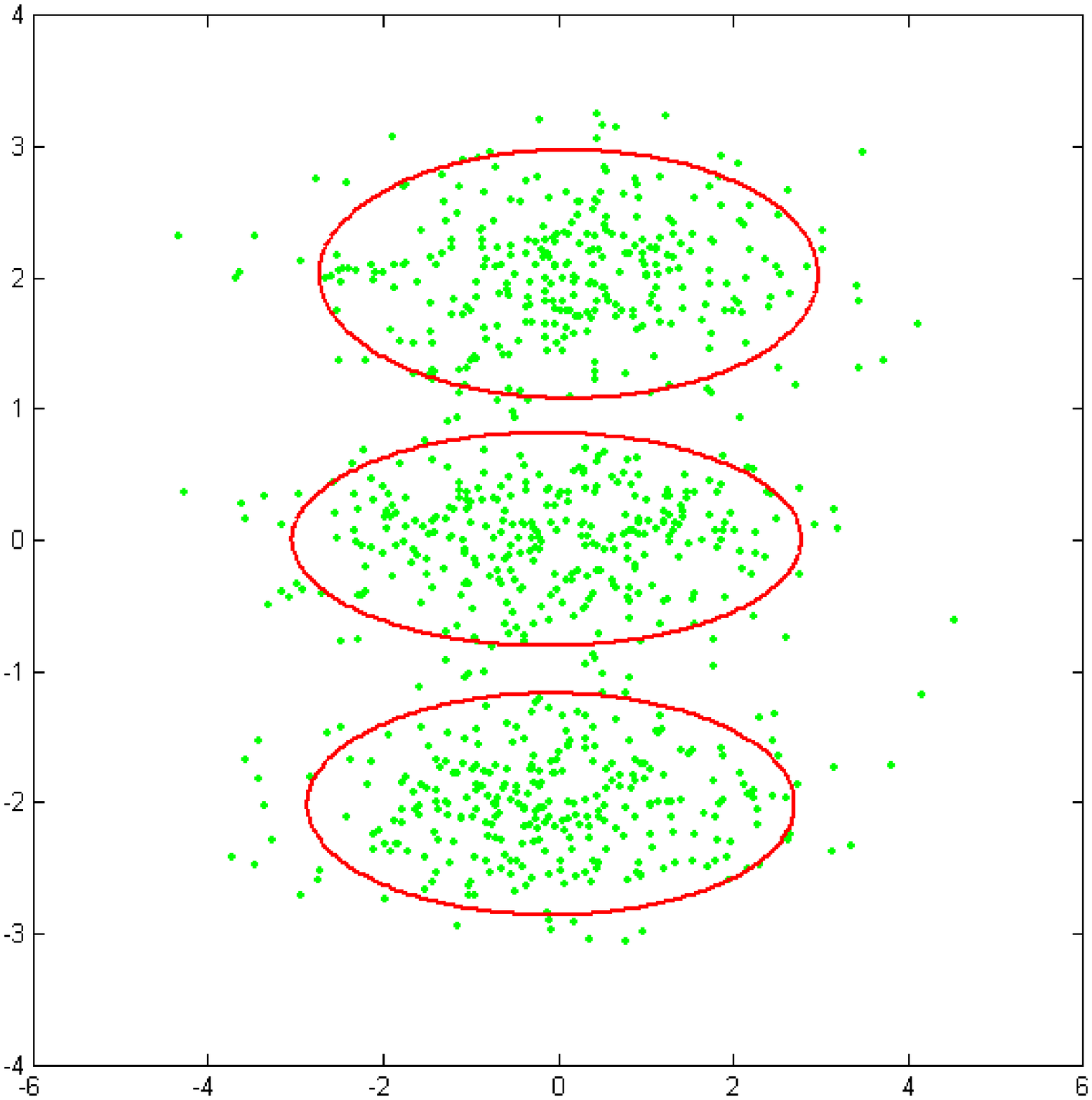}}\qquad
   \caption{\label{fig:diagcov}Parameter estimates at various
   stages of our algorithm on the three component Gaussian mixture
   model (a) Poor random initial guess (b) Local maximum
   obtained after applying EM algorithm with the poor initial
   guess (c) Exit point obtained by our algorithm (d) The final
   solution obtained by applying the EM algorithm using the exit point as the initial guess.
   }
 \end{figure*}
\section{Results and Discussion}
\label{sec:results}

Our algorithm has been tested on both synthetic and real datasets.
The initial values for the centers and the covariances were chosen
uniformly random. Uniform priors were chosen for initializing the
components. For real datasets, the centers were chosen randomly from
the sample points.

\begin{figure}[htp]
\centerline{
  \epsfig{figure=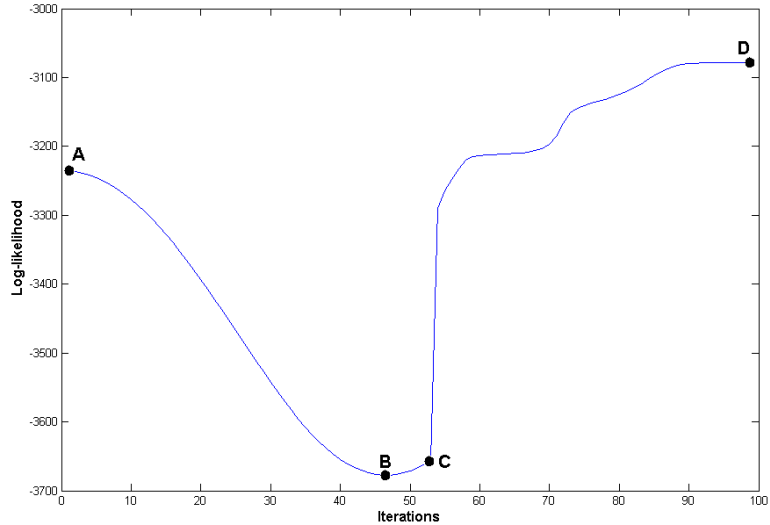, width=4.0in}
} \caption{Graph showing likelihood vs Evaluations. A corresponds to
the original local maximum (L=-3235.0). B corresponds to the exit
point (L=-3676.1). C corresponds to the new initial point (L=-3657.3) after moving out by
`$\epsilon$'. D corresponds to the new local maximum (L=-3078.7).}
\label{fig:eval}
\end{figure}

\subsection{Synthetic Datasets}
A simple synthetic data with 40 samples and 5 spherical Gaussian
components was generated and tested with our algorithm. Priors were
uniform and the standard deviation was 0.01. The centers for the
five components are given as follows: $\mu_1=[0.3~0.3]^T$,
$\mu_2=[0.5~0.5]^T$, $\mu_3=[0.7~0.7]^T$, $\mu_4=[0.3~0.7]^T$ and
$\mu_5=[0.7~0.3]^T$.



The second dataset was that of a diagonal covariance case containing
$n=900$ data points. The data generated from a two-dimensional,
three-component Gaussian mixture distribution with mean vectors at
$[0 ~-2]^T, [0~ 0]^T,[0 ~2]^T$ and same diagonal covariance matrix
with values 2 and 0.2 along the diagonal \cite{Ueda98}. All the
three mixtures have uniform priors. Fig. \ref{fig:diagcov} shows
various stages of our algorithm and demonstrates how the clusters
obtained from existing algorithms are improved using our algorithm.
The initial clusters obtained are of low quality because of the poor
initial set of parameters. Our algorithm takes these clusters and
applies the neighborhood-search stage and the EM stage simultaneously to
obtain the final result. Fig. \ref{fig:eval} shows the value of the
log-likelihood during the neighborhood-search stage and the EM
iterations.

In the third synthetic dataset, a more complicated overlapping
Gaussian mixtures are considered \cite{Figueiredo02}. The parameters
are as follows: $\mu_1=\mu_2=[-4~ -4]^T$ , $\mu_3 =[2~2]^T$ and
$\mu_4=[-1~-6]^T$. $\alpha_1=\alpha_2=\alpha_3=0.3$ and
$\alpha_4=0.1$.
\begin{displaymath}
     \Sigma_1=\left[ \begin{array}{cc} 1 &0.5\\  0.5 & 1 \end{array}
     \right]~~~~~~~~~~~~~~~\Sigma_2=\left[ \begin{array}{cc} 6 &-2\\  -2 & 6 \end{array}
     \right]
 \end{displaymath}
\begin{displaymath}
    \Sigma_3=\left[ \begin{array}{cc} 2 &-1\\  -1 & 2 \end{array}
     \right]~~~~~~~~~~~~~~~ \Sigma_4=\left[ \begin{array}{cc} 0.125 &0\\  0 &0.125  \end{array}
     \right]
 \end{displaymath}


\begin{table*}[htp]
\centering \caption{\protect Performance of TRUST-TECH-EM algorithm on
an average of 100 runs on various synthetic and real datasets compared with random start EM algorithm}
\begin{center}
\begin{small}
\begin{tabular}{|c|c|c|c|c|c|}
\hline
Dataset &Samples & Clusters & Features &  EM(mean $\pm$ std) & TT-EM(mean $\pm$ std)\\
\hline Spherical & 40& 5 & 2  & 38.07$\pm$2.12 & 43.55$\pm$0.6 \\
\hline Elliptical & 900& 3 & 2  &  -3235$\pm$0.34 & -3078.7$\pm$0.03 \\
\hline FC1& 500& 4 & 2  & -2345.5 $\pm$175.13 & -2121.9$\pm$ 21.16 \\
\hline FC2& 2000& 4 & 2  & -9309.9 $\pm$694.74 & -8609.7 $\pm$37.02 \\
\hline Iris & 150& 3 & 4  & -198.13$\pm$27.25 &-173.63$\pm$11.72\\
\hline Wine & 178& 3 & 13  & -1652.7$\pm$1342.1& -1618.3$\pm$1349.9\\
\hline
\end{tabular}
\end{small}
\end{center}
\label{TB:results5}
\end{table*}

\subsection{Real Datasets}
Two real datasets obtained from the UCI Machine Learning repository
\cite{Blake98} were also used for testing the performance of our
algorithm. Most widely used Iris data with 150 samples, 3 classes
and 4 features was used. Wine data set with 178 samples was also
used for testing. Wine data had 3 classes and 13 features. For these
real data sets, the class labels were deleted thus treating it as an unsupervised learning problem. Table~\ref{TB:results5} summarizes our
results over 100 runs. The mean and the standard deviations of the
log-likelihood values are reported. The traditional EM algorithm
with random starts is compared against our algorithm on both
synthetic and real data sets. Our algorithm not only obtains higher
likelihood value but also produces it with high confidence. The low
standard deviation of our results indicates the robustness of
obtaining the global maximum. In the case of the wine data, the
improvements with our algorithm are not much significant compared to
the other datasets. This might be due to the fact that the dataset
might not have Gaussian components. Our method assumes that the
underlying distribution of the data is mixture of Gaussians.
Table~\ref{TB:compresults5} gives the results of TRUST-TECH-EM
compared with other methods like split and merge EM and k-means+EM
proposed in the literature.




\begin{table}[htp]
\centering \caption{\protect Comparison of TRUST-TECH-EM with
other methods}
\begin{center}
\begin{tabular}{|c|c|c|}
\hline
Method & Elliptical & Iris\\
\hline RS+EM & -3235 $\pm$ 14.2 & -198 $\pm$ 27\\
\hline K-Means+EM & -3195 $\pm$ 54&-186 $\pm$ 10\\
\hline SMEM &-3123 $\pm$ 54&-178.5 $\pm$ 6\\
\hline TRUST-TECH-EM &-3079 $\pm$ 0.03 &-173.6 $\pm$ 11\\
\hline
\end{tabular}
\end{center}
\label{TB:compresults5}
\end{table}
\subsection{Discussion}
It will be effective to use TRUST-TECH-EM for those solutions that
appear to be promising. Due to the nature of the problem, it is very
likely that the nearby solutions surrounding the existing solution
will be more promising. One of the primary advantages of our method
is that it can be used along with other popular methods available
and improve the quality of the existing solutions. In clustering
problems, it is an added advantage to perform refinement of the
final clusters obtained. Most of the focus in the literature was on
new methods for initialization or new clustering techniques which
often do not take advantage of the existing results and completely
start the clustering procedure ``{\it from scratch}". Though shown
only for the case of multivariate Gaussian mixtures, our technique
can be effectively applied to any parametric finite mixture model.


\begin{table}[htp]
\centering \caption{\protect Number of iterations taken for
the convergence of the best solution. }
\begin{center}
\begin{tabular}{|c|c|c|}
\hline
Dataset & Avg. no. of  & No. of iterations \\
 &iterations & for the best solution\\
\hline Spherical &126& 73\\
\hline Elliptical &174& 86\\
\hline Full covariance &292&173\\
\hline
\end{tabular}
\end{center}
\label{TB:convresults}
\end{table}
Table \ref{TB:convresults} summarizes the average number of
iterations taken by the EM algorithm for the convergence to the
local optimal solution. We can see that the most promising solution
produced by our TRUST-TECH methodology converges much faster. In
other words, our method can effectively take advantage of the fact
that the convergence of the EM algorithm is much faster for high
quality solutions. We exploit this inherent property of the EM algorithm
to improve the efficiency of our algorithm. Hence, for obtaining the Tier-1 solutions using our algorithm, the
threshold for the number of iterations can be significantly lowered.

\newpage
\section*{APPENDIX-A: Proof of Theorem \ref{th:stabgrad}} \label{sec:appendix-A}
\begin{proof}
First, we will show that every bounded trajectory will converge to one of the equilibrium points. Second, we will show that every trajectory is bounded \cite{Chiang96}.
\begin{enumerate}
\item{Let $\Phi(x,t)$ denote the bounded trajectory starting at $x$. Computing the time derivative along the trajectory, we get
\begin{equation*}
\frac{d}{dt}f(\Phi(x,t))=-(\nabla f(\Phi(x,t)))^T(\nabla f(\Phi(x,t)))~\leq~0
\end{equation*}
Also, we know that $\frac{d}{dt}f(\Phi(x,t))~=~0$ if, and only if, $x \in E$. Hence, $f(x)$ is a Lyapunov function of the gradient system (\ref{def:loggrad}) and the $\omega$-limit point of any bounded trajectory consists of equilibrium points only, i.e. any bounded trajectory will approach one of the equilibrium point.}
\item{
Following the proof of preposition 1 presented in \cite{Chiang96}, we can show that every trajectory $\Phi(x,t)$ is bounded. However, we will have to show that the magnitude of the gradient of the log-likelihood function for the Gaussian mixture model is bounded on the entire domain of the parameter space.

\begin{equation}\label{eq:logeq}
log ~p(\mathcal{Y}|\Theta) =-\sum_{j=1}^n log \sum_{i=1}^k \alpha_i ~p({y}^{(j)}|{\theta}_i)
\end{equation}

Now, the domain of the parameter space is given as follows: 

$-\infty <\mu_i<\infty$ , $\Sigma_i$ is positive definite and $0\leq \alpha_i \leq 1$ where $\sum_{i=1}^{k}\alpha_i=1$. 

First, let us focus on $\alpha$ because it is a constrained variable.

\textbf {Derivative with $\alpha$} :

\begin{equation}\label{eq:partalpha}
    \frac{\partial f}{\partial \alpha_r}= \sum_{j=1}^{n}\left[ \frac{p({y}^{(j)}|{\theta}_r)}{\sum_{i=1}^{k}\alpha_i p({y}^{(j)}|{\theta}_i)}\right]
\end{equation}
As $\alpha \rightarrow 1$, we have
\begin{equation*}\label{eq:alphatends1}
    \frac{\partial f}{\partial \alpha_r}= \sum_{j=1}^{n}\left[ \frac{p({y}^{(j)}|{\theta}_r)}{1 \cdot p({y}^{(j)}|{\theta}_i)}\right] = n <\infty
\end{equation*}
As $\alpha \rightarrow 0$, we have
\begin{equation*}\label{eq:alphatends0}
    \frac{\partial f}{\partial \alpha_r}=\sum_{j=1}^{n}\left[ \frac{p({y}^{(j)}|{\theta}_r)}{\sum_{i=1,i\neq r}^{k}\alpha_i p({y}^{(j)}|{\theta}_i)}\right] <\infty
\end{equation*}

Hence, the derivatives with respect to $\alpha$ are bounded. 

\textbf {Derivative with $\mu$} :
\begin{equation}\label{eq:partmu}
    \frac{\partial f}{\partial \mu_r}= \sum_{j=1}^{n}\left[ \frac{\alpha_r \frac{1}{\sqrt{(2\pi)}\sigma_r} e^{-\frac{(x^{(j)}-\mu_r)^2}{2\sigma_r^2}}\cdot\frac{1}{\sigma_r^2}(x^{(j)}-\mu_r)}{\sum_{i=1}^{k}\alpha_i p({y}^{(j)}|{\theta}_i)}\right]
\end{equation}
This is obviously bounded for and $\mu \in \Re$.

\textbf {Derivative with $\sigma$} :
\begin{equation}\label{eq:partmu}
    \frac{\partial f}{\partial \sigma_r}= \sum_{j=1}^{n}\left[ \frac{\frac{1}{\sigma_r} e^{-\frac{(x^{(j)}-\mu_r)^2}{2\sigma_r^2}}\cdot \frac{(x^{(j)}-\mu_r)^2}{\sigma_{r^3}} -\frac{1}{\sigma_{r^2}} e^{-\frac{(x^{(j)}-\mu_r)^2}{2\sigma_r^2}} }{\sum_{i=1}^{k}\alpha_i p({y}^{(j)}|{\theta}_i)}\right]
\end{equation}

As $\sigma_r \rightarrow 0$ the exponential factor goes to zero faster than $\frac{1}{\sigma_r}$ goes to infinity. Hence, it is bounded. So, the gradient of the log-likelihood function is bounded in the  entire domain of the parameter space.

}
\end{enumerate}

\end{proof}

\chapter{Motif Refinement using Neighborhood Profile Search}
\label{ch:motif}

As a case study of the algorithm developed in the previous chapter, we describe its application to the motif finding problem in bioinformatics. The main goal of the motif finding problem is to detect novel,
over-represented unknown signals in a set of sequences (e.g.
transcription factor binding sites in a genome). The most widely
used algorithms for finding motifs obtain a generative probabilistic
representation of these over-represented signals and try to discover
profiles that maximize the information content score. Although these
profiles form a very powerful representation of the signals, the
major difficulty arises from the fact that the best motif
corresponds to the global maximum of a non-convex continuous
function. Popular algorithms like Expectation Maximization (EM) and
Gibbs sampling tend to be very sensitive to the initial guesses and
are known to converge to the nearest local maximum very quickly. In
order to improve the quality of the results, EM is used with
multiple random starts or any other powerful stochastic global
methods that might yield promising initial guesses (like projection
algorithms). Global methods usually do not give initial guesses
in the convergence region of the best local maximum but rather give some point that is in a promising region. In
this chapter, we apply the TRUST-TECH based Expectation Maximization (TT-EM) algorithm proposed in the previous chapter to this motif finding problem. It has the capability to search for alignments corresponding to Tier-1 local optimal solutions in the profile space. This search is performed in a systematic manner to explore the multiple local optimal solutions.
This effective search is achieved by transforming the original
optimization problem into a dynamical system with certain properties and obtaining more useful information about the nonlinear likelihood surface via the dynamical and topological properties of the dynamic system.

\begin{figure}[htp]
\centerline{
  \epsfig{figure=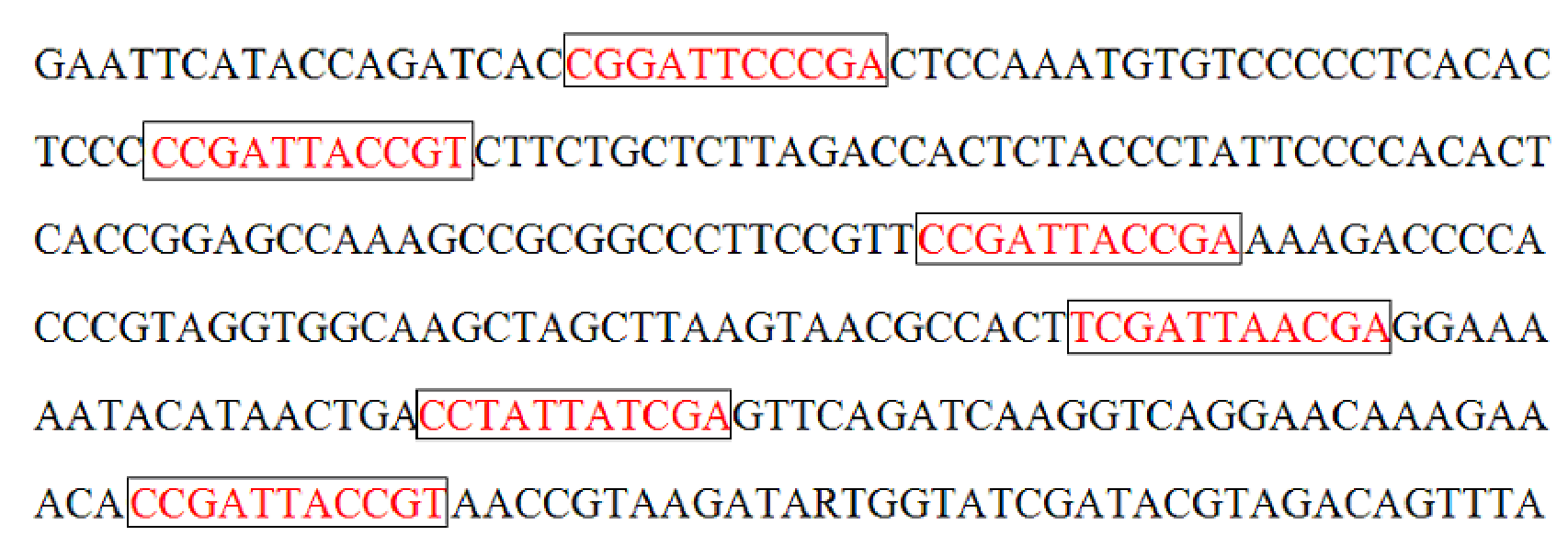, width=4.5in}
} \caption
{Synthetic DNA sequences containing some instance of the
pattern `CCGATTACCGA' with a maximum number of 2 mutations. The
motifs in each sequence are highlighted in the box. We have a (11,2)
motif where 11 is the length of the motif and 2 is the number of
mutations allowed.}
\label{fig:sequences}
\end{figure}

Recent developments in DNA sequencing have allowed biologists to
obtain complete genomes for several species. However, knowledge of
the sequence does not imply the understanding of how genes interact
and regulate one another within the genome. Many transcription
factor binding sites are highly conserved throughout the sequences
and the discovery of the location of such binding sites plays an
important role in understanding gene interaction and gene
regulation. Although there are several variations of the motif finding
algorithms, the problem studied in this chapter is defined as
follows: without any previous knowledge of the consensus pattern,
discover all the occurrences of the motifs and then recover a pattern
for which all of these instances are within a given number of
mutations (or substitutions). Despite the significant amount of
literature available on the motif finding problem, many do not
exploit the probabilistic models used for motif refinement
\cite{Lawrence93,Bailey94}. In this chapter, we consider a precise version of the motif discovery problem in
computational biology as discussed in \cite{Buhler01,Pevzner00}. The
planted (l,d) motif problem \cite{Pevzner00} considered here is described as follows: Suppose there is a fixed but unknown
nucleotide sequence $M$ (the {\it motif}) of length $l$. The problem
is to determine $M$, given $t$ sequences with $t_i$ being the length
of the $i^{th}$ sequence and each one containing a planted variant of
$M$. More precisely, each such planted variant is a substring that
is $M$ with exactly $d$ point substitutions (see
Fig.~\ref{fig:sequences}). More details about the complexity of the
motif finding problem is given in \cite{Pevzner00a}. A detailed
assessment of different motif finding algorithms was published
recently in \cite{Tompa05}.



\section{Relevant Background}
\label{sec:background} Existing approaches used to solve the motif
finding problem can be classified into two main categories
\cite{Eskin04}. The first group of algorithms utilizes a generative
probabilistic representation of the nucleotide positions to discover
a consensus DNA pattern that maximizes the information content
score. In this approach, the original problem of finding the best
consensus pattern is formulated as finding the global maximum of a
continuous non-convex function. The main advantage of this approach
is that the generated profiles are highly representative of the
signals being determined \cite{Durbin99}. The disadvantage, however,
is that the determination of the ``best" motif cannot be guaranteed
and is often a very difficult problem since finding global maximum
of any continuous non-convex function is a challenging task.
Current algorithms converge to the nearest local optimum instead of
the global solution. Gibbs sampling \cite{Lawrence93}, MEME
\cite{Bailey94}, greedy CONSENSUS algorithm \cite{Hertz99} and HMM
based methods \cite{Eddy98} belong to this category.

The second group uses patterns with `mismatch representation' which
defines a signal to be a consensus pattern and allows up to a certain
number of mismatches to occur in each instance of the pattern. The
goal of these algorithms is to recover the consensus pattern with
the highest number of instances. These methods view the
representation of the signals as discrete and the main advantage of
these algorithms is that they can guarantee that the highest scoring
pattern will be the global optimum for any scoring function.  The
disadvantage, however, is that consensus patterns are not as
expressive of the DNA signal as profile representations. Recent
approaches within this framework include Projection methods
\cite{Buhler01,Raphael04}, string based methods \cite{Pevzner00},
Pattern-Branching \cite{Price03}, MULTIPROFILER \cite{Keich02} and
other branch and bound approaches \cite{Eskin02,Eskin04}.

A hybrid approach could potentially combine the expressiveness of
the profile representation with convergence guarantees of the
consensus pattern. An example of a hybrid approach is the Random
Projection \cite{Buhler01} algorithm followed by EM algorithm
\cite{Bailey94}. It uses a global solver to obtain promising
alignments in the discrete pattern space followed by further local
solver refinements in continuous space \cite{Barash01,Segal02}.
Currently, only few algorithms take advantage of a combined discrete
and continuous space search \cite{Buhler01,Eskin04,Raphael04}. We consider the profile representation of the motif
and a new hybrid algorithm is developed to escape out of the local
maxima of the likelihood surface. Some motivations to develop the new hybrid algorithm are :
\begin {itemize}
\item{A motif refinement stage is vital and popularly used by many
pattern based algorithms (like PROJECTION, MITRA etc) which try to
find optimal motifs. } \item{The traditional EM algorithm used in
the context of motif finding converges very quickly to the nearest
local optimal solution (within 5-8 iterations) \cite{Blekas03}.} \item{There are
many other promising local optimal solutions in the close vicinity
of the profiles obtained from the global methods.}
\end{itemize}

In spite of the importance placed on obtaining a global optimal
solution in the context of motif finding, little work has been done
in the direction of finding such solutions \cite{Xing04}. There are
several proposed methods to escape out of the local optimal solution
to find better solutions in machine learning \cite{Elidan02} and
optimization \cite{Cetin93} related problems. Most of them are
stochastic in nature and usually rely on perturbing either the data
or the hypothesis. These stochastic perturbation algorithms are
inefficient because they will sometimes miss a neighborhood solution
or obtain an already existing solution. To avoid these problems, we
will apply TRUST-TECH-EM algorithm that can systematically explore multiple local maxima in a tier-by-tier manner. Our method is primarily based on transforming the log-likelihood function into its corresponding dynamical system and obtaining multiple local optimal solutions. The underlying theoretical details of TRUST-TECH is described in \cite{Chiang96, Lee04}.

\section{Preliminaries}
\label{sec:problem} We will first describe our problem formulation
and the details of the EM algorithm in the context of motif finding
problem. We will then describe some details of the dynamical system
of the log-likelihood function which enables us to search for the
nearby local optimal solutions.

\subsection{Problem Formulation}

In this section, we transform the the problem of finding the best possible motif into a problem of finding the global maximum of a highly nonlinear log-likelihood scoring function obtained from its profile representation. The log-likelihood surface is made of $3l$ variables which are treated as the unknown parameters that are to be estimated. We will now describe these parameters ($Q_{k,j}$) and then represent the scoring function in terms of these parameters.

Let $t$ be the total number of sequences and $n$ be the average
length of the sequences. Let $S=\{S_1,S_2...S_t\}$ be the set of $t$
sequences. Let $P$ be a single alignment containing the set of
segments $\{P_1,P_2,...,P_t\}$. $l$ is the length of the consensus
pattern. For further discussion, we use
the following variables

 \begin{tabbing}
\hspace{0.5in}$i~=~1~...~t~~~~~~~~~~~~--$ for $t$ sequences \\
\hspace{0.5in}$k~=~1~...~l~~~~~~~~~~~~--$ for positions within an $l$-mer\\
\hspace{0.5in}$j~\in~\{A,T,G,C\}~~~~--$ for each nucleotide
 \end{tabbing}

\begin{table}[h]
\centering \caption{\protect A count of nucleotides
${A,T,G,C}$ at each position $k={1..l}$ in all the sequences of the
data set. $k=0$ denotes the background count.}

\begin{center}
\begin{tabular}{|c|c|c|c|c|c|c|c|}
\hline $j$ &  $k=0$& $k=1$&$ k=2$&$ k=3$ &$k=4$& ~...~ &  $k=l$\\
\hline

${A}$ & $C_{0,1}$ & $C_{1,1}$ &$C_{2,1}$ &$C_{3,1}$ &$C_{4,1}$ &...
&$C_{l,1}$\\ \hline

${T} $& $C_{0,2}$ & $C_{1,2}$ &$C_{2,2}$ &$C_{3,2}$ &$C_{4,2}$ &...
&$C_{l,2}$\\ \hline

${G}$ & $C_{0,3}$ & $C_{1,3}$ &$C_{2,3} $&$C_{3,3}$ &$C_{4,3}$ &...
&$C_{l,3}$\\ \hline

${C}$ & $C_{0,4}$ & $C_{1,4}$ &$C_{2,4} $&$C_{3,4}$ &$C_{4,4}$ &...
&$C_{l,4}$\\ \hline

\end{tabular}
\end{center}
\label{TB:PSSM}
\end{table}

The count matrix can be constructed from the given alignments as
shown in Table ~\ref{TB:PSSM}. We define $C_{0,j}$ to be the
non-position specific background count of each nucleotide in all of
the sequences where $j \in \{ A,T,C,G \}$ is the running total of
nucleotides occurring in each of the $l$ positions. Similarly,
$C_{k,j}$ is the count of each nucleotide in the $k^{th}$ position
(of the $l-mer$) in all the segments in $P$.

\begin{equation}\label{eq:backcount}
    Q_{0,j}=\frac{C_{0,j}}{\sum_{J\in \{A,T,G,C\}}{C_{0,J}}}
\end{equation}

\begin{equation}\label{eq:forecount}
    Q_{k,j}=\frac{C_{k,j}+b_j}{t+\sum_{J\in \{A,T,G,C\}}{b_J}}
\end{equation}

Eq. (\ref{eq:backcount}) shows the background frequency of each
nucleotide where $b_j$ is known as the Laplacian or Bayesian
correction and is equal to $d*Q_{0,j}$ where $d$ is some constant
usually set to unity.  Eq. (\ref{eq:forecount}) gives the weight
assigned to the type of nucleotide at the $k^{th}$ position of the
motif.

A Position Specific Scoring Matrix (PSSM) can be constructed from
one set of instances in a given set of $t$ sequences. From Eqs. (\ref{eq:backcount}) and (\ref{eq:forecount}), it is obvious that
the following relationship holds:

\begin{equation}\label{eq:sum}
    \sum_{j \in \{A,T,G,C\}}Q_{k,j}=1 ~~~~~~~~~~\forall k=0,1,2,...l
\end{equation}

For a given $k$ value in Eq. (\ref{eq:sum}), each $Q$ can be represented
in terms of the other three variables. Since the length of the motif
is $l$, the final objective function (i.e. the information content
score) would contain $3l$ independent variables\footnote {Although,
there are $4l$ variables in total, because of the constraints
obtained from Eq. (\ref{eq:sum}), the parameter space will contain only
$3l$ independent variables. Thus, the constraints help in reducing
the dimensionality of the search problem.}.

To obtain the score, every possible $l-mer$ in each of the $t$
sequences must be examined.  This is done so by multiplying the
respective $Q_{i,j} / Q_{0,j}$ dictated by the nucleotides and their
respective positions within the $l-mer$. Only the highest scoring
$l-mer$ in each sequence is noted and kept as part of the alignment.
The total score is the sum of all the best scores in each sequence.

\begin{equation}\label{eq:final}
    A(Q) =\sum_{i=1}^t log(A)_i=\sum_{i=1}^t
    log\left(\prod_{k=1}^{l}\frac{Q_{k,j}}{Q_b}\right)_i\\
    =\sum_{i=1}^t \sum_{k=1}^l log(Q_{k,j}^{'})_i
\end{equation}

$Q_{k,j}^{'}$ is the ratio of the nucleotide probability to the
corresponding background probability, i.e. $Q_{k,j} / Q_b$.
$Log(A)_i$ is the score at each individual $i^{th}$ sequence where t
is the total number of sequences.  In Eq. (\ref{eq:final}), we
see that $A$ is composed of the product of the weights for each
individual position $k$. We consider this to be the Information
Content (IC) score which we would like to maximize. $A(Q)$ is the
non-convex $3l$ dimensional continuous function for which the global
maximum corresponds to the best possible motif in the dataset. In summary, we transform the problem of finding the optimal motif intoa problem of finding the global maximum of a non-convex continuous $3l$ dimensional function.


\subsection{Dynamical Systems for the Scoring Function}
In order to present our algorithm, we define the dynamical
system corresponding to the log-likelihood function and the PSSM.
The key contribution here is the development of this
nonlinear dynamical system which will enable us to realize the
geometric and dynamic nature of the likelihood surface. We construct
the following {\it gradient system} in order to locate critical
points of the objective function (\ref{eq:final}):

\begin{equation}
 \dot{Q}(t)=-\nabla A(Q)\label{eq:gradientsystem1}
\end{equation}

One can realize that this transformation preserves all of the
critical points \cite{Chiang96}. Now, we will describe the
construction of the gradient system and the Hessian in detail. In
order to reduce the dominance of one variable over the other, the
values of the each of the nucleotides that belong to the consensus
pattern at the position $k$ will be represented in terms of the
other three nucleotides in that particular column. Let $P_{ik}$
denote the $k^{th}$ position in the segment $P_i$. This will also
minimize the dominance of the eigenvector directions when the
Hessian is obtained. The variables in the scoring function are
transformed into new variables described in Table ~\ref{TB:hessian}.

\begin{equation}\label{eq:adsaq}
    A(Q)=\sum_{i=1}^t \sum_{k=1}^l log ~f_{ik}(w_{3k-2},w_{3k-1},w_{3k})_i
\end{equation}

where $f_{ik}$ can take the values \{$w_{3k-2}$, $w_{3k-1}$,
$w_{3k}$, $1-(w_{3k-2}+w_{3k-1}+w_{3k})$ \} depending on the $P_{ik}$ value.\\

\begin{table}[h]
\centering \caption{\protect A count of nucleotides $j \in
\{A,T,G,C\}$ at each position $k={1..l}$ in all the sequences of the
data set. $C_k$ is the $k^{th}$ nucleotide of the consensus pattern
which represents the nucleotide with the highest value in that
column. Let the consensus pattern be GACT...G and $b_j$ be the
background. }

\begin{center}
\begin{tabular}{|c|c|c|c|c|c|c|c|}
\hline $j$ &  $k=b$& $k=1$&$ k=2$&$ K=3$ &$k=4$&~...~&  $k=l$\\
\hline

${A}$ & $b_A$ & $w_1$ &$C_2$ &$w_7$ &$w_{10}$ &... &$w_{3l-2}$\\
\hline

${T} $& $b_T$ & $w_2$ &$w_4$ &$w_8$ &$C_4$ &... &$w_{3l-1}$\\
\hline

${G}$ & $b_G$ & $C_1$ &$w_5 $&$w_9$ &$w_{11}$ &... &$C_l$\\
\hline

${C}$ & $b_C$ & $w_3$ &$w_6 $&$C_3$ &$w_{12}$ &... &$w_{3l}$\\
\hline

\end{tabular}
\end{center}
\label{TB:hessian}
\end{table}

The first derivative of the scoring function is a one dimensional
vector with $3l$ elements.
\begin{equation}\label{eq:beq}
    \nabla A= \left[~ \frac{\partial A}{\partial w_1} ~~ \frac{\partial A}{\partial w_2} ~~\frac{\partial A}{\partial w_3}
    ~~.~.~.~.~~~ \frac{\partial A}{\partial w_{3l}}
    ~\right]^T
\end{equation}

and each partial derivative is given by
\begin{equation}\label{eq:peq}
    \frac{\partial A}{\partial w_p}= \sum_{i=1}^t \frac{\frac{\partial f_{ip}}{\partial w_p}}{f_{ik}(w_{3k-2},w_{3k-1},w_{3k})}
\end{equation}
\begin{equation*}\label{eq:peq}
~~~~\forall p=1,2~...~3l~~and~~k=round(p/3)+1
\end{equation*}

The Hessian $\nabla^2 A$ is a block diagonal matrix of block size
3X3. For a given sequence, the entries of the 3X3 block will be the
same if that nucleotide belongs to the consensus pattern ($C_k$).
This nonlinear transformation will preserve all the critical points on the likelihood
surface. The theoretical details of the proposed method and their advantages are published in
\cite{Chiang96}. If we can identify
all the saddle points on the stability boundary of a given local
maximum, then we will be able to find all the tier-1 local maxima.
However, finding all of the saddle points is computationally
intractable and hence we have adopted a heuristic by generating the
eigenvector directions of the PSSM at the local maximum. 

\section{Neighborhood Profile Search}
\label{sec:algorithm}

We will now describe our novel neighborhood search framework which is applied in the profile space. Our framework consists of the following three stages:

\begin{itemize}

\item{{\it Global stage} in which the promising
solutions in the entire search space are obtained.} \item{{\it
Refinement stage} (or {\it local stage}) where a local method is applied to the solutions
obtained in the previous stage in order to refine the profiles.}
\item{{\it Neighborhood-search stage} where the exit points are
computed and the Tier-1 and Tier-2 solutions are explored
systematically.}

\end{itemize}

In the global stage, a branch and bound search is performed on the
entire dataset. All of the profiles that do not meet a certain
threshold (in terms of a given scoring function) are eliminated in
this stage. Some promising initial alignments are obtained by applying these methods (like projection methods) on the
entire dataset. Most promising set of alignments are considered for
further processing. The promising patterns obtained are transformed into
profiles and local improvements are made to these profiles in the
refinement stage. The consensus pattern is obtained from each
nucleotide that corresponds to the largest value in each column of
the PSSM.  The $3l$ variables chosen are the nucleotides that
correspond to those that are not present in the consensus pattern.
Because of the probability constraints discussed in the previous
section, the largest weight can be represented in terms of the other
three variables.

To solve Eq. (\ref{eq:final}), current algorithms begin at random
initial alignment positions and attempt to converge to an alignment
of $l-mer$s in all of the sequences that maximize the objective
function. In other words, the $l-mer$ whose $log(A)_i$ is the
highest (with a given PSSM) is noted in every sequence as part of
the current alignment. During the maximization of $A(Q)$ function,
the probability weight matrix and hence the corresponding alignments
of $l-mer$s are updated.  This occurs iteratively until the PSSM
converges to the local optimal solution. The consensus pattern is
obtained from the nucleotide with the largest weight in each
position (column) of the PSSM. This converged PSSM and the set of
alignments correspond to a local optimal solution. The neighborhood-search stage
where the neighborhood of the original solution is explored in a
systematic manner is shown below:

\noindent \textbf{Input:} Local Maximum (A).
\\\textbf{Output:} Best Local Maximum in the neighborhood region.
\\\textbf{Algorithm:}\\ {\it Step 1:} Construct the PSSM for the alignments corresponding to the local maximum (A) using Eqs.(\ref{eq:backcount}) and (\ref{eq:forecount}).
\\{\it Step 2:} Calculate the eigenvectors of the Hessian matrix for this PSSM.
\\{\it Step 3:} Find exit points ($e_{1i}$) on the practical stability boundary along each eigenvector direction.
\\{\it Step 4:} For each exit point, the corresponding Tier-1 local maxima ($a_{1i}$) are obtained by applying the EM algorithm after the ascent step.
\\{\it Step 5:} Repeat this process for promising Tier-1 solutions to obtain Tier-2 ($a_{2j}$) local maxima.
\\{\it Step 6:} Return the solution that gives the maximum information content score among $\{A,~a_{1i},~a_{2j}
\}$.\\

\begin{figure}
\centerline{
  \epsfig{figure=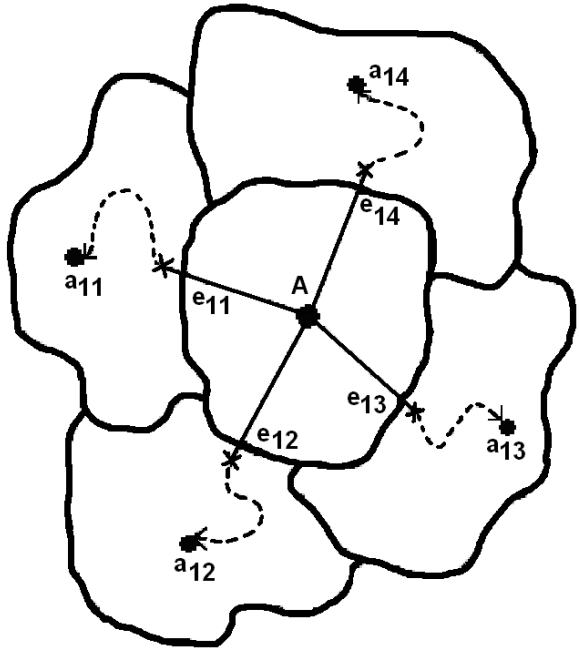, width=3.0in}
} \caption {Diagram illustrates the TRUST-TECH method of escaping
from the original solution ($A$) to the neighborhood local optimal
solutions ($a_{1i}$) through the corresponding exit points
($e_{1i}$). The dotted lines indicate the local convergence of the
EM algorithm.}

\label{fig:diagram}
\end{figure}

To escape out of this local optimal solution, our approach requires
the computation of a Hessian matrix (i.e. the matrix of second
derivatives) of dimension $(3l)^2$  and the $3l$ eigenvectors of the
Hessian. these directions were chosen as a general heuristic and are not problem dependent. Depending on the dataset that is being worked on, one can obtain even more promising directions. The main reasons for choosing the eigenvectors of the
Hessian as search directions are:
\begin{itemize}
\item{Computing the eigenvectors of the Hessian is related to finding
the directions with extreme values of the second derivatives, i.e.,
directions of extreme normal-to-isosurface change.}
\item{The eigenvectors of the Hessian will form the basis vectors for
the search directions. Any other search direction can be obtained by
a linear combination of these directions.} \item{This will make our
algorithm deterministic since the eigenvector directions are always
unique.}
\end{itemize}

\begin{figure}[htp]
\centerline{
  \epsfig{figure=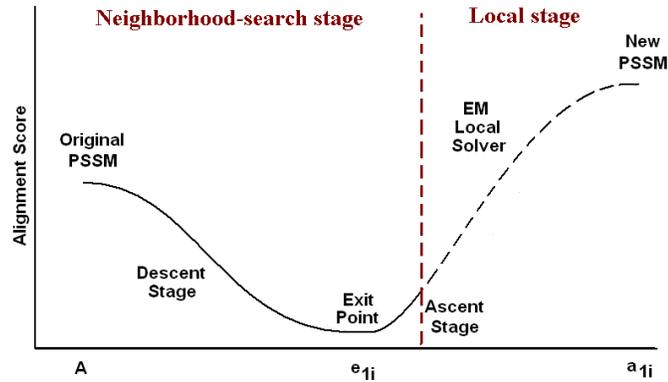, width=3.5in}
} \caption {A summary of escaping out of the local optimum to the
neighborhood local optimum.  Observe the corresponding trend of the alignment score at each step.}

\label{fig:summary}
\end{figure}

The value of the objective function is evaluated along these
eigenvector directions with some small step size increments. Since
the starting position is a local optimal solution, one will see a
steady decline in the function value during the initial steps; we
call this the {\it descent stage}. Since the Hessian is obtained
only once during the entire procedure, it is more efficient compared
to Newton's method where an approximate Hessian is obtained for
every iteration. After a certain number of evaluations, there may be
an increase in the value indicating that the stability boundary is reached. The point along this
direction intersecting the stability boundary is called the {\it exit
point}. Once the exit point has been reached, few more evaluations
are made in the direction of the same eigenvector to improve the chances of reaching a new convergence region. This procedure is
clearly shown in Fig ~\ref{fig:summary}. Applying the local method
directly from the exit point may give the original local maximum.
The ascent stage is used to ensure that the new guess is in a
different convergence zone. Hence, given the best local maximum
obtained using any current local methods, this framework allows us
to systematically escape out of the local maximum to explore
surrounding local maxima. The complete algorithm is shown below :

\noindent \textbf{Input:} The DNA sequences, length of the motif(l), Maximum
No. of Mutations(d) \\\textbf{Output:} Motif
(s)\\\textbf{Algorithm:}\\ {\it Step 1:} Given the sequences, apply
Random Projection algorithm to obtain different set of alignments.
\\{\it Step 2:} Choose the promising buckets and apply EM algorithm to
refine these alignments.
\\{\it Step 3:} Apply the TRUST-TECH method to obtain nearby promising local optimal solutions. \\{\it Step 4:}
Report the consensus pattern that corresponds to the best alignments and their corresponding PSSM.\\

The new framework can be treated as a hybrid approach between global
and local methods. It differs from traditional local methods by the ability to explore multiple local solutions in the neighborhood region in a
systematic and effective manner. It differs from global methods by working
completely in the profile space and searching a subspace efficiently
in a deterministic manner. However, the main difference of this work compared to the algorithm presented in the previous chapter is that the global method is performed in discrete space and the local method is performed in the continuous space. In other words, the both the global and local method do not optimize the same function. In such cases, it is even more important to search for neighborhood local optimal solutions in the continuous space.

\section{Implementation Details}
\label{sec:implementation}

Our program was implemented on Red Hat Linux version 9 and runs on a
Pentium IV 2.8 GHz machine. The core algorithm that we have
implemented is $TT\_EM$ described in Algorithm ~\ref{complete}.
$TT\_EM$ obtains the initial alignments and the original data
sequences along with the length of the motif. This procedure constructs the PSSM, performs EM refinement, and then
computes the Tier-1 and Tier-2 solutions by calling the procedure
$Next\_Tier$. The eigenvectors of the Hessian were computed using
the source code obtained from \cite{Press92}. $Next\_Tier$ takes a
PSSM as an input and computes an array of PSSMs corresponding to the
next tier local maxima using the TRUST-TECH methodology.

\begin{algorithm}
\caption{Motif $TT\_EM$($init\_aligns,seqs,l$)} \label{complete}
\begin{algorithmic}
\STATE$PSSM = Construct\_PSSM(init\_aligns)$ \STATE $New\_PSSM=
Apply\_EM(PSSM, seqs)$\STATE $TIER1 = Next\_Tier (seqs, New\_PSSM,
l)$ \FOR{$i=1$ to $3l$} \IF{$TIER1[i]<>zeros(4l)$}
     \STATE $TIER2[i][~] = Next\_Tier (seqs,
TIER1[i], l)$   \ENDIF \ENDFOR \STATE Return $best(PSSM,TIER1,
TIER2)$
\end{algorithmic}
\end{algorithm}

Given a set of initial alignments, Algorithm \ref{complete} will
find the best possible motif in the profile space in a tier-by-tier manner. For implementation considerations, we have shown only for two tiers. Initially, a PSSM is computed using $construct\_PSSM$ from
the given alignments. The procedure $Apply\_EM$ will return a new
PSSM that corresponds to the alignments obtained after the EM
algorithm has been applied to the initial PSSM. The details of the
procedure $Next\_Tier$ are given in Algorithm \ref{nexttier}. From a
given local solution (or PSSM), $Next\_Tier$ will compute all the
$3l$ new $PSSM$s corresponding to the tier-1 local optimal solutions. The second tier patterns are obtained by calling the
$Next\_Tier$ from the first tier solutions \footnote {New PSSMs
might not be obtained for certain search directions. In those cases,
a zero vector of length $4l$ is returned. Only those new PSSMs which
do not have this value will be used for any further processing.}.
Finally, the pattern with the highest score amongst all the PSSMs is
returned.

\begin{algorithm}
\caption{PSSMs[~] $Next\_Tier$($seqs, PSSM,l$)} \label{nexttier}
\begin{algorithmic}
\STATE$Score=eval(PSSM)$ \STATE
$Hess=Construct\_Hessian(PSSM)$\STATE $Eig[~]=Compute\_EigVec
(Hess)$ \STATE $MAX\_Iter=100$

\FOR{$k=1$ to $3l$} \STATE $PSSMs[k]= PSSM~~~~~~Count=0$ \STATE
$Old\_Score=Score~~~~~~~~~ep\_reached=FALSE$

\WHILE{$(!~ep\_reached)~ \&\& ~(Count<MAX\_Iter)$} \STATE
$PSSMs[k]=update(PSSMs[k],Eig[k],step)$ \STATE $Count~=~Count~+~1$
\STATE $New\_Score~=~eval(PSSMs[k])$

\IF{$(New\_Score~>~Old\_Score)$} \STATE $ep\_reached=TRUE$ \ENDIF

\STATE $Old\_Score=New\_Score$
    \ENDWHILE
\IF{$count<MAX\_Iter$}
    \STATE $PSSMs[k]=update(PSSMs[k],Eig[k],ASC)$
\STATE $PSSMs[k]=Apply\_EM(PSSMs[k],Seqs)$ \ELSE \STATE
$PSSMs[k]=zeros(4l)$ \ENDIF \ENDFOR \STATE \bf Return $PSSMs[~]$
\end{algorithmic}
\end{algorithm}

The procedure $Next\_Tier$ takes a PSSM, applies the TRUST-TECH
method and computes an array of PSSMs that corresponds to the next
tier local optimal solutions. The procedure $eval$ evaluates the
scoring function for the PSSM using (\ref{eq:final}). The procedures
$Construct\_Hessian$ and $Compute\_EigVec$ compute the Hessian
matrix and the eigenvectors respectively. $MAX\_iter$ indicates the
maximum number of uphill evaluations that are required along each of
the eigenvector directions. The neighborhood PSSMs will be stored in
an array variable $PSSMs$ (this is a vector). The original PSSM is updated with a
small step until an exit point is reached or the number of
iterations exceeds the $MAX\_Iter$ value. If the exit point is
reached along a particular direction, few more function evaluations are made to ensure that the PSSM has exited the original convergence region
and has entered a new one. The EM algorithm is then used during this
ascent stage to obtain a new PSSM \footnote{ For completeness, the
entire algorithm has been shown in this section. However, during the
implementation, several heuristics have been applied to reduce the
running time of the algorithm. For example, if the first tier
solution is not very promising, it will not be considered for
obtaining the corresponding second tier solutions.}.

The initial alignments are converted into the profile space and a
PSSM is constructed. The PSSM is updated (using the EM algorithm)
until the alignments converge to a local optimal solution. The
TRUST-TECH methodology is then employed to escape out of this local
optimal solution to compute nearby first tier local optimal
solutions.  This process is then repeated on promising first tier
solutions to obtain second tier solutions. As shown in Fig.
\ref{fig:diagram}, from the original local optimal solution,
various exit points and their corresponding new local optimal
solutions are computed along each eigenvector direction. Sometimes, two directions may yield the same local optimal solution. This can
be avoided by computing the saddle point corresponding to the exit
point on the stability boundary \cite{Reddy06f}. There can be many
exit points, but there will only be a unique saddle point
corresponding to the new local minimum. For computational efficiency, the
TRUST-TECH approach is only applied to promising initial alignments
(i.e. random starts with higher Information Content score).
Therefore, a threshold $A(Q)$ score is determined by the average of
the three best first tier scores after 10-15 random starts; any
current and first tier solution with scores greater than the
threshold is considered for further exploration. Additional random
starts are carried out in order to aggregate at least ten first tier
solutions. The TRUST-TECH method is repeated on all first tier
solutions above a certain threshold to obtain second-tier local optimal solutions.

\begin{figure}
\centerline{
  \epsfig{figure=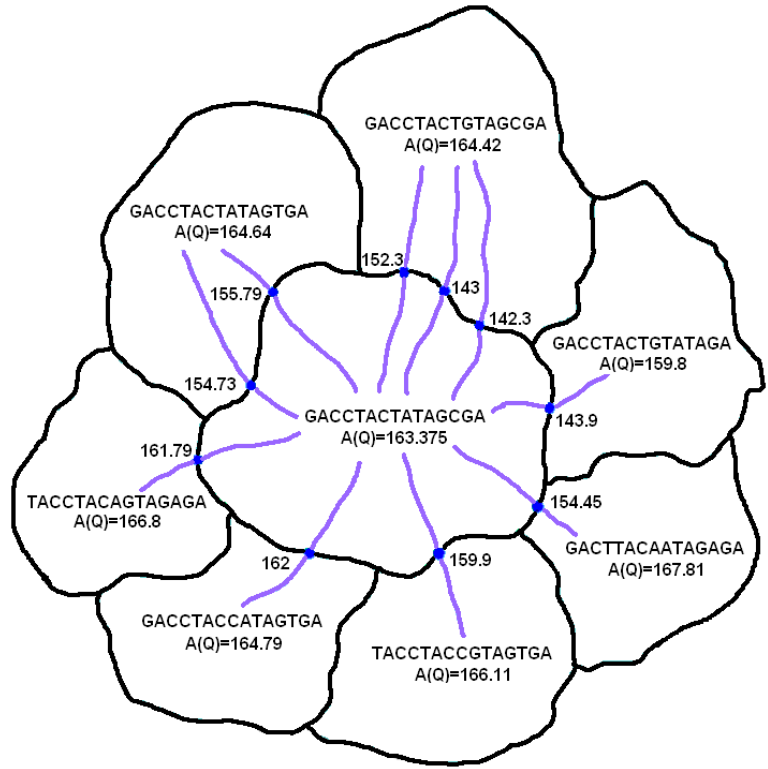, width=4.0in}
} \caption {2-D illustration of first tier improvements in a $3l$
dimensional objective function. The original local maximum has a
score of 163.375. The various Tier-1 solutions are plotted and the
one with highest score (167.81) is chosen.}

\label{fig:resultstier}
\end{figure}

\section{Experimental Results}
\label{sec:results} 

Experiments were performed on both synthetic
data and real data. Two different methods were used in the global
stage: random start and random projection. The main purpose of our work is not to demonstrate that our algorithm can outperform the
existing motif finding algorithms. Rather, the main work here
focuses on improving the results that are obtained from other
efficient algorithms. We have chosen to demonstrate the performance
of our algorithm on the results obtained from the random projection
method which is a powerful global method that has outperformed other
traditional motif finding approaches like MEME, Gibbs sampling,
WINNOWER, SP-STAR, etc. \cite{Buhler01}. Since the comparison was
already published, we mainly focus on the performance improvements
of our algorithm as compared to the random projection algorithm. For
the random start experiment, a total of $N$ random numbers between
$1$ and ($t-l+1$) corresponding to initial set of alignments are
generated. We then proceeded to evaluate our TRUST-TECH methodology
from these alignments.

Fig. \ref{fig:resultstier} shows the Tier-1 solutions obtained from
a given consensus pattern. Since the exit points are being used
instead of saddle points, our method might sometimes find the same
local optimal solution obtained before. As seen from the figure, the
Tier-1 solutions can differ from the original pattern by more than
just one nucleotide position. Also, the function value at the exit
points is much higher than the original value.

%
%
%
%

\subsection {Synthetic Datasets}
The synthetic datasets were generated by implanting some motif
instances into $t=20$ sequences each of length 600. Let $m$
correspond to one full random projection + EM cycle. We have set
$m=1$ to demonstrate the efficiency of our approach. We compared the
performance coefficient (PC) which gives a measure of the average
performance of our implementation compared to that of Random
Projection. The PC is given by :

\begin{equation}\label{eq:pc}
    PC=\frac{|K \cap P|}{|K \cup P|}
\end{equation}

\noindent where K is the set of the residue positions of the planted motif
instances, and P is the corresponding set of positions predicted by
the algorithm. Table \ref {TB:syn_das} gives an overview of the
performance of our method compared to the random projection
algorithm on the ({\it l,d}) motif problem for different {\it l} and
{\it d} values.

Our results show that by branching out and discovering multiple
local optimal solutions, higher $m$ values are not needed. A higher
$m$ value corresponds to more computational time because projecting
the $l$-mers into $k$-sized buckets is a time consuming task. Using
our approach, we can replace the need for randomly projecting {\it
l}-mers repeatedly in an effort to converge to a global optimum by
deterministically and systematically searching the solution space
modeled by our dynamical system and improving the quality of the
existing solutions. The improvements of our algorithm are clearly
shown in Table \ref{TB:syn_res}. We can see that, for higher length
motifs, the improvements are more significant.

\begin{sidewaystable}
\centering \caption{\protect  The consensus patterns and their
corresponding scores of the original local optimal solution obtained
from multiple random starts on the synthetic data. The best first
tier and second tier optimal patterns and their corresponding scores
are also reported. }
\begin{center}
{\scriptsize
\begin{tabularx}{8.5in}{|X||c|X|c|X|c|X|}
\hline
(l,d) &  Initial Pattern &Score&   First Tier Pattern&  Score&
Second Tier Pattern &Score \\ \hline
(11,2)&  AACGGTCGCAG &125.1& CCCGGTCGCTG& 147.1& CCCGGGAGCTG
&153.3\\ \hline (11,2)&
ATACCAGTTAC &145.7 &ATACCAGTTTC &151.3 &ATACCAGGGTC& 153.6 \\
\hline(13,3)& CTACGGTCGTCTT &  142.6 &CCACGGTTGTCTC &157.8&
CCTCGGGTTTGTC &  158.7\\ \hline (13,3)&  GACGCTAGGGGGT& 158.3&
GAGGCTGGGCAGT &161.7 &  GACCTTGGGTATT& 165.8\\ \hline (15,4)&
CCGAAAAGAGTCCGA& 147.5& CCGCAATGACTGGGT& 169.1& CCGAAAGGACTGCGT&
176.2\\ \hline (15,4)& TGGGTGATGCCTATG& 164.6& TGGGTGATGCCTATG&
166.7& TGAGAGATGCCTATG& 170.4\\ \hline (17,5) &TTGTAGCAAAGGCTAAA&
143.3 &CAGTAGCAAAGACTACC &  173.3& CAGTAGCAAAGACTTCC &  175.8\\
\hline (17,5)& ATCGCGAAAGGTTGTGG  & 174.1& ATCGCGAAAGGATGTGG &176.7
&ATTGCGAAAGAATGTGG &  178.3\\ \hline (20,6)& CTGGTGATTGAGATCATCAT&
165.9& CAGATGGTTGAGATCACCTT &   186.9& CATTTAGCTGAGTTCACCTT  & 194.9
\\ \hline(20,6)& GGTCACTTAGTGGCGCCATG &216.3& GGTCACTTAGTGGCGCCATG&
218.8& CGTCACTTAGTCGCGCCATG &   219.7\\ \hline
\end{tabularx}
}
\end{center}
 \label{TB:syn_res}
\end{sidewaystable}

\begin{figure}[htp]
\centerline{
  \epsfig{figure=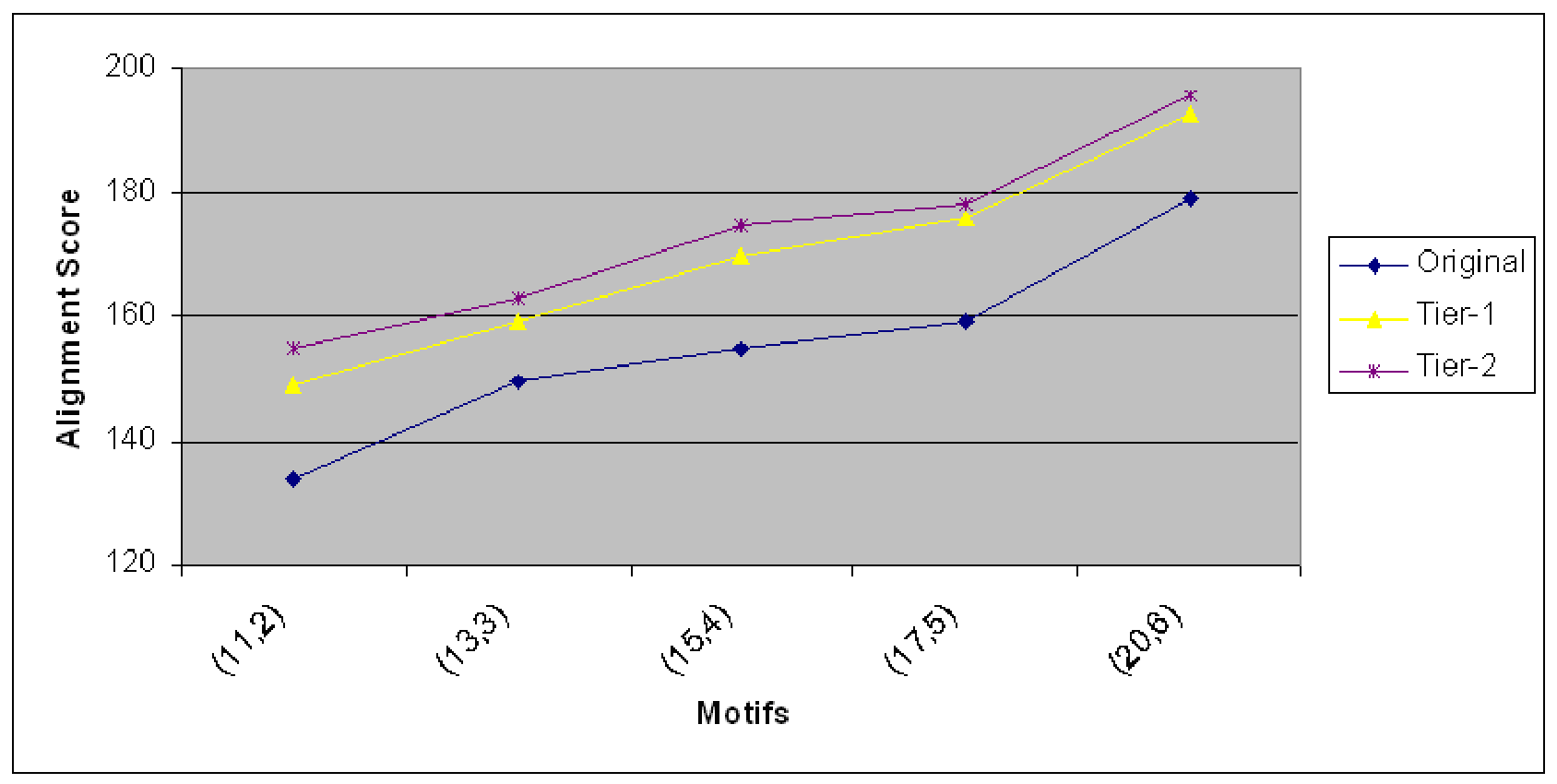, width=4.0in}
} \caption {The average scores with the corresponding first tier and
second tier improvements on synthetic data using the random starts
with TRUST-TECH approach with different ({\it l,d}) motifs.}
\label{fig:syn_rs}
\end{figure}

As opposed to stochastic processes like mutations in genetic
algorithms, our approach eliminates the stochastic nature and obtains
the nearby local optimal solutions systematically. Fig.
\ref{fig:syn_rs} shows the performance of the TRUST-TECH approach on
synthetic data for different ({\it l,d}) motifs. The average scores
of the ten best solutions obtained from random starts and their
corresponding improvements in Tier-1 and Tier-2 are reported. One
can see that the improvements become more prominent as the length of
the motif is increased. Table \ref{TB:syn_res} shows the best and
worst of these top ten random starts along with the consensus
pattern and the alignment scores.

\begin{figure}[htp]
\centerline{
  \epsfig{figure=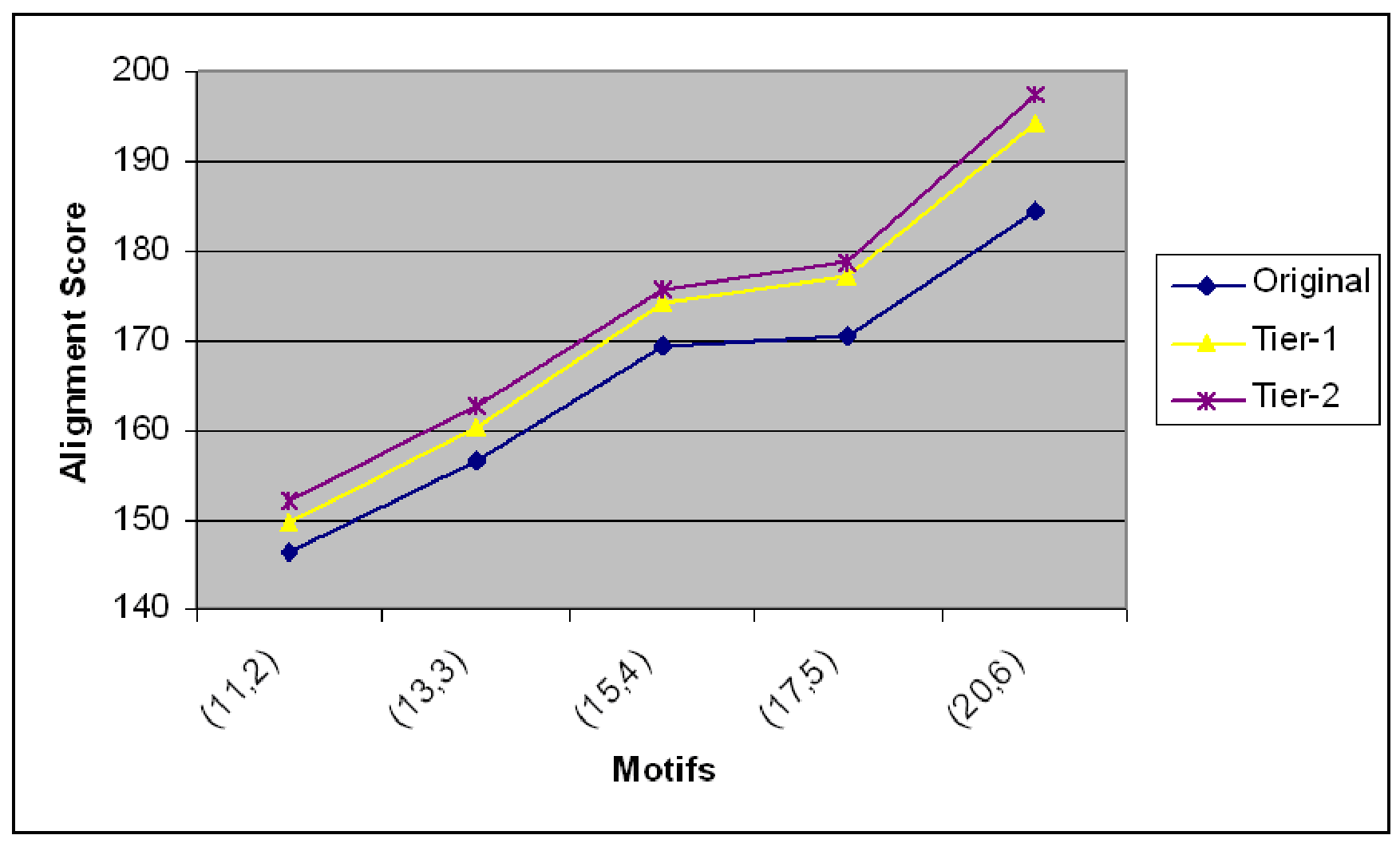, width=4.0in}
} \caption {The average scores with the corresponding first tier and
second tier improvements on synthetic data using the Random
Projection with TRUST-TECH approach with different ({\it l,d})
motifs.} \label{fig:syn_rp}
\end{figure}

With a few modifications, more experiments were conducted using the
Random Projection method.  The Random Projection method will
eliminate non-promising regions in the search space and gives a
number of promising sets of initial patterns. EM refinement is
applied to only the promising initial patterns. Due to the
robustness of the results, the TRUST-TECH method is employed only on
the top five local optima. The TRUST-TECH method is again repeated
on the top scoring first tier solutions to arrive at the second tier
solutions. Fig. \ref{fig:syn_rp} shows the average alignment scores
of the best random projection alignments and their corresponding
improvements in tier-1 and tier-2 are reported. In general, the
improvement in the first tier solution is more significant than the
improvements in the second tier solutions.

\begin{table}[htp]
\centering \caption{\protect The results of performance
coefficient with $m=1$ on synthetically generated sequences. The IC
scores are not normalized and the perfect score is 20 since there
are 20 sequences. }
\begin{center}
\begin{tabular}{|c|c|c|}
\hline

Motif  &  PC obtained using &PC obtained using\\
(l,d) &  Random Projection &TRUST-TECH method \\ \hline

(11,2)&  20& 20 \\  \hline(15,4) &14.875 &17\\  \hline (20,6)
&12.667&18\\ \hline
\end{tabular}
\end{center}
\label{TB:syn_das}
\end{table}

\subsection{Real Datasets}

Table~\ref{TB:realdata} shows the results of the TRUST-TECH
methodology on real biological sequences. We have chosen $l=20$ and
$d=2$. `{\it t}' indicates the number of sequences in the real data.
For the biological samples taken from \cite {Buhler01,Price03}, the
value $m$ once again is the average number of random projection + EM
cycles required to discover the motif. All other parameter values
(like projection size $k$=7 and threshold $s$=4) are chosen to be
the same as those used in the Random projection paper
\cite{Buhler01}. All of the motifs were recovered with $m=1$ using
the TRUST-TECH strategy. Without the TRUST-TECH strategy, the Random
Projection algorithm needed {\it multiple cycles} (m=8 in some cases
and m=15 in others) in order to retrieve the correct motif. This
elucidates the fact that global methods can only be used to a
certain extent and should be combined with refined local heuristics
in order to obtain better efficiency. Since the random projection
algorithm has outperformed other prominent motif finding algorithms
like SP-STAR, WINNOWER, Gibbs sampling etc., we did not repeat the
same experiments that were conducted in \cite{Buhler01}. Running one
cycle of random projection + EM is much more expensive
computationally. The main advantage of our strategy comes from the
deterministic nature of our algorithm in refining motifs.

\begin{table*}[htp]
\centering \caption{\protect Results of TRUST-TECH method on
biological samples. The real motifs were obtained in all the six
cases using the TRUST-TECH framework.}
\begin{center}
\begin{scriptsize}
\begin{tabular}{|c|c|c|c|c|}
\hline
\textbf{Sequence} &\textbf{Size}&\textbf{t}& \textbf{ Best
(20,2) Motif }&\textbf{ Reference Motif} \\\hline
E. coli CRP & 1890 & 18 &\underline {TGTGAAATAGATCACA}TTTT &TGTGANNNNGNTCACA \\
preproinsulin& 7689 &4 & GGAAATTGCAG\underline{CCTCAGCCC}& CCTCAGCCC \\
DHFR &800& 4&  CTGCA\underline{ATTTCGCGCCA}AACT& ATTTCNNGCCA \\
metallothionein &6823& 4 & CCCTC\underline{TGCGCCCGG}ACCGGT& TGCRCYCGG \\
c-fos &3695& 5&  \underline{CCATATTAGGACATCT}GCGT &CCATATTAGAGACTCT \\
yeast ECB& 5000 &5 &GTA\underline{TTTCCCGTTTAGGAAA}A
&TTTCCCNNTNAGGAAA \\ \hline
\end{tabular}
\end{scriptsize}
\end{center}
\label{TB:realdata}
\end{table*}

%

The TRUST-TECH framework broadens the search region in order to obtain an improved
solution which may potentially correspond to a better motif. In most
of the profile based algorithms, EM is used to obtain the nearest
local optimum from a given starting point. In our approach, we obtain promising results by computing multiple local optimal solutions in a tier-by-tier manner. We have shown on both real and synthetic data sets that
beginning from the EM converged solution, the TRUST-TECH approach is
capable of searching in the neighborhood regions for another
solution with an improved information content score. This will often
translate into finding a pattern with less hamming distance from the
resulting alignments in each sequence. Our approach has demonstrated
an improvement in the score on all datasets that it was tested on.
One of the primary advantages of the TRUST-TECH methodology is that
it can be used with different global and local methods. The main
contribution of our work is to demonstrate the capability of this
hybrid EM algorithm in the context of the motif finding problem. The TRUST-TECH approach can potentially use any global method and improve its
results efficiently. From these simulation results, we observe that motif refinement stage plays a vital
role and can yield accurate results deterministically.

In the future, we would like
to continue our work by combining other global methods available in
the literature with existing local solvers like EM or GibbsDNA that
work in continuous space. By following the example of
\cite{Tompa05}, we may improve the chances of finding more promising
patterns by combining our algorithm with different global and local
methods.

\chapter{Component-wise Smoothing for Learning Mixture Models}
\label{ch:smooth}

\section{Overview}

The task of obtaining an optimal set of parameters to fit a mixture model to a
given data can be formulated as a problem of finding the global maximum of its highly
nonlinear likelihood surface. This chapter introduces a new smoothing
algorithm for learning a mixture model from multivariate data. Our
algorithm is based on the conventional Expectation Maximization (EM)
approach applied to a smoothened likelihood surface. A family or
hierarchy of smooth log-likelihood surfaces is constructed using
convolution based approaches. In simple terms, our method first
smoothens the likelihood function and then applies the EM algorithm
to obtain a promising solution on the smooth surface. This solution
is used as an initial guess for the EM algorithm applied to the next
smooth surface in the hierarchy. This process is repeated until the
original likelihood surface is solved. The smoothing process reduces
the overall gradient of the surface and the number of local maxima.
This effective optimization procedure eliminates extensive search in
the non-promising regions of the parameter space. Results on
benchmark datasets demonstrate significant improvements of the
proposed algorithm compared to other approaches. Reduction in the
number of local maxima has also been demonstrated empirically
on several datasets.

As mentioned in Chapter \ref{ch:trust-tech-em}, one of the key problems with the EM algorithm is that it is a
`greedy' method which is sensitive to initialization. The
log-likelihood surface on which the EM algorithm is applied is very
rugged with many local maxima. Because of its greedy nature, EM
algorithm tends to get stuck at a local maximum that corresponds to
erroneous set of parameters for the mixture components. Obtaining an
improved likelihood function value not only provides better
parameter estimates but also enhances the generalization capability
of the given mixture models \cite{McLachlan00}. The fact that the local maxima
are not uniformly distributed makes it important for us to develop
algorithms that help in avoiding search in non-promising regions.
More focus needs to be given for searching the promising subspace by
obtaining promising initial estimates. This can be achieved by
smoothing the surface and obtaining promising regions and then
gradually trace back these solutions onto the original surface. In
this work, we develop a hierarchical smoothing algorithm for the
mixture modeling problem using convolution-based approach. We
propose the following desired properties for smoothing algorithms :
\begin {itemize}
\item{Preserve the presence of the global optimal solution.}
\item{Enlarge the convergent region (with respect to the EM algorithm) of the global optimal solutions and other promising sub-optimal solutions.}
\item{Reduce the number of local maximum or minimum.}
\item{Smooth different regions of the search space differently.}
\item{Avoid over smoothing which might make the surface too flat and cause convergence problems for the local solvers.}
\end{itemize}

\section{Relevant Background}
\label{sec:background}

Since EM is very sensitive to the initial set
of parameters that it begins with, several methods are proposed in
the literature to identify good initial points. All these methods are mentioned in Chapter \ref{ch:trust-tech-em}. In this section, we primarily focus on the literature relevant to smoothing algorithms.

Different smoothing strategies have been successfully used in
various applications for solving a diverse set of problems.
Smoothing techniques are used to reduce irregularities or random
fluctuations in time series data \cite{Shumway82,Beran99}. In the
field of natural language processing, smoothing techniques are also
used for adjusting maximum likelihood estimate to produce more
accurate probabilities for language models \cite{Chen96}.
Convolution based smoothing approaches are predominantly used in the
field of digital image processing for image enhancement by noise
removal \cite{Blake87,Chu98}. Other variants of smoothing techniques
include continuation methods \cite{Richter83,Dunlavy05} which are
used successfully in various applications. Different multi-level
procedures other than smoothing and its variants are clearly
illustrated in \cite{Teng99}.

For optimization problems, smoothing procedure helps in reducing the
ruggedness of the surface and helps the local methods to prevent the
local minima problem. It was used for the structure prediction of
molecular clusters \cite{Shao00}. This smoothing procedure obtains a hierarchy of smooth surfaces with a fewer and fewer local maxima. Promising initial points can be obtained by tracing back promising solutions at each level. This is an initialization procedure which has the capability to avoid searching non-promising regions. Obtaining the number of components using some model selection criterion \cite{McLachlan00} is not the primary focus of this work. In other words, our algorithm assumes that the number of components are known before hand. In summary, the main contributions are :

\begin {itemize}
\item{Develop convolution-based smoothing algorithms for obtaining optimal set of parameters.}
\item{Demonstrate that the density-based convolution on the entire dataset will result smoothing the likelihood surface with respect to the parameters.}
\item{Empirically show that the number of local maxima on the log-likelihood surface is reduced.}
\item{Show that smoothing is effective in obtaining promising initial set of parameters}
\end{itemize}

\begin{figure*}[htp]
   \centering
   \subfigure[Conventional method]{\includegraphics[width = 3.0 in]{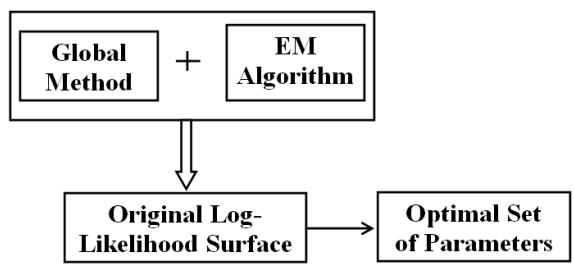}}\qquad
   \subfigure[Smoothing approach]{\includegraphics[width = 5.0 in]{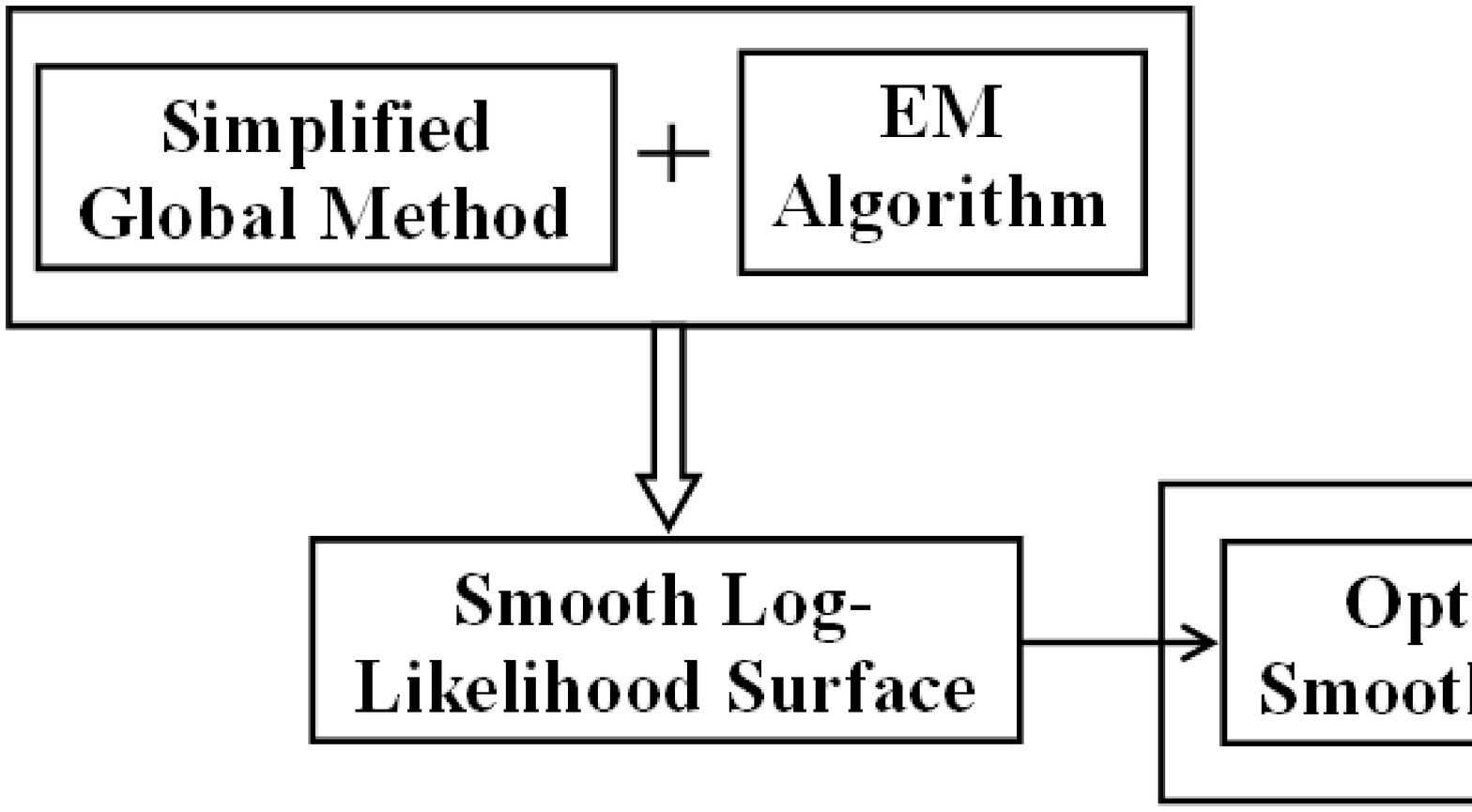}}\qquad

   \caption{\label{fig:block7} Block diagram of the traditional
   approach and the smoothing approach.}
 \end{figure*}
Fig. \ref{fig:block7} compares the conventional approach with the
smoothing approach. In the traditional approach, a global method in
combination with the EM algorithm is used to find the optimal set of
parameters on the log-likelihood surface. In the smoothing approach,
a simplified version of the global method is applied in combination
with the EM algorithm to obtain an optimal set of parameters on the
smooth surface which are again used in combination with the EM
algorithm to obtain optimal set of parameters on the original
log-likelihood surface. Since the smoothened log-likelihood surface
is easy to traverse (has fewer local maxima), one can gain
significant computational benefits by applying a simplified global
method compared to that of the conventional global method on the
original log-likelihood surface which are usually very expensive.

\begin{figure*}[htp]
\centerline{
  \epsfig{figure=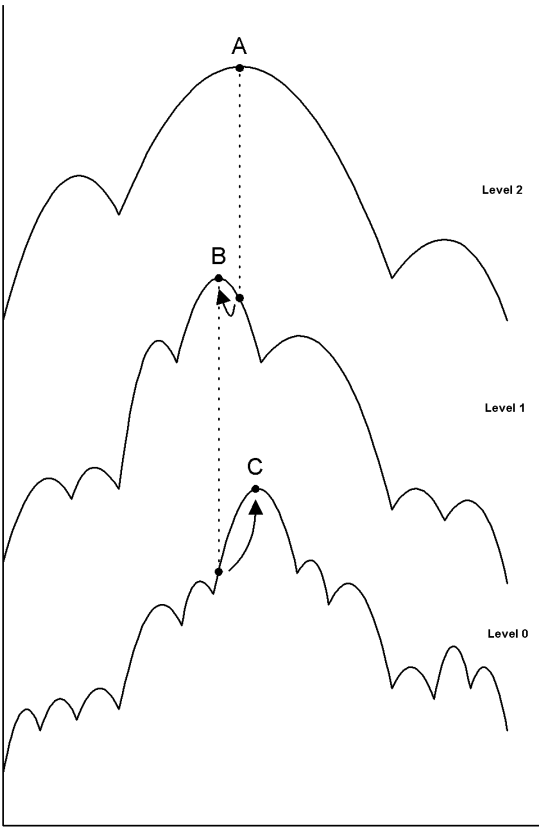, width=3.0in}
} \caption{Smoothing a nonlinear surface at different levels. Tracing the global maxima (A, B and C) at different levels.} \label{fig:smoothlevels}
\end{figure*}
Fig. \ref{fig:smoothlevels} shows a nonlinear surface and its smoothened versions in one dimension. During the smoothing process, there is no guarantee that the global maximum will retain its location. However, if the smoothing factor is not changed significantly, one can carefully traceback the global maximum by applying the EM algorithm at each level. For example, `C' is the global maximum of the original nonlinear surface which contains 11 local maxima and is indicated by level 0. Two smoothened versions at level 1 (with 7 local maxima) and level 2 (with 3 local maxima) can be constructed using kernel smoothing. `B' indicates the global maximum in level 1 and `A' indicates the global maximum in level 2. Obtaining the global maximum in level 2 is relatively easier compared to the global maximum in level 0 because of the presence of fewer local maxima. Once the point `A' is obtained, it is used as an initial guess for the EM algorithm at level 1 to obtain `B'. Similarly `C' in level 0 is obtained by applying EM algorithm initialized with `B'.\\

\section{Preliminaries}
\label{sec:problem} 
\begin{figure}[htp]
   \centering
   \subfigure[]{\includegraphics[width = 1.6 in]{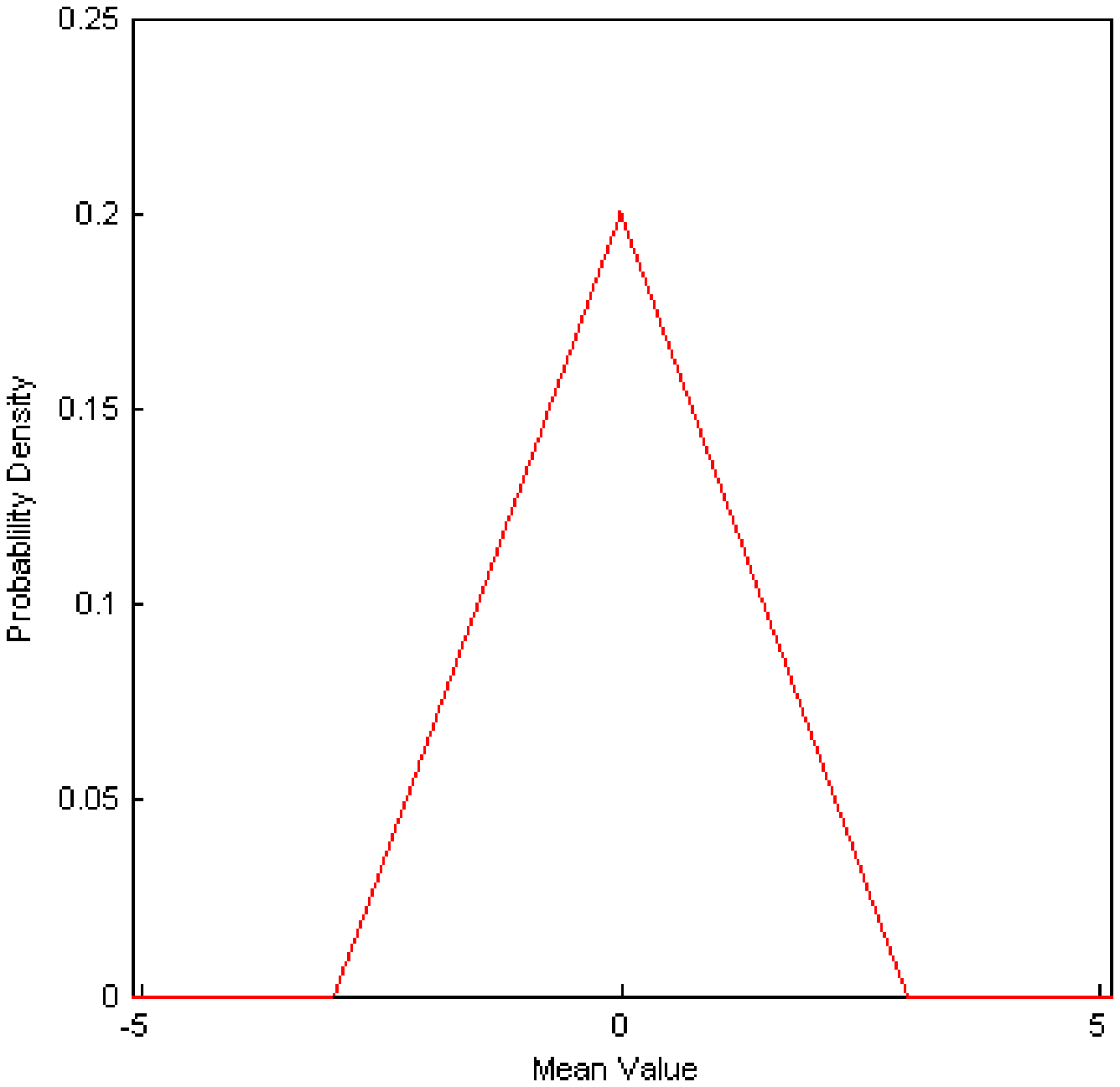}}\qquad
   \subfigure[]{\includegraphics[width = 1.6 in]{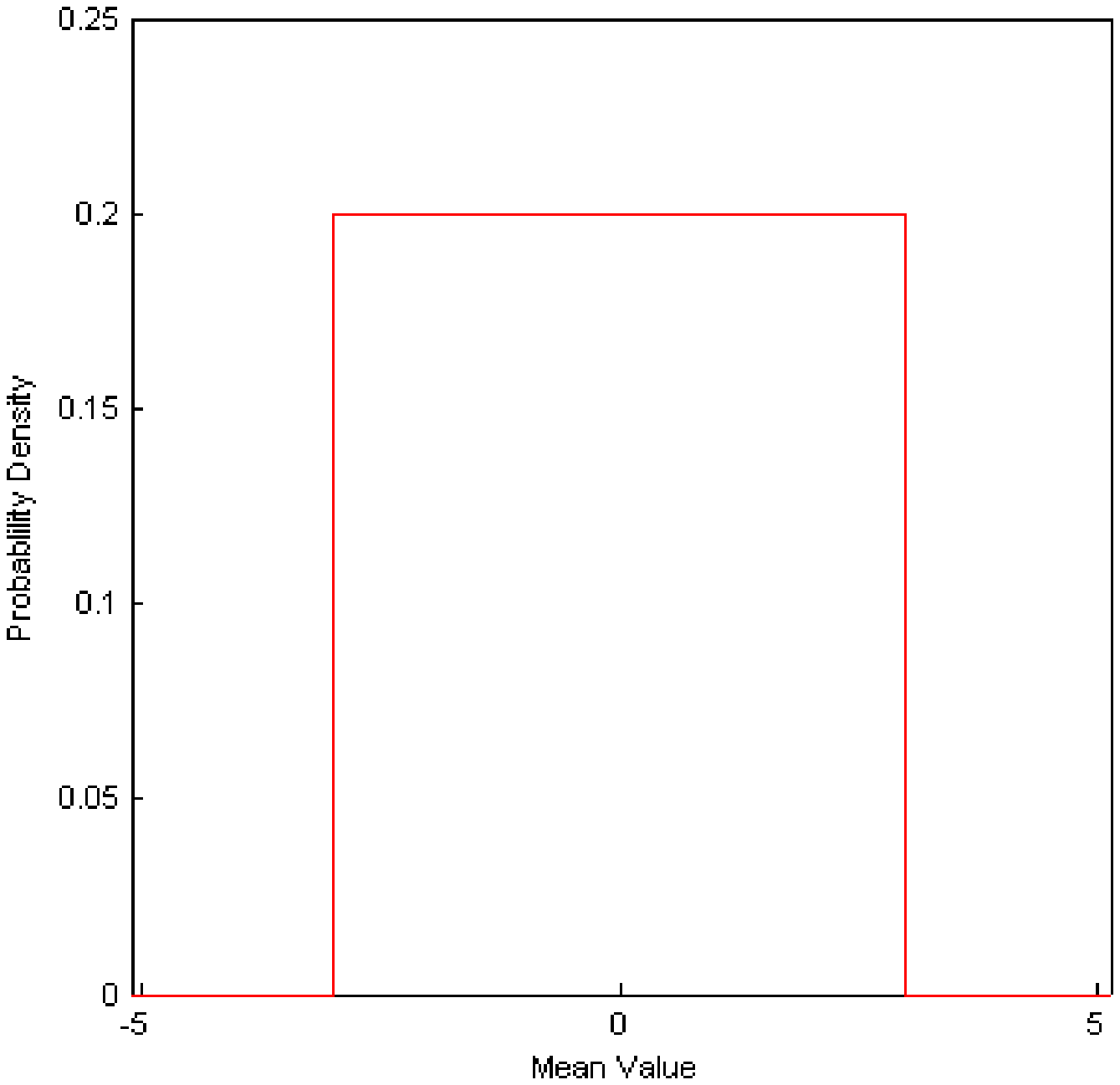}}\qquad
   \subfigure[]{\includegraphics[width = 1.6 in]{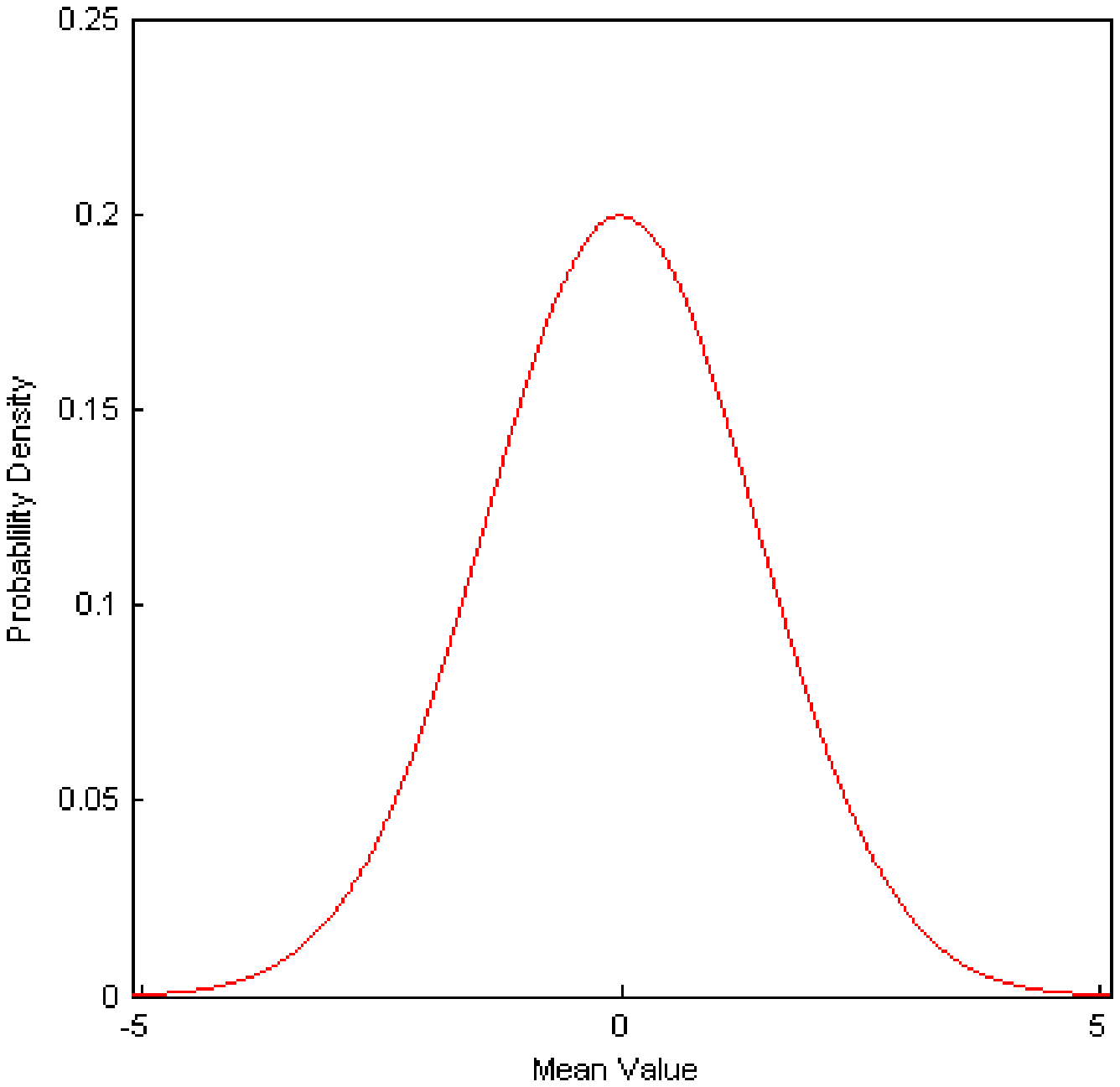}}\qquad
   \caption{\label{fig:kern}Different convolution kernels (a)
   Triangular function (b) Step function and (c) Gaussian function
   }
 \end{figure}

We will now introduce some preliminaries
on convolution kernels. For smoothing the mixture model, any kernel can be used for
convolution if it can yield a closed form solution in each E and M
step. Three widely used kernels are shown in Fig. \ref{fig:kern}. We
choose to use Gaussian kernel for smoothing the original
log-likelihood function for the following reasons :
\begin {itemize}
\item{When the underlying distribution is assumed to be generated from Gaussian components, Gaussian kernels give the optimal performance.}
\item{The analytic form of the likelihood surface obtained after smoothing is very similar to the original likelihood surface. }
\item{Since the parameters of the original components and the kernels will be of the same scale, changing the parameters correspondingly to scale will be much easier.}
\end{itemize}

Instead of convolving the entire likelihood surface directly, we
convolve individual components of the mixture model separately. Lets
consider a Gaussian density function parameterized by
$\theta_i$(i.e. $\mu_i$ and $\sigma_i$) under the assumption that
all the components have the same functional form (d-variate
Gaussians):


\begin{equation}\label{eq:gaussiandensity}
    p(x|\theta_i) =\frac{1}{\sigma_i \sqrt{2\pi}}e^{-\frac{(x-\mu_i)^2}{2\sigma_i^2}}
\end{equation}

Lets consider the following Gaussian kernel :
\begin{equation}\label{eq:convker}
    g(x) =\frac{1}{\sigma_0 \sqrt{2\pi}}e^{-\frac{(x-\mu_0)^2}{2\sigma_0^2}}
\end{equation}

\begin{equation}\label{eq:convolution}
\begin{split}
    p'(x|\theta_i) = p(x|\theta_i) \otimes g(x)
    &=\frac{1}{\sigma_i
    \sqrt{2\pi}}e^{-\frac{(x-\mu_i)^2}{2\sigma_i^2}}\otimes    \frac{1}{\sigma_0
    \sqrt{2\pi}}e^{-\frac{(x-\mu_0)^2}{2\sigma_0^2}}\\ &\quad     =\frac{1}{\sqrt{2\pi (\sigma_i^2+\sigma_0^2)}}e^{-\frac{(x-(\mu_i+\mu_0))^2}{2(\sigma_i^2+\sigma_0^2)}}
\end{split}
\end{equation}

{\it Convolution of Gaussians:} When a Gaussian density function
with parameters $\mu_1$ and $\sigma_1$ is convolved with a Gaussian
kernel with parameters $\mu_0$ and $\sigma_0$, then the resultant
density function is also Gaussian with mean $(\mu_1+\mu_0)$ and
variance $(\sigma_1^2+\sigma_0^2)$. The proof for this claim is
given in Appendix-A.

\begin{figure}[htp]
\centerline{
  \epsfig{figure=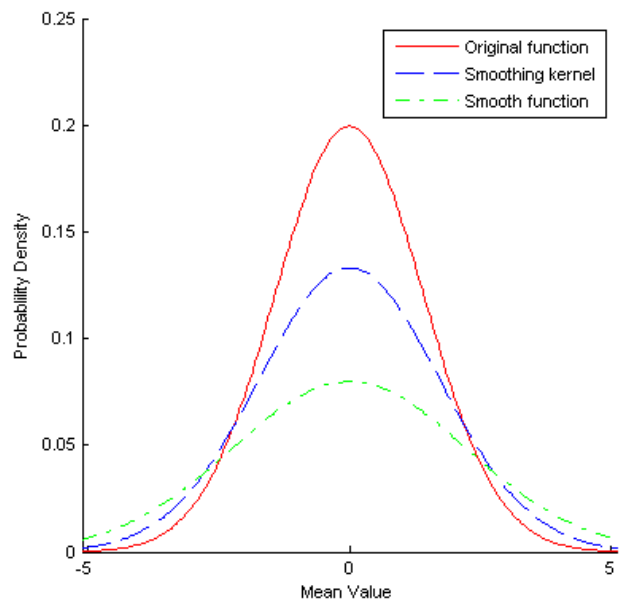, width=3.25in}
} \caption{The effects of smoothing a Gaussian density function with
a Gaussian kernel.} \label{fig:smooth}
\end{figure}

Now, the new smooth density can be obtained by convolving with the
Gaussian kernel given by Eq. (\ref{eq:convker}). Convolving two
Gaussians to obtain another Gaussian is shown graphically in Fig.
\ref{fig:smooth}. It can also be observed that if the mean of one of
the Gaussians is zero, then the mean of the resultant Gaussian is
not shifted. The only change is in the variance parameter. Since,
shifting mean is not a good choice for optimization problems and we
are more interested in reducing the peaks, we chose to increase the
variance parameter without shifting the mean.

\section{Smoothing Log-Likelihood Surface}
\label{sec:smooth}

The overall log-likelihood surface can be convolved using a Gaussian
kernel directly. This is not a feasible approach because of the
following reasons :

\begin {itemize}
\item{It results in an analytic expression that is not easy to work on and computing the EM updates will become cumbersome.}
\item{It is Computationally very expensive.}
\item {Different regions of search space must be smoothened differently. Choosing parameters to do this task is hard.}

\end{itemize}

To avoid the first problem, we exploit the structure of the problem.
Since the log-likelihood surface is obtained from individual
densities, smoothing each component's individual density function
will smoothen the overall log-likelihood surface. This will also
give the flexibility to chose the kernel parameters which is
discussed in following subsection.

After computing the new density ($p'$), we can define the
\begin{equation}\label{eq:pdash}
    p'(x|\Theta) = \sum_{i=1}^k \alpha_i
    ~p'(x|\theta_i)
\end{equation}

Now, the smooth log-likelihood function is given by:

\begin{equation}
f'(\mathcal{X}, \Theta)=~ \sum_{j=1}^n ~log~\sum_{i=1}^k \alpha_i
    ~p'(x^{(j)}|\theta_i) \label{eq:smooth}
\end{equation}

\begin{thm}\label{th:convmu}{\it (Density Smoothing):}
Convolution of a Gaussian density function with parameters $\mu_1$
and $\sigma_1$ with a Gaussian kernel with parameters $\mu_0=0$ and
$\sigma_0$ is equivalent to convolving the function with respect to
$\mu_1$.
\end{thm}
{\it Proof: See Appendix-B.}\\

\begin{figure*}[htp]
\centerline{
  \epsfig{figure=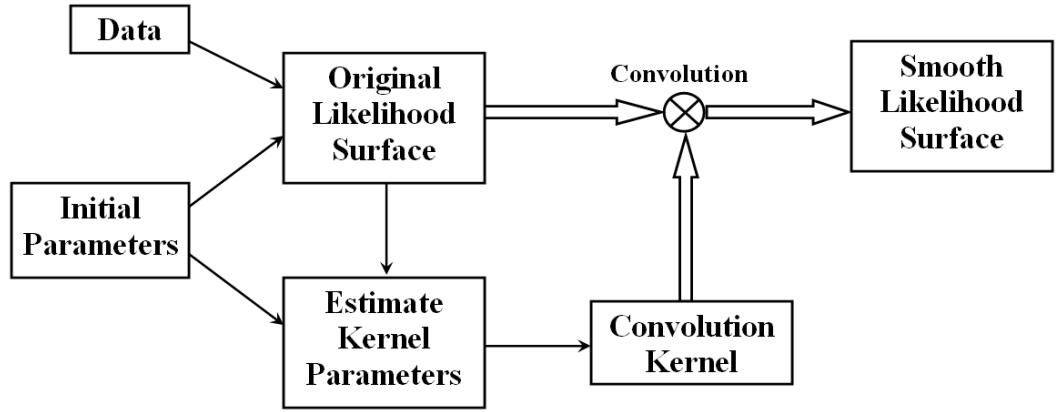, width=5.5in}
} \caption{Block Diagram of the smoothing approach. Smooth
likelihood surface is obtained by convolving the original likelihood
surface with a convolution kernel which is chosen to be a Gaussian
kernel in our case.} \label{fig:block1}
\end{figure*}

Fig. \ref{fig:block1} shows the block diagram of the smoothing
procedure. The original likelihood surface is obtained from the
initial set of parameters and the given dataset. The kernel
parameters are chosen from the initial set of parameters and the
original log-likelihood surface. The kernel is then convolved with
the original log-likelihood surface to obtain smooth log-likelihood
surface.

\subsection{Kernel Parameters}
The parameters of the smoothing kernel can be chosen to be fixed so
that they need not depend on the parameters of individual
components. Fixed kernels will be effective when the underlying
distribution comes from similar components. The main disadvantage of
choosing a fixed kernel is that some of the components might not be
smoothened while others might be over smoothened. Since, the
Gaussian kernel has the property that the convolution sums up the
parameters, this can also be treated as
{\it Additive smoothing}. To avoid the problems of fixed kernel
smoothing, we introduce the concept of variable kernel smoothing. Each component will be treated differently and smoothed
according to the existing parameter values. This smoothing strategy
is much more flexible and works effectively in practice. Since, the
kernel parameters are effectively multiplied, this smoothing can be
considered as {\it Multiplicative Smoothing}. In other words,
$\sigma_0$ must be chosen individually for different components and
it must be a function of $\sigma_i$. Both these approaches don't
allow for smoothing the mixing weight parameters ($\alpha$'s).

\subsection{EM Updates}
For both of the above mentioned smoothing kernels, the following
equations are valid. The complete derivations of these EM equations
for the case of fixed kernel smoothing is given in Appendix-C. The
$Q-function$ of the EM algorithm applied to the smoothened
log-likelihood surface is given by:

 \begin{equation}\label{eq:qfuncgmm}
 Q(\Theta|\widehat{\Theta}(t))= \sum_{j=1}^{n}\sum_{i=1}^{k}  w_i^{(j)}[log\frac{1}{\sqrt{2\pi(\tilde{\sigma}_i^2)}}
 \\-\frac{(x^{(j)}-\tilde{\mu}_i)^2}{2\tilde{\sigma}_i^2}+log ~\alpha_i]
\end{equation}

where
 \begin{equation}\label{eq:expectz}
w_i^{(j)}=\frac{\frac{\alpha_i(t)}{\tilde{\sigma}_i}e^{-\frac{1}{2\tilde{\sigma}_i^2}(x^{(j)}-\tilde{\mu}_i(t))^2}}{\sum_{i=1}^k
\frac{\alpha_i(t)}{\tilde{\sigma}_i}e^{-\frac{1}{2\tilde{\sigma}_i^2}(x^{(j)}-\tilde{\mu}_i(t))^2}}
 \end{equation}

$\tilde{\Theta}$  represents the smoothened parameters. The updates for the maximization step in the case of GMMs are given
as follows :

\begin{eqnarray}
\tilde{\mu}_i(t+1) = \frac{\sum_{j=1}^{n}w_i^{(j)}x^{(j)}}{\sum_{j=1}^{n}w_i^{(j)}}\\
\tilde{\sigma}_i^2(t+1) = \frac{\sum_{j=1}^{n}w_i^{(j)} (x^{(j)}-(\tilde{\mu}_i(t+1))^2}{\sum_{j=1}^{n}w_i^{(j)}}\\
\tilde{\alpha}_i(t+1)=\frac{1}{n}\sum_{j=1}^{n}w_i^{(j)}
\label{eq:update}
\end{eqnarray}

\section{Algorithm and its implementation}
\label{sec:algorithm}

\begin{figure}[htp]
\centerline{
  \epsfig{figure=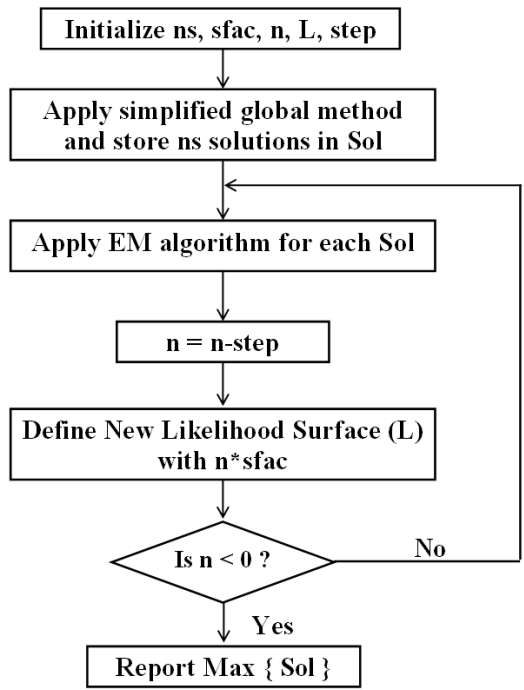, width=3.0in}
} \caption{Flowchart of the smoothing algorithm} \label{fig:flow}
\end{figure}

This section describes the smoothing algorithm in detail and
explains the implementation details. The basic advantage of the
smoothing approach is that a simplified version of the global method can be used to
explore fewer promising local maxima on the smoothened surface.
These solutions are used as initial guesses for the EM algorithm
which is again applied to the next level of smoothing. Smoothing
will help to avoid search in non-promising areas of the
parameter space. Fig. \ref{fig:flow} gives the flow chart of our
smoothing algorithm.

Before describing the algorithm, we first introduce certain
variables that are used. The likelihood surface (defined by L)
depends on the parameters and the available data. The smoothing
factor ($sfac$) determines the extent to which the likelihood
surface needs to be smoothened (which is usually chosen by
trial-and-error). $ns$ denotes the number of
solutions that will be traced. $n$ determines number of levels in
the smoothing hierarchy. It is clear that there is a trade-off
between the number of levels and the accuracy of this method. Having many levels might increase the accuracy of the
solutions, but it is computationally expensive. On the other hand,
having few levels is computationally very cheap, but we might have
to forgo the quality of the final solution. Deciding these
parameters is not only user-specific but also depends significantly
on the data that is being modeled. Algorithm \ref{alg:smooth}
describes the smoothing approach.

\begin{algorithm}
\caption{Smooth-EM Algorithm} \label{alg:smooth}
\begin{algorithmic}
\STATE \textbf{Input:} Parameters $\Theta$, Data  $\mathcal{X}$,
Tolerance $\tau$, Smooth factor $Sfac$, number of levels $nl$,
number of solutions $ns$ \STATE \textbf{Output:}
$\widehat{\Theta}_{MLE} $ \STATE \textbf{Algorithm:} \STATE
step=1/nl ~~~~ Sfac=Sfac/nl \STATE
L=Smooth($\mathcal{X},\Theta$,nl*Sfac) \STATE
Sol=Global($\mathcal{X},\Theta$, L,ns)
 \WHILE{n $\geq$ 0}

\STATE nl=nl-step \STATE L=Smooth($\mathcal{X},\Theta$,nl*Sfac)

\FOR {i=1:ns}\STATE Sol(i)=EM(Sol(i),$\mathcal{X}$,L,$\tau$) \ENDFOR

   \ENDWHILE
\STATE $\widehat{\Theta}_{MLE}$ =max\{Sol\}

\end{algorithmic}
\end{algorithm}

The algorithm takes smoothing factor, number of levels, number of
solutions, parameters set and the data as input and computes the global maximum on the log-likehood surface. Smooth function returns the
likelihood surface corresponding to smoothing factor at each level.
Initially, a simple global method is used to identify promising
solutions ($ns$) on the smooth likelihood surface which are stored
in $Sol$. With these solutions as initial estimates, we then apply
EM algorithm on the likelihood surface corresponding to the next
level smooth surface. The EM algorithm also returns $ns$ number of
solutions corresponding to the $ns$ number of initial estimates. At
every iteration, new likelihood surface is constructed with a
reduced smoothing factor. This process is repeated until the
smoothing factor becomes zero which corresponds to the original
likelihood surface. Though, it appears to be a daunting task, it can
be easily implemented in practice. The main idea is to construct a
family or hierarchy of surfaces and carefully trace the promising
solutions from the top most surface to the bottom most one. In terms
of tracing back the solutions to uncoarsened models, our method
resembles other multi-level methods proposed in
\cite{Karypis99,Dasgupta00}. The main difference is that the
dimensionality of the parameter space is not changed during the
smoothing (or coarsening) process.

\section{Results and Discussion}
\label{sec:results}

Our algorithm has been tested on three different datasets. The
initial values for the centers were chosen from the available data
points randomly. The covariances were chosen randomly and uniform
prior is assumed for initializing the components.

A simple synthetic data with 40 samples and 5 spherical Gaussian
components was generated and tested with our algorithm. Priors were
uniform and the standard deviation was 0.01. The centers for the
five components are given as follows : $\mu_1=[0.3~0.3]^T$,
$\mu_2=[0.5~0.5]^T$, $\mu_3=[0.7~0.7]^T$, $\mu_4=[0.3~0.7]^T$ and
$\mu_5=[0.7~0.3]^T$.

\begin{figure}[htp]
\centerline{
  \epsfig{figure=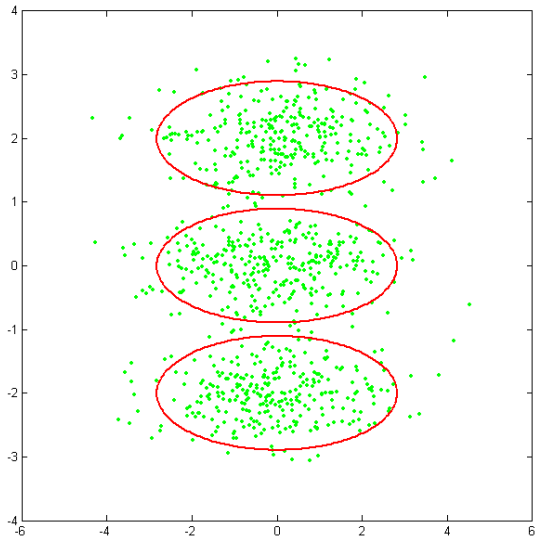, width=2.8in}
} \caption{True mixture of the three Gaussian components with 900
samples.} \label{fig:true}
\end{figure}

\begin{figure}[htp]
\centerline{
  \epsfig{figure=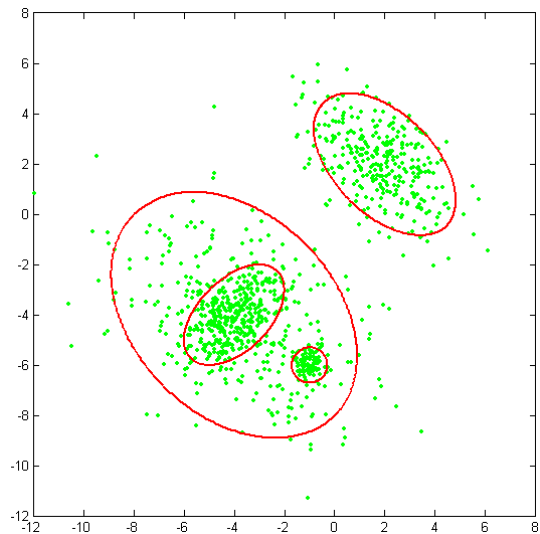, width=2.8in}
} \caption{True mixtures of the more complicated overlapping
Gaussian case with 1000 samples.} \label{fig:third}
\end{figure}

The second dataset was that of a diagonal covariance case. The data
generated from a two-dimensional, three-component Gaussian mixture
distribution \cite{Ueda98} with mean vectors at $[0 ~-2]^T, [0~
0]^T,[0 ~2]^T$ and same diagonal covariance matrix with values 2 and
0.2 along the diagonal. All the three mixtures have uniform priors.
The true mixtures with data generated from these three components
are shown in Fig. \ref{fig:true}. In the third synthetic dataset, a
more complicated overlapping Gaussian mixtures are considered
\cite{Figueiredo02}. It has four components with 1000 data samples
(see Fig. \ref{fig:third}). The parameters are as follows :
$\mu_1=\mu_2=[-4~ -4]^T$ , $\mu_3 =[2~2]^T$ and $\mu_4=[-1~-6]^T$.
$\alpha_1=\alpha_2=\alpha_3=0.3$ and $\alpha_4=0.1$.

\begin{displaymath}
     C_1=\left[ \begin{array}{cc} 1 &0.5\\  0.5 & 1 \end{array}
     \right]~~~~~~~~~~~~~~~C_2=\left[ \begin{array}{cc} 6 &-2\\  -2 & 6 \end{array}
     \right]
 \end{displaymath}
\begin{displaymath}
    C_3=\left[ \begin{array}{cc} 2 &-1\\  -1 & 2 \end{array}
     \right]~~~~~~~~~~~~~~~ C_4=\left[ \begin{array}{cc} 0.125 &0\\  0 &0.125  \end{array}
     \right]
 \end{displaymath}

%

\begin{figure*}[htp]
   \centering
   \subfigure[]{\includegraphics[width = 2.7 in]{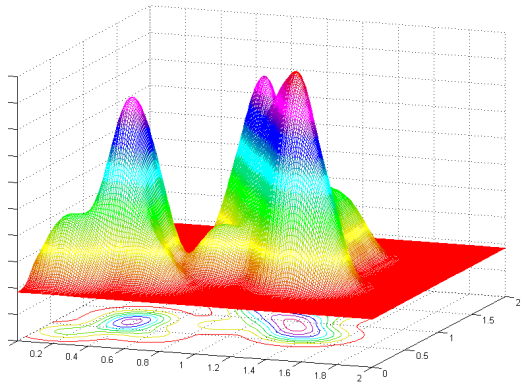}}\qquad
   \subfigure[]{\includegraphics[width = 2.7 in]{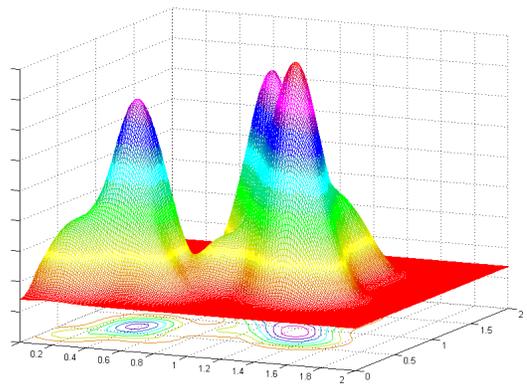}}\qquad
   \subfigure[]{\includegraphics[width = 2.7 in]{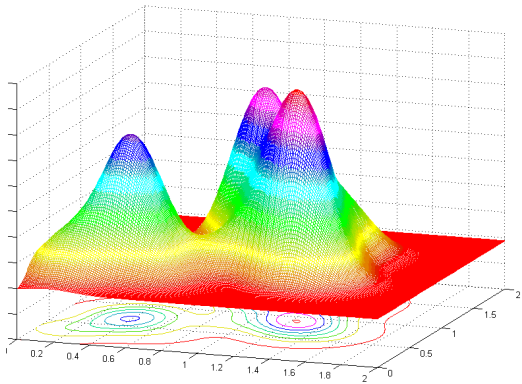}}\qquad
   \subfigure[]{\includegraphics[width = 2.7 in]{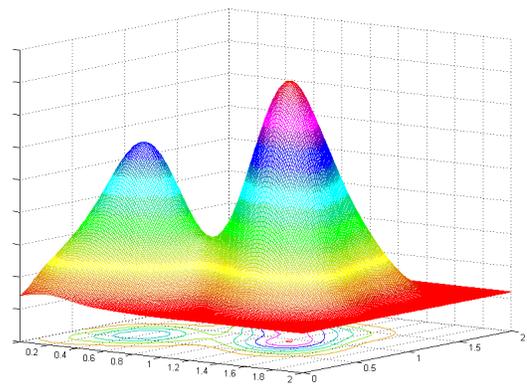}}\qquad
   \caption{\label{fig:diagcov}Various stages during the smoothing process. (a) The original log-likelihood surface which is very rugged (b)-(c) Intermediate smoothened surfaces (d)
   Final smoothened surface with only two local maxima.
   }
 \end{figure*}
\subsection{Reduction in the number of local maxima}
One of the main advantages of the proposed smoothing algorithm is to
ensure that the number of local maxima on the likelihood surface has
been reduced. To the best of our knowledge, there is no theoretical
way of estimating the amount of reduction in the number of unique
local maximum on the likelihood surface. We hence use empirical
simulations to justify the fact that the procedure indeed reduces
the number of local maxima. Fig. \ref{fig:diagcov} demonstrates the
capability of our algorithm to reduce the number of local maxima. In
this simple case, there were six local maxima originally, which were
reduced to two local maxima after smoothing. Other stages during the
transformation are also shown. 

\begin{figure*}
   \centering
   \subfigure[Spherical Dataset]{\includegraphics[width = 2.55 in]{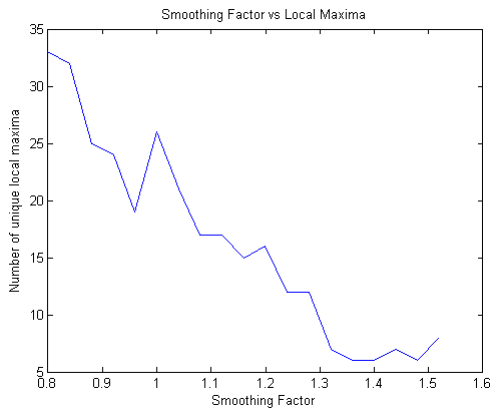}}\qquad
   \subfigure[Elliptical Dataset]{\includegraphics[width = 2.55 in]{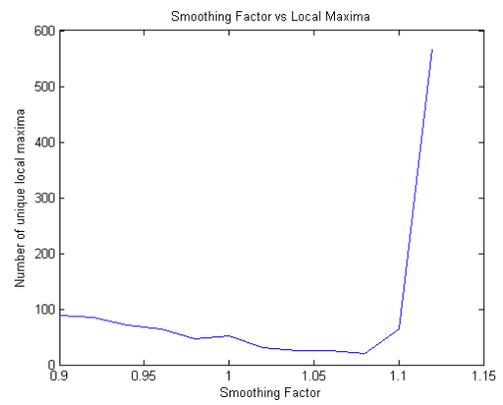}}\qquad
   \subfigure[Iris Dataset]{\includegraphics[width = 2.55 in]{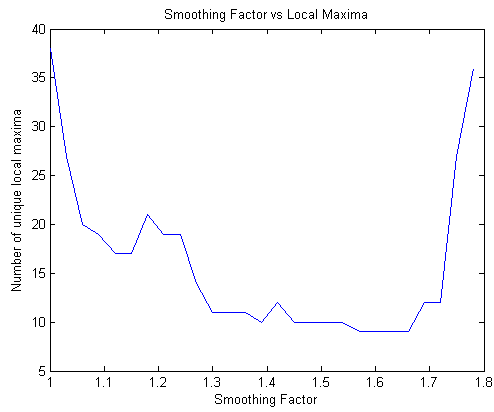}}\qquad
   \caption{\label{fig:local} Reduction in the number of local
   maxima for various datasets.
   }
 \end{figure*}

Fig. \ref{fig:local} shows the variation of the number of local
maxima with respect to the smoothing factor for different datasets.
One can see that if the smoothing factor is increased beyond a
certain threshold value ($\sigma_{opt}$), the number of local maxima
increases rapidly. This might be due to the fact that over-smoothing
the surface will make the surface flat, thus making it difficult for
the EM to converge. Experiments were conducted using 1000 random
starts and the number of unique local maxima were stored.

\subsection{Smoothing for Initialization}
Smoothing the likelihood surface also helps in the optimization
procedure. Experiments were conducted using 100 random starts. The
average across all the starts is reported. The surface is then
smoothened and some promising solutions are used to trace the
local optimal solutions and the average across all these starts are
reported. Table \ref{TB:compresults} summarizes the results obtained
directly with the original likelihood and the smoothened likelihood.
We have used only two levels and tracked three solutions for each
level. The two main claims (reduction in the number of local maximum
and better initial estimates) about the contributions have been justified.

\begin{table}[htp]
\centering \caption{\protect Comparison of smoothing algorithm
with the random starts. Mean and standard deviations across 100
random starts are reported.}
\begin{center}
\begin{tabular}{|c|c|c|}
\hline
Dataset & RS+EM & Smooth+EM\\
\hline Spherical & 36.3 $\pm$ 2.33 & 41.22 $\pm$ 0.79\\
\hline Elliptical & -3219 $\pm$ 0.7&-3106 $\pm$ 12\\
\hline Full covariance &-2391.3 $\pm$ 35.3&-2164.3 $\pm$ 18.56\\
\hline Iris &-196.34 $\pm$ 15.43&-183.51 $\pm$ 2.12\\
\hline
\end{tabular}
\end{center}
\label{TB:compresults}
\end{table}

More sophisticated global methods like Genetic algorithms, simulated annealing, adaptive partitioning \cite{Tang94} etc. and their
simplified versions can also be used in combination with our approach. Since the main focus of our work is to demonstrate the
smoothing capability, we used multiple random restarts as our global method.

Our algorithm is based on the conventional Expectation Maximization (EM) approach applied to a smoothened likelihood surface. A hierarchy of smooth surfaces is constructed and optimal set of parameters are obtained by tracing back the promising solutions at each level. The basic idea here is to obtain promising solutions in the smoothened surface and carefully trace the corresponding solutions in the intermediate surfaces by applying EM algorithm at every level. This smoothing process not only reduces the overall gradient of the surface but also reduces the number of local maxima. This is an effective optimization procedure that eliminates extensive search in the non-promising areas of the parameter space. Benchmark results demonstrate a significant improvement of the proposed algorithm compared to other existing methods.

One can apply TRUST-TECH method for tracing the optimal solutions across different smooth levels. Applying an EM algorithm might not guarantee to trace the optimal solution at each level. Especially, if the smoothing procedure distorts the surface to a considerable amount, it will be difficult to trace the optimal solution by merely applying the EM algorithm. Using TRUST-TECH, one can avoid this problem. Even if there is a considerable distortion in the surface, TRUST-TECH can help in searching the neighborhood regions effectively. In summary, we can use TRUST-TECH (instead of EM) to trace the optimal solutions.
\newpage
\section*{APPENDIX-A: Convolution of two Gaussians} \label{sec:appendix-A}

\begin{proof}
Lets consider two gaussian density functions with parameters
$\theta_1$ and $\theta_0$.

\begin{eqnarray}
P(x|\theta_1) =\frac{1}{ \sqrt{2\pi}\sigma_1}e^{-\frac{(x-\mu_1)^2}{2\sigma_1^2}}\\
P(x|\theta_0) =\frac{1}{\sqrt{2\pi}\sigma_0 }e^{-\frac{(x-\mu_0)^2}{2\sigma_0^2}}\\
\end{eqnarray}

By definition of convolution, we have

\begin{equation}\label{eq:conv1}
    g(t) \otimes h(t) =\int_{-\infty}^{\infty} g(\tau)h(t-\tau)d\tau
\end{equation}
\begin{eqnarray}
\label{eq:conv2}
    c(x|\theta_1,\theta_0)=p(x|\theta_1) \otimes p(x|\theta_0)\\ = \int_{-\infty}^{\infty} ~\frac{1}{ \sqrt{2\pi}\sigma_1}e^{-\frac{((x-\tau)-\mu_1)^2}{2\sigma_1^2}} ~\frac{1}{\sqrt{2\pi}\sigma_0
}e^{-\frac{(\tau-\mu_0)^2}{2\sigma_0^2}} d\tau\\
\\  = ~\frac{1}{ \sqrt{2\pi}\sigma_1}~\frac{1}{\sqrt{2\pi}\sigma_0} \int_{-\infty}^{\infty} e^{-\frac{\sigma_0^2(\tau-x+\mu_1)^2+\sigma_1^2(\tau-\mu_0)^2}{2\sigma_1^2\sigma_0^2}} d\tau
\end{eqnarray}

After rearranging the terms that are independent of $\tau$ and
further simplification, we get

\begin{equation}\label{eq:conv21}
  c(x|\theta_1,\theta_0)=  \frac{e^{-\frac{(x-(\mu_1+\mu_0))^2}{2(\sigma_1^2+\sigma_0^2)}}}{\sqrt{2\pi}\sigma_1~\sqrt{2\pi}\sigma_0}
\int_{-\infty}^{\infty}
e^{-\frac{(\tau-\mu_{\tau})^2}{2\sigma_{\tau}^2}} d\tau
\end{equation}

where
\begin{equation}\label{eq:conv3}
\mu_{\tau}=\frac{\mu_0\sigma_1^2+(x-\mu_1)\sigma_0^2}{(\sigma_1^2+\sigma_0^2)}\\
\sigma_{\tau}=\frac{\sigma_1^2\sigma_0^2}{(\sigma_1^2+\sigma_0^2)}
\end{equation}

Hence, we have

\begin{equation}\label{eq:conv4}
\begin{split}
  c(x|\theta_1,\theta_0)&=\frac{e^{-\frac{(x-(\mu_1+\mu_0))^2}{2(\sigma_1^2+\sigma_0^2)}}}{\sqrt{2\pi(\sigma_1^2+\sigma_0^2)}}
\int_{-\infty}^{\infty}
\frac{1}{\sqrt{2\pi}~\sigma_{\tau}}~e^{-\frac{(\tau-\mu_{\tau})^2}{2\sigma_{\tau}^2}}
d\tau
\\ &\quad =\frac{1}{\sqrt{2\pi
(\sigma_1^2+\sigma_0^2)}}e^{-\frac{(x-(\mu_1+\mu_0))^2}{2(\sigma_1^2+\sigma_0^2)}}
\end{split}
\end{equation}
(because the quantity inside the integral is 1 for a Gaussian
density function.)
\end{proof}

\newpage
\section*{APPENDIX-B: Proof of Theorem \ref{th:convmu}} \label{sec:appendix-B}
\begin{proof}
Convolution of Gaussian density with respect to the mean is shown
below:

\begin{equation}\label{eq:conv5}
    \tilde{c}(x,\theta_0)= \int_{-\infty}^{\infty} ~\frac{1}{ \sqrt{2\pi}\sigma_1}e^{-\frac{(x-(\mu_1-\tau)^2}{2\sigma_1^2}} ~\frac{1}{\sqrt{2\pi}\sigma_0
}e^{-\frac{\tau^2}{2\sigma_0^2}} d\tau\\
\end{equation}

Now, consider Eq. (\ref{eq:conv2}). Substituting  $\mu_0=0$ and
replacing $\tau$ with $-\tau$, we get
\begin{equation}\label{eq:conv6}
    c(x,\theta_0)= \int_{-\infty}^{\infty} ~\frac{1}{ \sqrt{2\pi}\sigma_1}e^{-\frac{(x+\tau-\mu_1)^2}{2\sigma_1^2}} ~\frac{1}{\sqrt{2\pi}\sigma_0
}e^{-\frac{\tau^2}{2\sigma_0^2}} d\tau\\ 
\end{equation}

From Eqs. (\ref{eq:conv5}) and (\ref{eq:conv6}), we can see that
convolution of a Gaussian density function with a Gaussian density
with zero mean is equivalent to convolving the function with respect
to mean.\end{proof}

\newpage
\section*{APPENDIX-C: Derivations for EM updates} \label{sec:appendix-C}

For simplicity, we show the derivations for EM updates in the fixed
kernel case. Lets consider the case where a fixed Gaussian kernel
with parameters $\mu_0$ and $\sigma_0$ which will be used to
convolve each component of the GMM. We know that

\begin{equation}\label{eq:log1}
    log ~p(\mathcal{X},\mathcal{Z}|\Theta)=log \prod_{j=1}^n ~p(x^{(j)}|z^{(j)},\Theta)~\cdot~p(z^{(j)})\\
\end{equation}
For the $j^{th}$ data point, we have
\begin{equation}\label{eq:log2}
    p(x^{(j)}|z^{(j)},\Theta)~\cdot~p(z^{(j)})\\  =\prod_{i=1}^k   \left[ \frac{1}{\sqrt{2\pi
(\sigma_i^2+\sigma_0^2)}}e^{-\frac{(x-(\mu_i+\mu_0))^2}{2(\sigma_i^2+\sigma_0^2)}} p(z_i^{(j)}=1)\right]^{z_i^{(j)}}\\
\end{equation}

Hence,
\begin{equation}\label{eq:finalsol}
\begin{split}
    log &~p(\mathcal{X},\mathcal{Z}|\Theta)=\sum_{j=1}^n log  ~p(x^{(j)}|z^{(j)},\Theta)\cdot
    p(z^{(j)})\\ &\quad
    = \sum_{j=1}^n\sum_{i=1}^k log\left[\frac{1}{\sqrt{2\pi
(\sigma_i^2+\sigma_0^2)}}e^{-\frac{(x-(\mu_i+\mu_0))^2}{2(\sigma_i^2+\sigma_0^2)}}
p(z_i^{(j)}=1) \right]^{z_i^{(j)}}\\ &\quad
=\sum_{j=1}^n\sum_{i=1}^k z_i^{(j)} [ -
log(\sqrt{2\pi(\sigma_i^2+\sigma_0^2)})-\frac{(x-(\mu_i+\mu_0))^2}{2(\sigma_i^2+\sigma_0^2)}
+log~ \alpha_i ]
\end{split}
\end{equation}
{\bf Expectation Step : } 
For this step, we need to compute the $Q$-function which is the
expected value of Eq.~(\ref{eq:finalsol}) with respect to the hidden
variables.
\begin{eqnarray}
\label{eq:qfunc}
\begin{split}
    Q(\Theta|\Theta^{(t)}) = E_z\left[log~p(\mathcal{X},\mathcal{Z}|\Theta)|\mathcal{X},\Theta^{(t)} \right] \\
    =\sum_{j=1}^n\sum_{i=1}^k E_z[z_i^{(j)}] [ -
log(\sqrt{2\pi(\sigma_i^2+\sigma_0^2)})\\~~~~~~~~~~~~~~~~~~~~~~~~~~~~~-\frac{(x-(\mu_i+\mu_0))^2}{2(\sigma_i^2+\sigma_0^2)}
+log~ \alpha_i ]
\end{split}
\end{eqnarray}
To compute the Expected value of the hidden variables $(w_i^{(j)})$,
\begin{eqnarray}
\label{eq:wij}
\begin{split}
    w_i^{(j)}=E_z[z_i^{(j)}]=\sum_{c=0}^1 c*p(z_i^{(j)}=c|\Theta^{(t)},x^{(j)})  \\
    =\frac{p(x^{(j)}|\Theta^{(t)},z_i^{(j)}=1)~p(z_i^{(j)}=1|\Theta^{(t)})}{p(x^{(j)}|\Theta^{(t)})}\\
=\frac{\frac{1}{\sqrt{(\sigma_i^2+\sigma_0^2)}}e^{-\frac{(x-(\mu_i+\mu_0))^2}{2(\sigma_i^2+\sigma_0^2)}}\alpha_i^{(t)}}{\sum_{m=1}^k
\frac{1}{\sqrt{
(\sigma_m^2+\sigma_0^2)}}e^{-\frac{(x-(\mu_m+\mu_0))^2}{2(\sigma_m^2+\sigma_0^2)}\alpha_m^{(t)}}
}
\end{split}
\end{eqnarray}

\noindent {\bf Maximization Step : }The maximization step is given by the following equation :
\begin{equation}\label{eq:max1}
    \frac{\partial }{\partial \Theta_i} Q(\Theta|\widehat{\Theta}(t))= 0
\end{equation}
where $\Theta_i$ are the parameters of the $i^{th}$ component. Due
to the assumption made that each data point comes from a single
component, solving the above equation becomes trivial. The updates
for the maximization step in the case of GMMs are given as follows :

\begin{eqnarray}
(\mu_i+\mu_0) = \frac{\sum_{j=1}^{n}w_i^{(j)}x^{(j)}}{\sum_{j=1}^{n}w_i^{(j)}}\\
(\sigma_i^2+\sigma_0^2) = \frac{\sum_{j=1}^{n}w_i^{(j)} (x^{(j)}-(\mu_i+\mu_0))^2}{\sum_{j=1}^{n}w_i^{(j)}}\\
\alpha_i=\frac{1}{n}\sum_{j=1}^{n}w_i^{(j)} \label{eq:update1}
\end{eqnarray}

%
%
%
%

\chapter {TRUST-TECH based Neural Network Training}
\label{ch:training}

Supervised learning using artificial neural networks has numerous
applications in various domains of science and engineering.
Efficient training mechanisms in a neural network play a vital role
in deciding the network architecture and the accuracy of the
classifier. Most popular training algorithms tend to be greedy and
hence get stuck at the nearest local minimum of the error surface.
To overcome this problem, some global methods (like multiple
restarts, genetic algorithms, simulated annealing etc.) for
efficient training make use of stochastic approaches in combination
with local methods to obtain an effective set of training
parameters. Due to the stochastic nature and lack of effective
{\it fine tuning} capability, these algorithms often fail to obtain an
optimal set of training parameters. In this chapter, a new method to
improve the local search capability of training algorithms is
proposed. This new method takes advantage of TRUST-TECH to compute neighborhood local minimum of the error
surface. The proposed approach obtains multiple local optimal
solutions surrounding the current local optimal solution in a
systematic manner. Empirical results on different machine learning
datasets indicate that the proposed algorithm outperforms current
algorithms available in the literature.


\section{Overview}
\begin{figure}[htp]
\centerline{
  \epsfig{figure=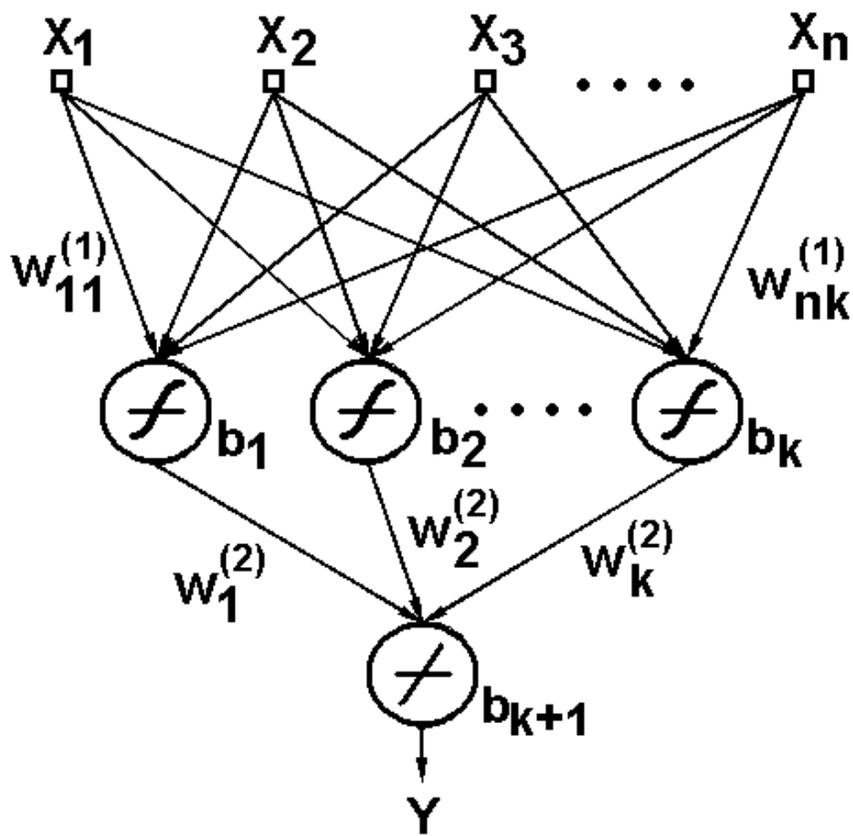, width=4.5in}
} \caption{The Architecture of MLP with single hidden layer having k
hidden nodes and the output layer having a single output node.
$x_1,x_2,...,x_n$ is an n-dimensional input feature vector. $w_{ij}$ are the
weights and $b_1,b_2,...,b_k$ are the biases for these $k$ nodes.
The activation function for the hidden nodes is sigmoidal and for
the output node is linear.} \label{fig:network}
\end{figure}

Artificial neural networks (ANN) were developed analogous to the
human brain for the purpose of improving conventional learning
capabilities. They are used for a wide variety of applications in
diverse areas such as function approximation, time-series prediction,
medical diagnosis, character recognition, load forecasting, speaker
identification and risk management. These networks serve as
excellent approximators of nonlinear continuous functions
\cite{Leung03}. However, using an artificial neural network to model
a system usually involves dealing with certain difficulties in achieving
the best representation of the classification problem.

The two challenging tasks in the process of learning using ANNs are
network architecture selection and optimal training. In deciding the
architecture for the feedforward neural network (also known as Multi-Layer
Perceptron, MLP), a larger network will always provide better
prediction accuracy for the data available. However,
such a large network that is too complicated and customized to some
given problem will lose its generalization capability for the unseen
data \cite{Branke95}. Also, every additional neuron translates to increased hardware
cost. Hence, it is vital to develop algorithms that can exploit the
potential of a given architecture which can be achieved by obtaining
the global minimum of the error on the training data. Hence, the
goal of optimal training of the network is to find a set of weights
that achieves the global minimum of the mean square error (MSE) \cite{Haykin99}.
Fig.~\ref{fig:network} shows the architecture of a single hidden
layer neural network with $n$ input nodes, $k$ hidden nodes and $1$
output node. The network is trained to deliver the output value
($Y_i$) for the $i^{th}$ sample at the output node which will be
compared to the actual target value~($t_i$).

\begin{figure*}[htp]
   \centering
   \subfigure[]{\includegraphics[width = 2.2 in]{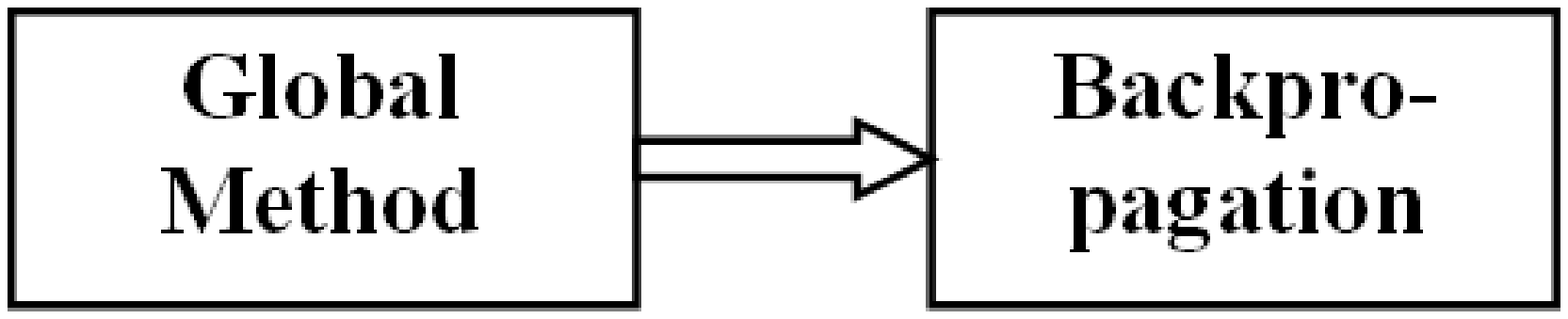}}\qquad
   \subfigure[]{\includegraphics[width = 3.2 in]{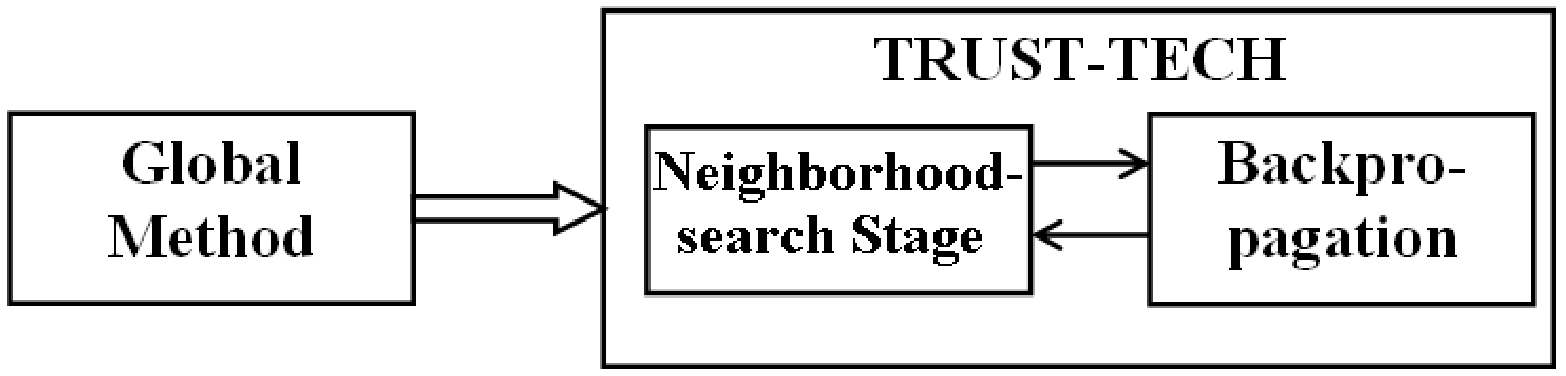}}\qquad
   \caption{\label{fig:comp}Comparison between the two frameworks
   (a) Traditional approach and (b) TRUST-TECH based approach. The main difference is the inclusion of the stability region based neighborhood-search stage that can explore the neighborhood solutions.
   }
 \end{figure*}
The main focus of this work is to develop a robust training
algorithm for obtaining the optimal set of weights of an artificial
neural network. Several training algorithms have been extensively
studied in the literature \cite{Haykin99}. Backpropagation (BP)
algorithm is a very robust deterministic local method that have
received significant attention. Though BP is comparatively cheaper
in terms of time and easy to implement, it can only obtain local
optimal solutions. On the contrary, some global methods like multiple random starts, genetic algorithms and simulated annealing can identify promising regions of the weight space, but are essentially stochastic in nature and computationally expensive. Expecting such stochastic algorithms to fine-tune the training weights will be even more time
consuming. Thus, there is a necessity to efficiently search for good
solutions in promising regions of the solution space, which can be
accomplished by the newly proposed TRUST-TECH based algorithm. In
this chapter, we introduce a novel algorithm that will search the
weight space in a systematic tier-by-tier manner. Fig.~\ref{fig:comp} compares the
traditional approach with our proposed approach. The main difference
between the two approaches is the inclusion of the neighborhood-search stage, where an improved
set of training weights are obtained by systematically exploring the
neighborhood local optimal solutions in a promising subspace.

\section{Relevant Background}
\label{sec:background}

The performance of a feedforward neural network is usually gauged by measuring the MSE of its outputs from the expected target values. The goal of optimal training is to find a set of parameters that
achieves the global minimum of the MSE
\cite{Bishop95,Shang96,Leung03}. For a $n$-dimensional dataset, the
MSE over $Q$ samples in the training set is given as follows :

\begin{equation}\label{eq:MSE}
    C(W) =\frac{1}{Q}\sum_{i=1}^Q \left[t(i)-y(X,W)\right]^2\\
\end{equation}

\noindent where $t(i)$ is the target output for the $i^{th}$ sample, $X$ is the input vector and $W$ is
the weight vector. The MSE as a function of the weight parameters is highly nonlinear containing several local minima. The
network's weights and thresholds must be set so as to minimize the
prediction error made by the network. Since it is not possible to
analytically determine the global minimum of the error surface, the
neural network training is essentially an exploration of the error
surface for an optimal set of parameters that attains this globally
optimal solution.

Training algorithms can be broadly classified into `{\it local}' and
`{\it global}' methods. Local methods begin at some initial points
and deterministically move towards a local minimum. From an
initial random configuration of weights and thresholds, these local
training methods incrementally(greedily) seek for improved solution until they
reach a local minimum. Typically, some form of the gradient information
at the current point on the error surface is calculated and used to
make a downhill move. Eventually, the algorithm stops at a low
point, which usually is a local minimum. In the context of training neural networks, this local minima problem
is a well-studied research topic \cite{Gori92}. The most commonly used training method in MLP is the
backpropagation algorithm \cite{Rumelhart86} which has been tested
successfully for different kinds of problems. Despite having many variants, BP faces the problem of stopping at local
minimum instead of proceeding towards the global minimum
\cite{Hinton92,Shang96}. Modifications \cite{Lehmen88} to the basic
BP model have been suggested to help the algorithm escape from being
trapped in a local minimum. However, while these improved methods
reduce the tendency to sink into local minimum by providing some
form of perturbations to the search direction, it does not train the
network to converge to a global minimum within a reasonable number
of iterations \cite{Ingman91, Wang99}. Based on the movement towards
improved solutions, local methods can be subdivided into two
categories:

\begin {enumerate}
\item{{\bf Line search methods: } These algorithms select some descent direction (based on the
gradient information) and minimize the error function value along this particular
direction. This process is repeated until a local minimum is
reached. Most popular choices for the descent directions are
Newton's direction or conjugate direction. In the context of neural
networks, apart from the obvious steepest descent methods, other
widely used line search algorithms are Newton's method
\cite{battiti92}, the BFGS method \cite {Osowski96} and conjugate
gradient methods~\cite{Charalambous92,Moller93}. }
\item {{\bf Trust region methods:} Trust
region methods are by far the fastest convergent methods compared to
the above mentioned line-search methods. The surface is assumed to
be a simple model (like a parabola) such that the minimum can be
located directly if the model assumption is good which usually
happens when the initial guess is close to the local minimum. They
require more storage space compared to conjugate gradient methods
\cite{Hagan94} and hence are not effective for large-scale
applications. }
\end{enumerate}

\begin{figure*}[htp]
\centerline{
  \epsfig{figure=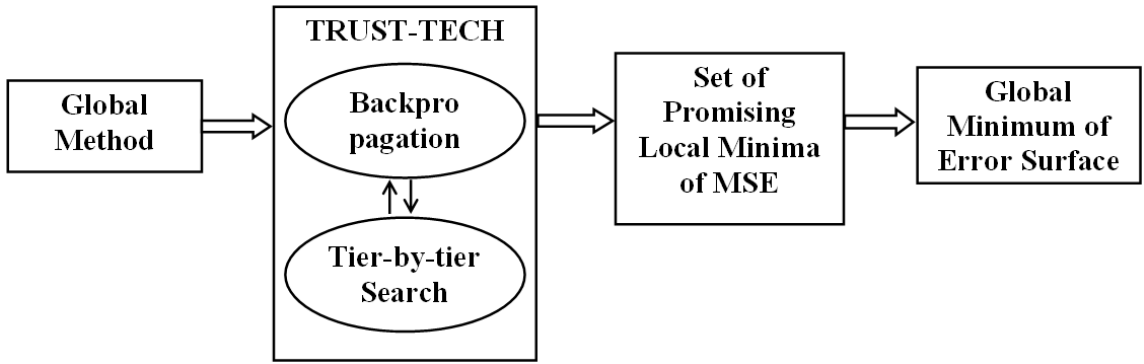, width=6.0in}
} \caption{Block Diagram of the TRUST-TECH based training method.}
\label{fig:block}
\end{figure*}

All these local methods discussed so far assume that they already
have an initial guess to begin with. Usually the quality of the
final solution depends significantly on the initial set of
parameters available. Hence, in practice, none of these local methods are used by themselves. They are usually combined with stochastic
global methods which yield a promising set of parameters in the
weight space. These global methods explore the entire error surface
and thus the chance of attaining a near-global optimal solution is
high. More advanced techniques like Genetic algorithms
\cite{Branke95} and simulated annealing \cite{Amato91} are applied
in combination with standard BP in order to obtain more promising
solutions and avoid being stuck at local minimum \cite{Reeves93}. The
use of various global optimization algorithms for finding an
effective set of training parameters is comprehensively given in
\cite{Shang96}. Although these methods (asymptotically) guarantee
convergence to the global minimum, it usually exhibits very slow
convergence even for simple learning tasks. Though methods can
explore the entire solution space effectively and obtain promising
local optimal solutions, it lacks fine-tuning capabilities to obtain
a precise final solution and requires local methods like BP to be
employed. Other traditional methods like Monte Carlo method, Tabu
search, ant colony optimization and particle swarm optimization are
also stochastic in nature and suffer from the same problems
described above.

From the above discussion, one can realize that there
is a clear gap between global and local methods. Typically, most of the successful
practical algorithms are a combination of these global and local
methods. In other words, these two approaches do not communicate
well between each other. Approaches that might resemble our
methodology are TRUST \cite{Cetin93} and dynamic tunneling
\cite{RoyChowdhury99}. These methods attempt to move out of the
local minimum in a stochastic manner. The training algorithm
proposed in this chapter differs from these two methods by
deterministically escaping out of the local minimum and
systematically exploring multiple local minima on the error surface
in a tier-by-tier manner in order to advance towards the global
minimum. This approach is based on the fundamental concepts of
stability regions that were established in \cite{Chiang96,Lee04}.
Fig. \ref{fig:block} shows the block diagram of the TRUST-TECH
methodology. Basically, a global method yields points in
certain promising regions of the search space. These points
are used as initial guesses to search the neighborhood subspace in a systematic manner.
TRUST-TECH relies on a robust, fast local method to obtain a local
optimal solution. It explores the parameter subspace in a  tier-by-tier manner
by transforming the function into its corresponding dynamical system
and exploring the neighboring stability regions. Thus, it gives a
set of promising local optimal solutions from which a global minimum
is selected. In this manner, TRUST-TECH can be treated as an
effective interface between the global and local methods, which
enables the communication between these two methods. It also allows
the flexibility of choosing different global and local methods
depending on their availability and performance for certain specific
classification tasks.

\section{Training Neural Networks}
\label{sec:training}

Without loss of generality, we consider a feedforward neural network
with one input layer, one hidden layer and one output layer.
Specifically, the output layer contains only one node that will
yield all the possible target values depending on its activation
function. Table \ref{TB:nota} gives the notations used in the rest of this chapter.
\begin{table}[h]
\centering \caption{\protect Description of the notations
used}
\begin{center}
\begin{tabular}{cl}
\hline  Notaion &   Description \\
 \hline

 Q & Number of training samples\\
 X & Input vector\\
 W & Weight vector\\
 n & Number of features\\
 k & Number of hidden nodes\\
 $w_{0j}$ & weight between the output node and the $j^{th}$ hidden node \\
 $w_{ij}$ & weight between the $i^{th}$ input node and the $j^{th}$ hidden node \\
 $b_0$ & bias of the output node \\
 $b_j$ & bias of the $j^{th}$ hidden node \\
 $\phi_1$ & Activation function of the hidden nodes\\
 $\phi_2$ & Activation function of the output node\\
 $t_i$ & target value of the $i^{th}$ input sample\\
 y & output of the network\\
 $e_i$ & Error for the $i^{th}$ input sample\\

 \hline

\end{tabular}
\end{center}
\label{TB:nota}
\end{table}

Let $k$ be the number of hidden nodes in the hidden layer and the
input vector is $n$-dimensional. Then the final nonlinear mapping
of our model is given by :

\begin{equation}\label{eq:MSE}
    y(W,X) =\phi_2\left( \sum_{j=1}^k w_{0j} \phi_1\left(\sum_{i=1}^n w_{ij}x_i+b_{j}\right)+b_{0}\right)
\end{equation}

where $\phi_1$ and $\phi_2$ are the activation functions of the
hidden nodes and the output nodes respectively. $\phi_1$ and
$\phi_2$ can be same functions or can be different functions. We
have chosen to use $\phi_1$ to be sigmoidal and $\phi_2$ to be
linear. Results in the literature \cite{Gorse97}, suggest that this
set of activation functions yield the best results for feedforward
neural networks. As shown in Fig.~\ref{fig:network}, $w_{0j}$
indicate the weights between the hidden layer and the output layer
and $w_{ij}$ indicate the weights between the input layer and the
hidden layer. $b_j$ are the biases of the $k$ hidden nodes and
$b_{0}$ is the bias of the output node. $x_i$ is the $n$-dimensional
input feature vector and $X_i$ indicates the $i^{th}$ training
sample. The task of the network is to learn associations between the
input-output pairs {$(X_1,t_1),(X_2,t_2),...,(X_Q,t_Q)$}. The weight vector to be optimized is constructed as follows:

\begin{equation*}\label{eq:Wvec}
{\scriptstyle
W=}(w_{01},w_{02},..,w_{0k},..,w_{n1},w_{n2},..,w_{nk},b_0,b_1,b_2..,b_k)^T
\end{equation*}

which includes all the weights and biases that are to be computed.
Hence, the problem of training neural networks is $s$-dimensional
unconstrained minimization problem where $s=(n+2)k+1$.
\begin{equation}\label{eq:minJW}
    \min_W C(\scriptstyle W)
\end{equation}

The mean squared error which is to be minimized can be written as

\begin{equation}\label{eq:JW}
    C({\scriptstyle W}) = \frac{1}{Q} \sum_{i=1}^{Q} e_i^2({\scriptstyle
  W})
\end{equation}
where the error
\begin{equation}\label{eq:ei}
    e_i({\scriptstyle W}) = t_i-y({\scriptstyle W},{\scriptstyle X_i})
\end{equation}

The error cost function $C(\cdot)$ averaged over all training data
is a highly nonlinear function of the synaptic vector
${\scriptstyle W}$. Ignoring the constant for simplicity, it can be
shown that
\begin{center}
\begin{eqnarray}
    \nabla C({\scriptstyle W}) = J^T({\scriptstyle W})e({\scriptstyle
    W})\\
\nabla^2 C({\scriptstyle W}) = J^T({\scriptstyle
W})J({\scriptstyle W})+S({\scriptstyle W})
\label{eq:firstderivative}
\end{eqnarray}
\end{center}

where $J({\scriptstyle W})$ is the Jacobian matrix

\[ J({\scriptstyle W}) = ~ \left[ \begin{array}{ccccc} \frac{\partial
e_1}{\partial {\scriptstyle W_1}} & \frac{\partial e_1}{\partial
{\scriptstyle W_2}} &.&.& \frac{\partial e_1}{\partial
{\scriptstyle W_N}}\vspace{0.05in}
\\\frac{\partial
e_2}{\partial {\scriptstyle W_1}} & \frac{\partial e_2}{\partial
{\scriptstyle W_2}} &.&.& \frac{\partial e_2}{\partial
{\scriptstyle W_N}}\vspace{0.05in}
\\ .&.&. &    &.\\
.&.& & .   &.\vspace{0.05in}\\
\frac{\partial e_Q}{\partial {\scriptstyle W_1}} & \frac{\partial
e_Q}{\partial {\scriptstyle W_2}} &.&.& \frac{\partial
e_Q}{\partial {\scriptstyle W_N}}\vspace{0.05in}

\end{array} \right]\]

and
\begin{equation}\label{eq:secondderivative}
    S({\scriptstyle W}) = \sum_{i=1}^{Q} e_i({\scriptstyle
  W}) \nabla^2 e_i({\scriptstyle W})\end{equation}

Generally, if we would like to minimize $J({\scriptstyle W})$ with
respect to the parameter vector ${\scriptstyle W}$, any variation
of Newton's method can be written as

\begin{equation}\label{eq:firstderivative}
\begin{split}
    \Delta {\scriptstyle W} &= -\left[\nabla^2 C({\scriptstyle W})\right]^{-1}~\nabla C({\scriptstyle W}) \\
    &=-\left[J^T({\scriptstyle W})J({\scriptstyle W})+S({\scriptstyle W})\right]^{-1} J^T({\scriptstyle W})e({\scriptstyle W}) \\
\end{split}
\end{equation}

\section{Problem Transformation}
\label{sec:trans} We explore the geometrical structure of the error surface to explore multiple local
optimal solutions in a systematic manner. Firstly, we describe the
transformation of the original minimization problem into its corresponding
nonlinear dynamical system and then propose a new TRUST-TECH based training algorithm for
finding multiple local optimal solutions.

This section mainly deals with the transformation of the original
error function into its corresponding nonlinear dynamical
system and introduces some terminology pertinent to comprehend our
algorithm. This transformation gives the correspondence between all
the critical points of the error surface and that of its
corresponding gradient system. To analyze the geometric structure of
the error surface, we build a {\it generalized gradient system}
described by
\begin{equation}\label{eq:dwdt}
    \frac{d{\scriptstyle W}}{dt} = -A({\scriptstyle W})\nabla C({\scriptstyle W})
\end{equation}

where the error function $C$ is assumed to be twice differentiable
to guarantee unique solution for each initial condition
${\scriptstyle W}(0)$ and $A({\scriptstyle W})$ is a positive
definite symmetric matrix for all ${\scriptstyle W} \in \Re^n$. It is interesting to note the
relationship between Eqs. (\ref{eq:dwdt}) and (\ref{eq:firstderivative})
and obtain different local solving methods used to find the nearest
local optimal solution with guaranteed convergence. For example, if
$A({\scriptstyle W})=I$, then it is a naive error back-propagation
algorithm. If $A({\scriptstyle W})=[J({\scriptstyle W})^T
J({\scriptstyle W})]$ then it is the Gauss-Newton method and if
$A({\scriptstyle W})=[J({\scriptstyle W})^T J({\scriptstyle W})+\mu
I]$ then it is the Levenberg-Marquardt method. This transformation of the original error function to its corresponding dynamical system will enable us to transform the problem of finding multiple local minima on the error surface into the problem of finding multiple stable equilibrium points of its corresponding dynamical system. This will enable us to apply TRUST-TECH method for training neural networks and obtain promising solutions.

\section{TRUST-TECH based Training} \label{sec:algorithm} 
\begin{figure}[htp]
\centerline{
  \epsfig{figure=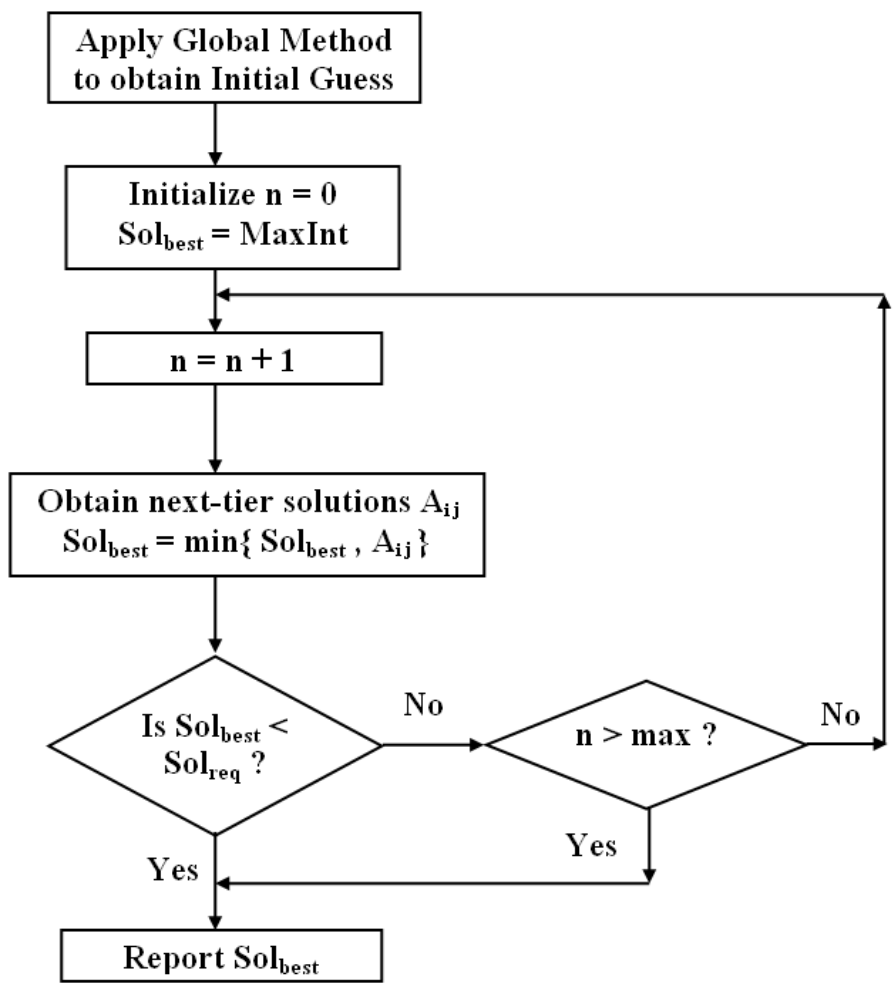, width=5.0in}
} \caption{Flow chart of our method} \label{fig:flowchart}
\end{figure}

The proposed TRUST-TECH based algorithm for training neural networks, uses a promising starting point ($A^*$) as input and outputs the best local minimum of the neighborhood in the weight space. Figure \ref{fig:flowchart} shows the flowchart of our approach.

\begin{figure}[htp]
\centerline{
  \epsfig{figure=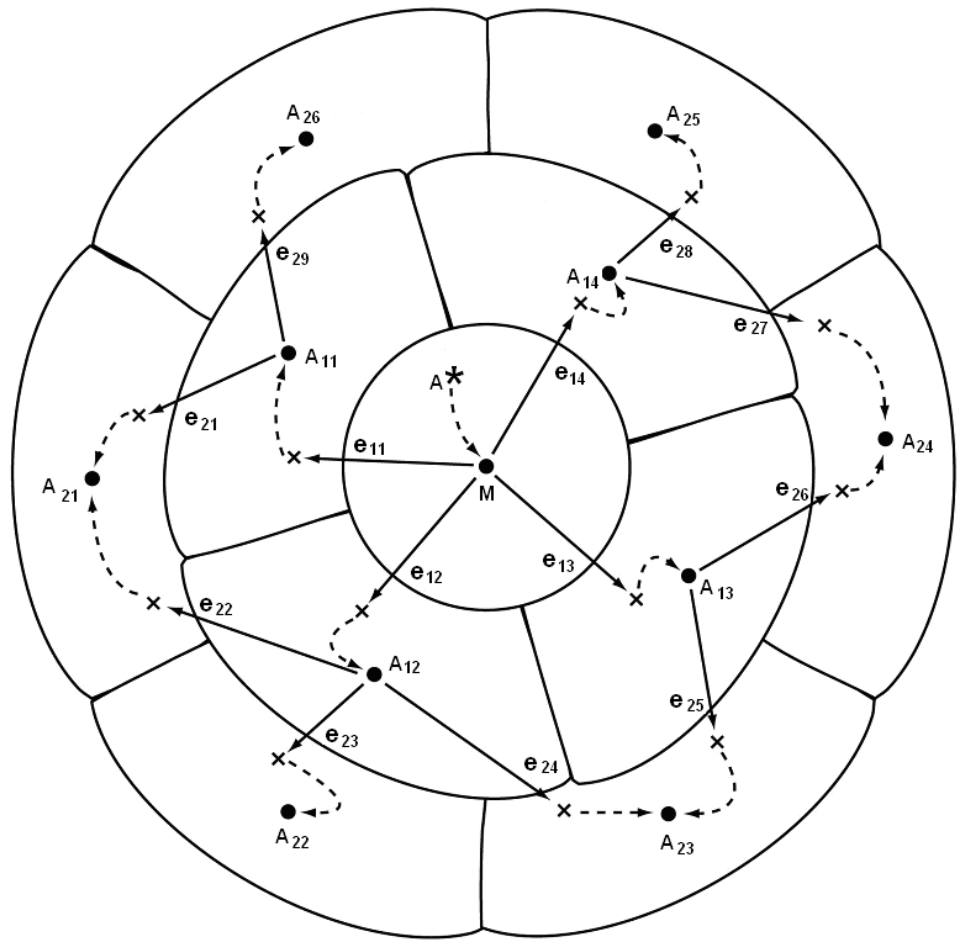, width=5.0in}
} \caption{Diagram Illustrating two tier exit point strategy. The
`A*' represents the initial guess. Dotted arrows represent the
convergence of the local solver. Solid arrows represent the gradient
ascent linear searches along eigenvector directions. `X' indicates a
new initial condition in the neighboring stability region. $M$
represents the local minimum obtained by applying local methods from
'X'. $A_{1i}$ indicates Tier-1 local minima. $e_{1i}$ are the exit
points between $M$ and $A_{1i}$. Similarly, $A_{2j}$ and $e_{2j}$
are the second-tier local minima and their corresponding exit points
respectively. } \label{fig:Diagram}
\end{figure}

\noindent \textbf{Input:} Initial guess($A^*$), Tolerance ($\tau$), Step size
($s$)
\\\textbf{Output:} Best local minimum ($A_{ij}$) in the neighborhood
\\\textbf{Algorithm:}\\
{\it Step 1: Obtaining good initial guess ($A^*$):} The initial
guess for the algorithm can be obtained from global search
methods or from a purely random start. Some domain knowledge about
the specific dataset that the network is being trained on, might help in
eliminating non-promising set of initial weights.
\\{\it Step 2: Moving to the local minimum ($M$):}
Using an appropriate local solver (such as conjugate-gradient, quasi-Newton or Levenberg-Marquardt), the local optimum $M$ is
obtained using $A^*$ as the initial guess. \\{\it Step 3: Determining the search
direction ($d_j$):} The eigenvectors $d_j$ of the Jacobian are
computed at $m_i$. These eigenvector directions might lead to
promising regions of the subspace. Other search directions can also be chosen based on the specific problem that is being dealt. \\
{\it Step 4: Escaping from the local minimum: } Taking small step
sizes away from $m_i$ along the $d_j$ directions increases the objective
function value till it hits the stability boundary. However, the
objective function value then decreases after the search trajectory
moves away from the exit point. This new point is used as
initial guess and local solver is applied again (go to Step 2).
\\ {\it Step 5: Finding Tier-1 local minima ($A_{1i}$):} Exploring the neighborhood of the local optimal solution corresponding to
the initial guess leads to tier-1 local minima. Exploring from tier-$k$ local minima leads to tier-$k+1$ local minima. \\
{\it Step 6: Exploring Tier-$k$ local minima ($A_{kj}$):} Explore all other tiers
in the similar manner described above (see Fig.~\ref{fig:Diagram}). From all these solutions, the best one is chosen to be the desired global optimum.\\
{\it Step 7: Termination Criteria:} The procedure can be terminated
when the best solution obtained so far is satisfactory (lesser than
$sol_{req}$) or a predefined maximum number of tiers is explored.

\section{Implementation Details}
\label{sec:implementation} All programs were implemented in MATLAB
v6.5 and run on Pentium IV 2.8 GHz machines. This section
describes the various implementation details used in our
simulations. The following issues are discussed in detail : (i)
Architecture and local methods, (ii) Initialization Schemes and
(iii) TRUST-TECH.

\subsection{Architecture and Local Methods}
As described in the introduction section, we have chosen to
demonstrate the capability of our newly proposed TRUST-TECH algorithm on a
network with single hidden layer and an output layer containing only one
output node. This architecture is not complicated and has the
capability to precisely demonstrate the problems with the existing
approaches. A network described here contains $n$ (number of attributes)
input nodes which is equal to the number of features available in
the dataset, one hidden layer with $k$ nodes and one output node.
Thus, each network has $nk$ weights and $k$ biases to the hidden
layer, and $k$ weights and one bias to the output node. Hence,
training a neural network is necessarily a search
problem of dimensionality $(n+2)k + 1$. Each hidden node has a
tangent-sigmoid transfer function and the output node has a pure
linear transfer function. The number of nodes in the hidden layer is
determined by incrementally adding hidden nodes, and selecting the architecture that achieves a compromise between minimal error value and minimal number of nodes. The trust region based Levenberg-Marquardt
algorithm is chosen because of efficiency in terms of time and space
consumption. It utilizes the approximation of the Jacobian in its
iterative gradient descent, which will be used for generating
promising directions in TRUST-TECH.
\subsection{Initialization Schemes}  Two different initialization
schemes were implemented. The most basic global method which is
multiple random starts with initial set of parameters between -1 and
1. More effective global method namely Nguyen-Widrow (NW) algorithm
\cite{Nguyen90} has also been used to test the performance of our
algorithm. The NW algorithm is implemented as the standard
initialization procedure in MATLAB. In both cases, the best initial set of parameters in terms of
training error is chosen and improved with our TRUST-TECH algorithm.

\begin{algorithm}
\caption{$New\_Wts~$ $TRUST\_TECH$($NET, Wts,s,\tau$)}
\label{trustech}
\begin{algorithmic}
\STATE $Wts=Train(NET,Wts,\tau)$ \STATE$Error=Estimate(NET, Wts)$

    \STATE $Thresh=c*Error$
\STATE$Wts1[~]=Neighbors(NET,Wts,s,\tau)$ \FOR{$k=1$ to
$size(Wts1)$}

\IF{$Estimate(NET,Wts1[k])<Thresh$} \STATE
$Wts2[k][~]=Neighbors(NET,Wts1,s,\tau)$ \ENDIF

\ENDFOR \STATE \bf Return $best(Wts,Wts1,Wts2)$
\end{algorithmic}
\end{algorithm}

\subsection{TRUST-TECH} It is effective to apply the TRUST-TECH methodology to those promising solutions obtained from
stochastic global methods. Algorithm~\ref{trustech} describes the two-tier TRUST-TECH algorithm. $NET$
assumes to have a fixed architecture with a single output node. $s$
is the step size used for evaluating the objective function value till it obtains an exit point. $\tau$ is the tolerance of error used for the convergence of the local method. $Weights$ give the initial set of
weight parameter values. $Train$ function implements the
Levenberg-Marquardt method that obtains the local optimal solution
from the initial condition. The procedure $Estimate$ computes the MSE value of the network model. A threshold
value ($Thresh$) is set based on this MSE value. The procedure
$Neighbors$ returns all the next tier local optimal solutions from a
given solution. After obtaining all the tier-1 solutions,
the procedure $Neighbors$ is again invoked (only for promising solutions) to obtain
the second-tier solutions. The algorithm finally compares the
initial solution, tier-1 and tier-2 solutions and returns the
network corresponding to the lowest error amongst all these solutions.

\begin{algorithm}
\caption{$Wts[~]$  $Neighbors~$($NET, Wts$,~s,~$\tau$)}
\label{Neighbors}
\begin{algorithmic}

    \STATE $[Wts,Hess]=Train(NET,Wts,\tau)$
\STATE $evec=Eig\_Vec (Hess)$ \STATE $Wts[~]=NULL$

\FOR{$k=1$ to $size(evec)$} \STATE $Old\_Wts= Wts$ \STATE $
ext\_Pt=Find\_Ext(NET,Old\_Wts,s,evec[k])$

\IF{$(ext\_Pt)$} \STATE $New\_Wts=Move(NET,Old\_Wts,evec[k])$

    \STATE $New\_Wts=Train(NET,New\_Wts,\tau)$
    \STATE $Errors=Estimate(NET,New\_Wts)$
\STATE $Wts[~]=Append(Wts[~],New\_Wts,Errors)$

\ENDIF \ENDFOR \STATE \bf Return $Wts[~]$
\end{algorithmic}
\end{algorithm}

The approximate Hessian matrix obtained during the updation in the
Levenberg-Marquardt method used for computing the search
direction. Since there is no optimal way of obtaining the search
directions, the Eigen vectors of this Hessian matrix are used as
search directions. Along each search direction, the exit point is
obtained by evaluating the function value along that particular
direction. The step size for evaluation is chosen to be the average
step size taken during the convergence of the local procedure. The
function value increases initially and then starts to reduce
indicating the presence of exit point on the stability boundary.
$Move$ function ensures that a new point (obtained from the exit
point) is located in a different (neighboring) stability region.
From this new initial guess, the local method is applied again to
obtain a new local optimal solution. For certain directions, an exit point might not be encountered. For
these directions, the search for exit points will be stopped after
evaluating the function for certain number of steps. This avoids
inefficient use of resources required to search in non-promising
directions.

\section{Experimental Results}
\label{sec:results}
\subsection{Benchmark Datasets}
The newly proposed training method is evaluated using seven
benchmark datasets taken from the UCI machine learning repository
available at \cite{Blake98}. Since the main focus of our work is
the development of TRUST-TECH based training algorithm, only simple experiments were
conducted for choosing the architecture of the neural network.
The hidden nodes in the hidden layer are added incrementally and the
training error is computed. The final architecture is chosen with a fixed number of hidden nodes after which there is no significant improvement in the training error even when a hidden node is added. The number of nodes where the improvement
in the training error is not significant is chosen as the final
architecture. Table~\ref{TB:data1} summarizes the datasets. It gives
the number of samples, input features, output classes along with the
number of hidden nodes of the optimal architecture. These datasets
have varying degrees of complexity in terms of sample size, output
classes and the class overlaps. Here is the description of the datasets:

\begin{enumerate}

\item{\it Cancer : } This dataset contains data from cancer patients. It has 683 samples out of which 444 are benign cases and 239 are malignant cases. 9 attributes describing the tumor were used for classification.

\item{\it Diabetes :} This dataset gives information about patients who have some signs of diabetes according to World Health
Organization criteria. Each sample has 8 real valued attributes. A
total of 768 samples with 500 negative cases and 268 positive cases
are
available.

\item{\it Image : }This dataset contains images which were drawn
randomly from a database of 7 outdoor images. The images were hand-segmented to create a classification for every pixel. There are 19
attributes that describe each instance (which is a 3x3 region) of a
given image. The dataset contains a total of 2310 samples.

\item{\it Ionosphere : }This radar data was collected by a system
consisting a phased array of 16 high-frequency antennas with total
transmitted power on the order of 6.4 kilowatts. The targets were
free electrons in the ionosphere. The dataset consists of 351 samples with 34
attributes. The classification task here is to separate good
radar signals from that of the bad ones.

\item {\it Iris : }This dataset contains 3 classes of 50 samples
each, where each class refers to a type of iris plant. It is
relatively simple dataset where one class is linearly separable from
the other two, but the other two have significant overlap and are
not linearly separable from each other. The four attributes
considered for classification are sepal length, sepal width, petal
length and petal width. All
attributes are measured in centimeters.

\item {\it Sonar : } This dataset is used for the classification
of sonar signals. The task is to discriminate sonar signals
bounced off a metal cylinder from those bounced off a roughly
cylindrical rock. The dataset contains a total of 208 samples (111
for mines and 97 for rocks). The data set contains signals that were obtained
from a variety of different aspect angles, spanning 90 degrees for
the cylinder and 180 degrees for the rock. Each pattern is a set of
60 numbers in the range 0.0 to 1.0 that represents the energy within
a particular frequency band, integrated over a certain period of
time.

\item {\it Wine : }This dataset was obtained from the results of a chemical
analysis of wines derived from three different cultivars. The
analysis determined the quantities of 13 constituents found in each
of the three types of wines. A total of 178 samples with the
following distribution (59,71,48).
\end{enumerate}

\begin{table}[htp]
\centering \caption{\protect Summary of Benchmark Datasets.
}
\begin{center}
\begin{tabular}{|c|c|c|c|c|c|}
\hline  &   Sample &Input &   Output & Hidden&
Search \\
Dataset &   Size &Features &   Classes& Nodes&
Variables \\
$(\mathcal{D})$&(Q) &  (n) &   (p) & (H)&
(n+2)k+1\\
\hline

\hline Cancer & 683&9 &  2&    5 &  56 \\\hline Diabetes & 178&  8&
3 & 4& 61\\ \hline Image& 2310&19& 7 &   8 &169 \\ \hline
Ionosphere& 351&34 & 2& 9& 325 \\ \hline Iris  &150& 4 &  3&  3& 19 \\ \hline Sonar & 208& 60& 2&  8  & 497 \\
\hline Wine & 178& 13&  3 &   4& 61\\  \hline
\end{tabular}
\end{center}
\label{TB:data1}
\end{table}

\subsection{Error Estimation}
To demonstrate the generalization capability (and hence the
robustness) of the training algorithm, ten-fold cross validation is
performed on each dataset. This practice of cross validation
effectively removes any bias in the dataset segmentation. The use of
the validation dataset allows early stopping of the local method and
prevents over-fitting to a particular dataset. Essentially, each
dataset is partitioned into ten folds of approximately equal size.
Let these folds are denoted by $T_1,T_2,...T_{10}$. Each time, the
validation set will be $T_i$ in which the the target labels will be
deleted. The test set is $T_j$ for $j=(i+1) ~mod~ 10$. The
training set comprises of the rest of the dataset and is given by :

\begin{equation}\label{eq:JW}
\sum_{\substack{k=1\\k\neq i~k\neq j}}^{10} T_k\end{equation} The
final MSE is the average of all the errors obtained across each
of the ten folds. Usually, training error is much lower than the
test error because the network is modeled using the training
data and this data will be more accurately classified compared to the
unseen test data. All the network parameters including
the architecture and the set of weights are obtained using the training data.
Once the final model is fixed, the accuracy on the test data will
provide an estimate of the generalization capability of the network
model and the training algorithm.

\subsection{Classification Accuracy}
The criteria of evaluation is given by the classification accuracy
of the network model. The classification accuracy is given by the
following formula :

\begin{equation}\label{eq:classacc}
  \%\ accuracy = \frac{diff(~t(i),y(W,X)~)}{Q}*100
\end{equation}

\noindent where $diff$ gives the number of misclassified samples. Tables~\ref{TB:results} and~\ref{TB:results1} shows the improvements in the train error and the test error using TRUST-TECH methodology. For effective implementation, only the best five tier-1 and
corresponding tier-2 solutions were obtained using the TRUST-TECH strategy. For some of the datasets, there had been considerable
improvement in the classifier performance.

\begin{table*}[htp]
\centering \caption{\protect Percentage improvements in the
classification accuracies over the training and test data using
TRUST-TECH with multiple random restarts. }
\begin{center}
{\footnotesize
\begin{tabularx}{5.8in}{|X||c|c|c|c|c|c|}
\hline
 &\multicolumn{3}{c|}{Train Error}&\multicolumn{3}{c|}{Test Error}\\
\cline{2-7}
Dataset &MRS+BP&TRUST-&Gain&MRS+BP&TRUST-&Gain\\
&&TECH&&&TECH&\\
\hline\hline

Cancer & 2.21  &  1.74   & 27.01 &  3.95   & 2.63 &   50.19 \\
\hline Image &9.37 &8.04  &  16.54  & 11.08  & 9.74  &  13.76 \\
\hline Ionosphere &2.35 &0.57 &312.28 & 10.25 &  7.96 &   28.77 \\
\hline Iris &1.25 &1.00 &25.00 &3.33 &2.67 &24.72 \\ \hline Diabetes
& 22.04 &20.69 &6.52 &23.83 &20.58& 15.79 \\ \hline Sonar& 1.56&
0.72 &116.67& 19.17 &12.98& 47.69 \\ \hline Wine &4.56& 3.58& 27.37&
14.94 &6.73 &121.99 \\ \hline
\end{tabularx}
}
\end{center}
\label{TB:results}\end{table*}

\begin{table*}[htp]
\centering \caption{\protect Percentage improvements in the
classification accuracies over the training and test data using
TRUST-TECH with MATLAB initialization. }
\begin{center}
{\footnotesize
\begin{tabularx}{5.8in}{|X||c|c|c|c|c|c|}
\hline
 &\multicolumn{3}{c|}{Train Error}&\multicolumn{3}{c|}{Test Error}\\
\cline{2-7}
Dataset &NW+BP&TRUST-&Gain&NW+BP&TRUST-&Gain\\
&&TECH&&&TECH&\\
\hline\hline

Cancer & 2.25  &  1.57 &42.99& 3.65 &   3.06 &   19.06 \\
\hline  Image  & 7.48  &  5.17  &  44.82 &  9.39  &  7.40   & 26.90\\
\hline Ionosphere & 1.56 &0.92 &   69.57 &  8.67   & 6.54  &  32.57\\
\hline Iris &1.33   & 0.67   & 100.00 & 3.33 &   2.67  &  25.00  \\
\hline Diabetes & 21.41 &19.55 &9.53 &23.70 &21.09& 12.37 \\ \hline
Sonar& 2.35 &0.42 &456.96 & 17.26 &  14.38  & 20.03 \\ \hline  Wine
& 7.60 &1.62 &370.06 & 14.54 &4.48 &224.82 \\ \hline
\end{tabularx}
}
\end{center}
\label{TB:results1}\end{table*}

\subsection{Visualization}
\begin{figure*}[htp]
   \centering
   \subfigure[Wine Dataset]{\includegraphics[width = 2.55 in]{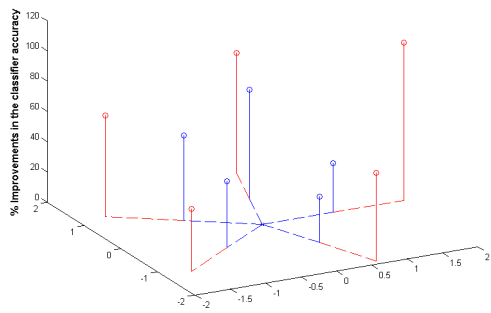}}\qquad
   \subfigure[Diabetes Dataset]{\includegraphics[width = 2.55 in]{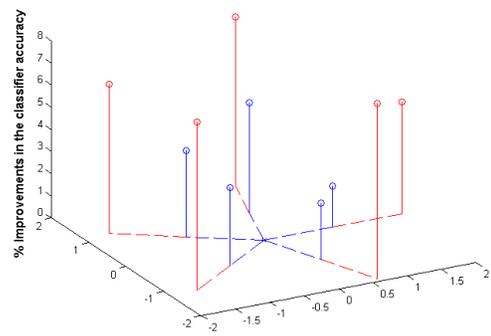}}\qquad
   \subfigure[Cancer Dataset]{\includegraphics[width = 2.55 in]{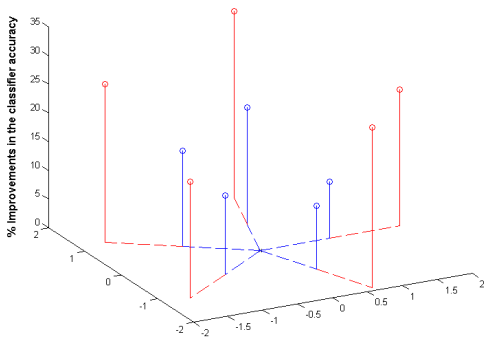}}\qquad
   \subfigure[Image Dataset]{\includegraphics[width = 2.55 in]{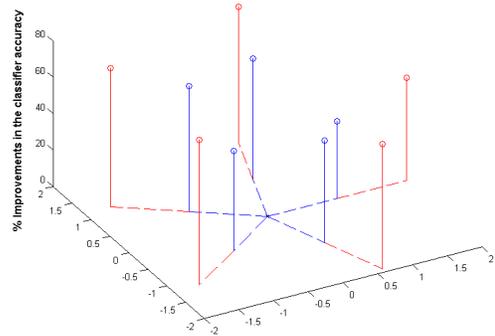}}\qquad

   \caption{\label{fig:spider} Spider web diagrams showing the tier-1 and tier-2 improvements using TRUST-TECH
   method on various benchmark datasets. The basis two axes are chosen arbitrarily and the
   vertical axis represents the improvements in the classifier
   accuracy. The distances between each tier are normalized to
   unity.
   }
 \end{figure*}

The improvements of the TRUST-TECH method are demonstrated using
spider web diagrams. Spiderweb diagram (shown in Fig.~\ref{fig:spider}) is a pretentious way to demonstrate the accuracy improvements
in a tier-by-tier manner. The circle in the middle of the plot
represents the starting local optimal solution. The basic two
dimensions are chosen arbitrarily for effective visualization and
the vertical axis is the percentage improvement in the
classification accuracy. Unit distances are used between the tiers
and the improvements are averaged out for 10 folds. The five
vertical lines surrounding the center circle are the best five local
minima obtained from a tier-1 search across all folds. The tier-2 improvements are also plotted. It should be noted that
the best tier-1 solution need not give the best second tier
solution.

Most successful algorithms for training
artificial neural networks make use of some stochastic approaches in
combination with backpropagation to obtain an effective set of
training parameters. Due to the limited fine-tuning capability of
these algorithms, even the best solutions that they can provide are
locally optimal. In this chapter, a new TRUST-TECH based method for improving the local
search capability of these training algorithms is proposed. This
method improves the neural network model thus allowing improved
classification accuracies by providing a better set of training
parameters. Because of the non-probabilistic in nature, multiple runs of our method
from a given initial guess will provide exactly the same results.
Different global and local methods work effectively on different
datasets. The proposed TRUST-TECH based training algorithm allows the user to have the
flexibility of choosing different global and local techniques for
training.

\chapter{Evolutionary TRUST-TECH}
\label{ch:evoltrust}

This chapter discusses the advantages of using the proposed TRUST-TECH based schemes in combination with the widely used evolutionary algorithms. The main focus of this work is to demonstrate the use of having a deterministic search compared to a stochastic one for exploring the neighborhood during the global stage. Most of the popular stochastic global optimization methods use some probabilistic neighborhood search algorithms for exploring the promising subspaces. Due to this probabilistic nature, one might obtain solutions that were already found or sometimes, even miss some promising solutions. Adding a deterministic search schemes like TRUST-TECH will help this neighborhood search to find promising solutions more effectively. First, we provide an overview of evolutionary computation, and describe the evolutionary algorithms in detail.

\section{Overview}
\label{sec:overview}

Research efforts on developing computational models have been rapidly growing in recent times. Amongst the several ways of developing effective computational models, evolutionary computation have become very popular. The evolutionary computational models \cite{Fraser57} use the well-studied computational models of evolutionary processes as key elements. There are a variety of evolutionary computational models that have been proposed and studied which we will refer to as evolutionary algorithms. In simple terms, they simulate the process of evolution of the individual components via processes of selection and reproduction. These processes depend on the fitness of the individuals as defined by an environment. Evolutionary algorithms maintain a population of individuals that evolve according to rules of selection and other genetic operators, such as recombination and mutation. Each individual in the population receives a measure of its fitness in the environment. Of all these operations mentioned above, selection is the main one that can exploit the available fitness information and mainly considers those individuals with high fitness value. Recombination and mutation perturb those individuals, providing general search strategies and heuristics for exploring the solution space. These algorithms are sufficiently complex to provide robust and powerful adaptive search mechanisms especially when the number of optimal solutions grow exponentially with the problem complexity.
\begin{algorithm}
\caption{Evolutionary Algorithm} \label{ea}
\begin{algorithmic}
\STATE \textbf{Input:} Initial Population, no. of iterations  \STATE \textbf{Output:}
promising solutions  \STATE \textbf{Algorithm:}

\STATE $t=0$
\STATE Initialize population P(t)
\STATE Evaluate P(t)

\WHILE{not done}
\STATE $t=t+1$
\STATE parent selection P(t)
\STATE recombine P(t)
\STATE mutate P(t)
\STATE evaluate P(t)
\STATE survive P(t)

\ENDWHILE

\end{algorithmic}
\end{algorithm}

Algorithm~\ref{ea} outlines a simple evolutionary algorithm (EA). A population of individual structures is initialized and then evolved from generation to generation by repeated applications of evaluation, selection, recombination, and mutation. The initial and final population size N is generally constant in an evolutionary algorithm. In other words, the initial fixed number of solutions are chosen and they evolve into more and more promising solutions as the time progresses. In our case, the time will be indicated in terms of the number of iterations taken to evolve.

An evolutionary algorithm typically initializes its population randomly, although some apriori information and domain specific knowledge can also be used to obtain promising starting points. If promising starting values are chosen then good solutions can be obtained in a fewer iterations compared to the number of iterations taken for an algorithm that was started with random points. Evaluation measures the fitness of each individual according to its worth in the environment. The complexity of the evaluation process is highly problem dependent. It may be as simple as computing a fitness function or as complex as running an elaborate simulation. Selection is usually performed in two steps, parent selection and survival. Parent selection decides who becomes parents and how many children the parents can have. Children are created via recombination, which exchanges information between parents. This recombination mechanism is the vital component of an evolutionary algorithm because it will provide new individuals in the environment. Mutation is an important step in the algorithm which perturbs the children to obtain minor changes in the solution. The children are then evaluated using the fitness function and the survive step decides who survives in the population.

\section{Variants of Evolutionary Algorithms}
\label{sec:ea}

The algorithm described above is the basic backbone of a simple evolutionary algorithm. Several variants and improvements for this basic architecture have been proposed in the literature and is currently an active topic of research. We will now discuss the three most popular and well-studied variations of evolutionary algorithms. The three methodologies are : 1. Evolutionary programming \cite{Fogel66}, 2. Evolution strategies \cite{Beyer02} and 3. Genetic algorithms \cite{Holland75,Goldberg89}. At a higher level, all these methods implement an evolutionary algorithm but the details of their implementation are completely different. They differ in the choice of problem representation, types of selection mechanism, forms of genetic operators, and performance measures.

Evolutionary programming (EP), developed by Fogel et al. \cite{Fogel66} traditionally used problem representations that are specific to the application domain. For example, in real-valued optimization problems, the individuals within the population are real-valued vectors. Similarly, ordered lists are used for traveling salesman problems, and graphs for applications with finite state machines. The basic scheme is similar to an evolutionary algorithm except that the recombination is generally not performed since the forms of mutation used are adaptive. These mutations are quite flexible that can produce perturbations that are similar to recombination, if desired. One of the heavily studied aspects of this approach is the extent to which an evolutionary algorithm is affected by its choice of the perturbation rates used to produce variability and the novelty in evolving populations.

Genetic algorithms (GAs), developed by Holland \cite{Holland75, Goldberg89} are arguably the most well known form of evolutionary algorithms. They have been traditionally used in a more domain independent setting, namely, bit-strings. Those individuals with higher relative fitness are more likely to be selected as parents. N children are created via recombination from the N parents. The N children are then mutated and the best survivors will replace the N parents in the population. It should be noted that there is a  strong emphasis on mutation and crossover.

The third category is an evolution strategy (ES). It follows the basic EA architecture and the number of children created is usually greater than N. After initialization and evaluation, individuals are selected uniformly random. Survival is deterministic and allows the N best children to survive. Like EP, considerable effort is made on adapting mutation as the algorithm runs by allowing each variable within an individual to have an adaptive mutation rate that is normally distributed with a zero expectation. Unlike EP, however, recombination does play an important role in evolution strategies, especially in adapting mutation.

These three approaches (EP, ESs, and GAs) have inspired an increasing amount of research and development of new forms of evolutionary algorithms for use in specific problem solving contexts. In this chapter, we focus on the aspect of mutations and improve their performance. The mutations usually allow us to obtain individuals with minor changes. In terms of the parameter space, mutations will merely perturb the solutions to obtain new solutions that might potentially be more promising than the original solutions. Most of the stochastic optimization methods which try to obtain global optimal solutions perform some kind of neighborhood search. In the popularly used simulated annealing technique \cite{Kirkpatrick83}, the neighborhood search is performed by local moves.

However, performing the neighborhood search in a stochastic manner will have many problems like:

\begin{itemize}
\item{One might not know the extent to which the mutation has to be performed on an individual.}
\item{It is difficult to understand the locations of the phenotype where the mutation should occur.}
\end{itemize}

For both the problems mentioned above, one can realize that performing a more systematic neighborhood search for obtaining better solutions will help in performing this mutation step in a better manner. In this chapter, we replace this concept of mutation by other local search techniques. Our first model will use a local refinement strategy instead of mutations. Our second model will use the TRUST-TECH based neighborhood search for searching the nearby local optimal solutions. Of course, it might incur some additional computational cost to perform these sophisticated neighborhood search strategies, but they will almost surely have the advantage of performing a nearly perfect local search which can have the potential of reducing the total number of iterative steps taken for convergence. In other words, we alter the mutation aspect of the algorithm so that it can result in the reduction of the total number of generations (computational time) required by the complete algorithm.

\begin{figure}
   \centering
   \subfigure[Evolutionary Local Refinement]{\includegraphics[width = 3.0 in]{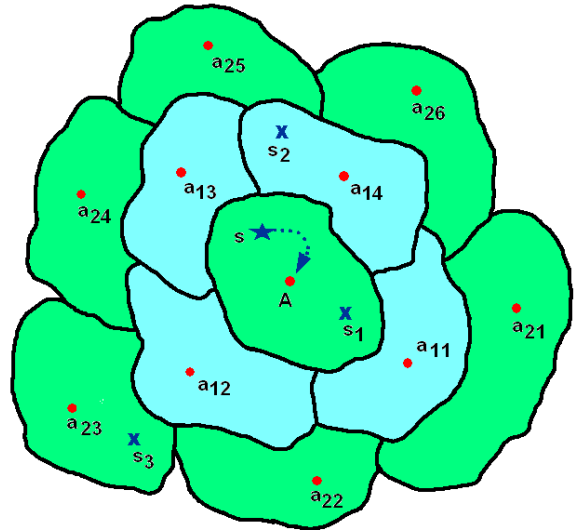}}\qquad
   \subfigure[Evolutionary TRUST-TECH]{\includegraphics[width = 3.0 in]{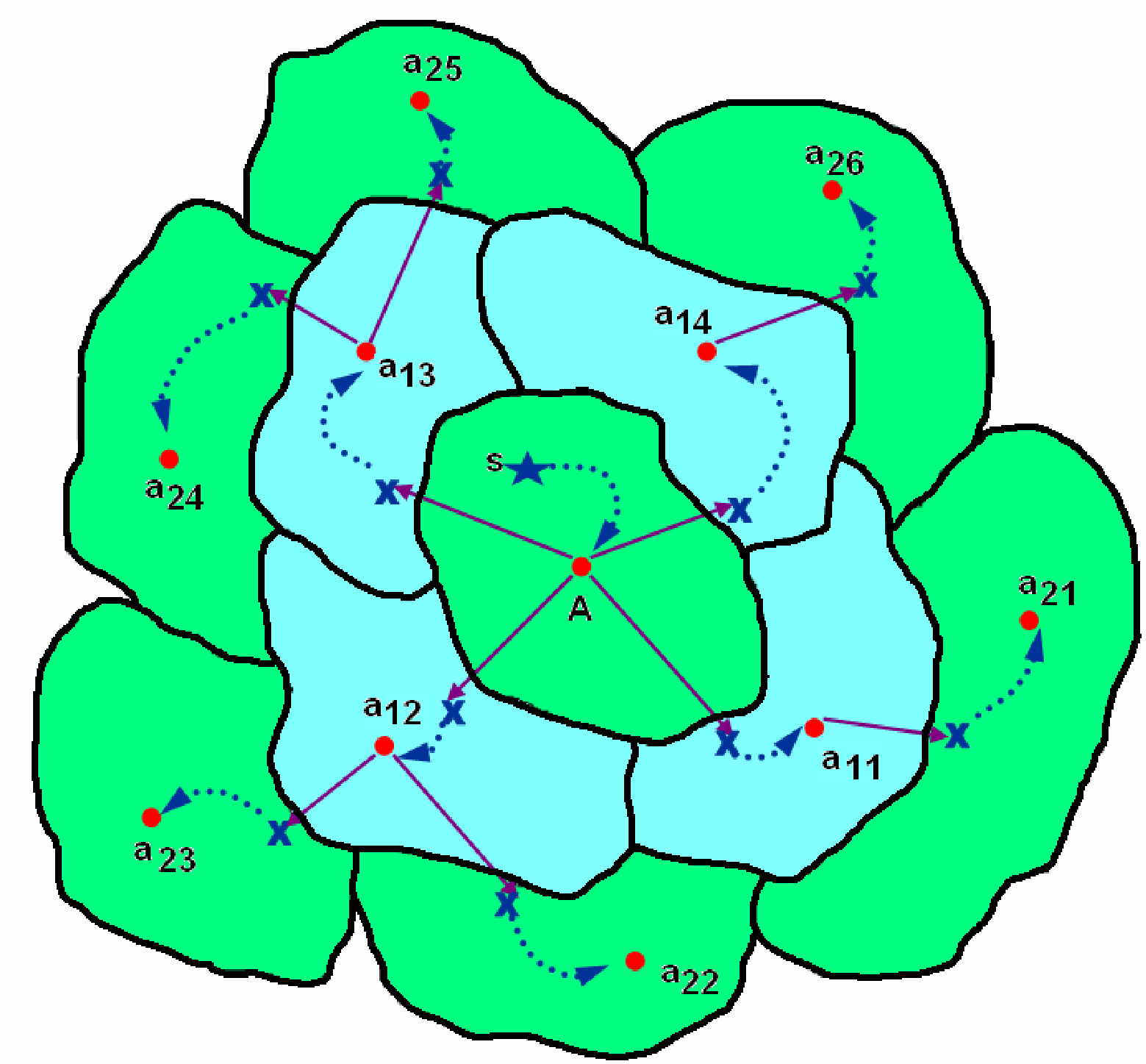}}\qquad
   \caption{\label{fig:algo} Topology of the nonlinear surface discussed in the introduction chapter. The initial point obtained after the recombination is `s'. The original concept of mutations will randomly perturb the point in the parameter space and obtain different points ($s_1,s_2$ and $s_3$). Two different evolutionary models (a) Local refinement strategy where the local optimization method is applied with `s' as initial guess to obtain `A'. (b) Evolutionary TRUST-TECH methodology where the surrounding solutions are explored systematically after the recombination.
   }
   \label{fig:evolloctrust}
\end{figure}

\section{Evolutionary TRUST-TECH Algorithm}
\label{sec:evoltrust}

\begin{algorithm}
\caption{Evolutionary Algorithm with Local Refinement} \label{els}
\begin{algorithmic}
\STATE \textbf{Input:} Initial Population, no. of iterations  \STATE \textbf{Output:}
promising solutions  \STATE \textbf{Algorithm:}
\STATE $t=0$
\STATE Initialize population P(t)
\STATE Evaluate P(t)
\WHILE{not done}
\STATE $t=t+1$
\STATE parent selection P(t)
\STATE recombine P(t)
\STATE Local Refinement P(t)
\STATE evaluate P(t)
\STATE survive P(t)
\ENDWHILE

\end{algorithmic}
\end{algorithm}

First, we will describe the evolutionary algorithm with local refinement strategy. In this model, the mutation is replaced with a local refinement strategy. This can be either in discrete space or a continuous space. The important point here is to ensure that a solution which is better than the existing solution is always obtained. The local refinement strategy will obtain the corresponding local optimal solution. In the case of mutations, there is no guarantee that the new point has a higher fitness function value than the original point. To ensure a better fitness value, one can apply this local refinement which is a greedy procedure to obtain a better solution. All other aspects of the evolutionary algorithm remain unchanged. Algorithm~\ref{els} describes the evolutionary algorithm with local refinement strategy.

\begin{algorithm}
\caption{Evolutionary TRUST-TECH} \label{evtrust}
\begin{algorithmic}
\STATE \textbf{Input:} Initial Population, no. of iterations  \STATE \textbf{Output:}
promising solutions  \STATE \textbf{Algorithm:}

\STATE $t=0$
\STATE Initialize population P(t)
\STATE Evaluate P(t)

\WHILE{not done}
\STATE $t=t+1$
\STATE parent selection P(t)
\STATE recombine P(t)
\STATE TRUST-TECH P(t)
\STATE evaluate P(t)
\STATE survive P(t)
\ENDWHILE
\end{algorithmic}
\end{algorithm}
Fig~\ref{fig:evolloctrust} clearly demonstrates both the proposed models. Let `s' denote the point obtained after the recombination process. Mutation will randomly perturb this point to obtain another point. It can be either $s_1,s_2$ or $s_3$. In all these cases, there is no guarantee that the new point has a higher fitness function value. However, applying a local refinement strategy, `s' will converge to `A' which has either equal or higher fitness value. The mutations are completely replaced using this local refinement method. There might be other promising solutions with higher fitness function values. Hence, in the evolutionary TRUST-TECH mechanism, this local refinement strategy is replaced by the TRUST-TECH methodology. The TRUST-TECH used in this step is identical to the algorithm presented in the previous chapter. The only difference being that the starting point is obtained from the recombination operator. Using the TRUST-TECH strategy, neighborhood solutions are obtained and the best one is retained for the next iteration in the evolutionary process.

\section{Experimental Results}

For our implementation, we started with 10 initial points and refined them through the evolutionary process. During each selection stage, two parents are chosen randomly and recombined. Amongst the four individuals (2 parents and 2 children), only two individuals with higher fitness function value are retained and the other two are discarded. The population is refined for 100 generations.  Each time only 10 recombinations are performed. For the local refinement strategy, we used the Levenberg-Marquardt method. For the TRUST-TECH implementation, only the first-tier solutions were obtained for computational efficiency. Amongst all the tier one solutions, the best solution is chosen and all the rest are discarded.

\begin{table*}[htp]
\centering \caption{\protect Results of Evolutionary TRUST-TECH model }
\begin{center}
\begin{tabular}{|c|c|c|c|}
\hline
Dataset & EA & EA+LR &EA+TRUST-TECH\\
\hline Pima & 0.1642 & 0.1432&0.1367\\
\hline Cancer & 0.0193 &0.01654&0.01423\\
\hline Wine &0.0775&0.0651&0.0472\\
\hline
\end{tabular}\end{center}
\label{TB:resultsevol}\end{table*}
Table~\ref{TB:resultsevol} reports the results of our approach. For the Local refinement and TRUST-TECH strategies, we used only 50 generations. The MSE over the training has been reported. One of the main observations from our experiments is that it is not necessary that every iteration during the local refinement will yield a better score than the original model. For example, if the local refinement strategy yields a better score at a particular generation, then its improvement during the next few generations might not be significant. This applies to even TRUST-TECH method as well. However, as it is evident from the results, there can be significant improvements in terms of MSE and this can be achieved at fewer number of iterations of the evolutionary algorithm. 
\section{Parallel Evolutionary TRUST-TECH}
From the results shown in the previous section, one can see that TRUST-TECH can help to provide faster convergence of the traditional evolutionary algorithms. The proposed framework can be easily extended to work on parallel machines. This will work in a similar manner to any other parallel evolutionary algorithm. The local refinement and neighborhood strategies proposed here can work independent for each of the selection and recombination individuals chosen. Two individuals are selected for recombination and then the local refinement can take place on a different local machine. This way, all the local refinements can be performed in different machines and the final results can be evaluated and the next generation parents are chosen in a centralized machine. This centralized machine (or the server) can obtain the results from all the local machines which perform the neighborhood search.

\chapter{Conclusion and Future Work}
\label{ch:conclusion}
This chapter concludes our discussion and highlights the most important contributions of this thesis. It also discusses the future research directions that one might want to pursue using the models presented in this thesis.

\section{Conclusion}
\label{sec:conclusion}

In this thesis work, we develop TRUST-TECH based methods for various problems related to areas of heuristic search, optimization and learning. We demonstrate the applicability and effectiveness of these methods for practical and high-dimensional nonlinear optimization problems. One of the main ideas of this framework is to transform the original optimization problem into a dynamical system with certain properties and obtain more useful information about the nonlinear surface via the dynamical and topological properties of the dynamical system.

In Chapter~\ref{ch:saddle}, we apply a stability boundary following procedure for obtaining saddle points on various potential energy surfaces that arise in the field of computational biology and computational chemistry. To find a saddle point, following the stability boundary is computationally more efficient than directly searching for a saddle point from a given local minimum. This algorithm works deterministically and requires very few user-specific parameters. The procedure has been successfully tested on a 525-dimensional problem. For energy surfaces that show symmetric behaviour, we propose a simplified version of this procedure.

Nonlinear optimization problems arise in several domains in science and engineering. Several algorithms had been proposed in the
optimization literature for solving these problems efficiently. Typically, optimization methods can be classified into two categories: (1) Global methods and (2) Local methods. Global methods are powerful stochastic methods that search the entire parameter space and obtain promising regions. Local methods, on the other hand, are deterministic methods and usually converge to a locally optimal solution that is nearest to a given initial point. There is a clear gap between these two methods and there is a need for a method that can search in the neighborhood regions.

The proposed TRUST-TECH based methods systematically search for neighborhood local optimal solutions by exploring the dynamic and geometric characteristics of stability boundaries of a nonlinear dynamical system corresponding to the nonlinear function of interest. This framework consists of three stages namely: (i) Global stage, (ii) Local stage and (iii) Neighborhood-search stage. These methods have been successfully used for various machine learning problems and are demonstrated in the context of both supervised (training artificial neural networks in Chapter~\ref{ch:training}) and unsupervised (expectation maximization in Chapter~\ref{ch:trust-tech-em}) machine learning problems. In both these scenarios, obtaining a global optimal solution in the parameter space corresponds to exploiting the complete potential of the given model. More complicated models might achieve the same function value but, they tend to loose the generalization capability. Our methods are tested on both synthetic and real datasets and the advantages of using this TRUST-TECH based framework are clearly manifested. The improvements in the performance of the expectation maximization algorithm are demonstrated in the context of mixture modeling and general likelihood problems such as the motif finding problem.

This framework not only reduces the sensitivity to initialization, but also allows the flexibility for the practitioners to use various global and local methods that work well for a particular problem of interest. Thus, it can act as a flexible interface between the local method (EM) and other global methods. This interface plays a vital role in problems where the functions optimized by the global method and the local method are not the same. For example, in the motif finding problem, a global method is used in the discrete space and a local method is used in the continuous space. The points obtained as a result of the global method need not be in the convergence region (of the local method) of the most promising solutions. In such cases, applying tier-by-tier search can significantly improve the quality of the solutions as demonstrated by the results of finding optimal motifs in Chapter~\ref{ch:motif}. Also, this framework has the potential to work with any global and local method by treating them as a black-box (without knowing their detailed algorithms).

In Chapter~\ref{ch:smooth}, we propose a novel smoothing framework in the context of Gaussian mixture models. The proposed component-wise kernel smoothing approach can reduce the number of local maxima on the log-likelihood surface and can potentially obtain promising initial points for the EM algorithm. Performing component-wise smoothing will maintain the structure of the log-likelihood function and hence the EM can be applied directly with a few modifications.

Frameworks for combining TRUST-TECH with other hierarchical stochastic algorithms such as smoothing algorithms and evolutionary algorithms are also proposed and tested on various datasets. In Chapter~\ref{ch:evoltrust}, it has been shown that the neighborhood-search stage can be incorporated into the global stage in order to improve the performance of the global stage in terms of the quality of the solutions and the speed of convergence.


\newpage
\section{Future Work}
\label{sec:future}
The algorithm proposed for finding saddle points can be applied to the problem of finding pseudo-native like structures and their corresponding transition states during the protein folding process. The TRUST-TECH based Expectation-Maximization algorithm can be extended to other widely used EM related problems for the family of probabilistic graphical models such as k-means clustering, training Hidden Markov Models, Mixture of Factor Analyzers, Probabilistic Principal Component Analysis, Bayesian Networks etc. Extension of these techniques to Markov Chain Monte Carlo methods (like Gibbs sampling) is also feasible. Several real-world applications such as image segmentation, gene finding, speech processing and text classification can benefit significantly from these methods. Extensions to constrained optimization problems appears to be a promising direction as well. Different global methods and local solvers can be used along with TRUST-TECH framework to study the flexibility of this framework. For machine learning problems, automatically choosing a model is an important and difficult problem to tackle. TRUST-TECH based methods are generic enough to incorporate any model selection criterion (such as Akaike information criterion (AIC) and Bayesian information criterion (BIC)) into the objective function. Basically, this term is added in the objective function in order to penalize complex models.

As a continuation of the smoothing work, the effects of convolving Gaussian components with other kernels must be investigated. Efficient algorithms for choosing the smoothing parameter automatically based on the available data can be developed. Though applied for Gaussian mixture models in this paper, convolution based smoothing strategies can be treated as powerful optimization tools that can enhance the search capability significantly. The novel neural network training algorithm can be extended to the problem of simultaneously deciding the architecture and the training parameters. This problem is characterized by a combination of a set of discrete (for architecture) and continuous (for training) variables. Its performance on large scale applications like character recognition, load forecasting etc. must be tested. The power of evolutionary TRUST-TECH has been demonstrated only in the case of training neural networks problem. The algorithm is not specific to neural network and can be demonstrated for general nonlinear programming problems in the future.

In the context of optimization, TRUST-TECH can also help the local solvers to converge to desired optimal solution. Especially, when a local method is applied to a point near the stability boundary, it might diverge or might converge to a different local optimal solution at a very slow rate. TRUST-TECH methodology can be used to integrate the system and obtain new points that are much closer to the desired local optimal solution. Using this new point as initial condition, there are higher chances that the local method will converge to the desired local optimal solution. The number of integration steps required and the step size for integration are research topics that can be pursued in the future.
 
Most of the problems that were discussed in this thesis primarily focus on finding critical points. Usually, the trajectory of convergence is not of much importance in optimization and machine learning problems. Transformation of the nonlinear objective function to it corresponding gradient system is proposed in this thesis. One can extend this work by scaling the dynamical system so that it preserves the location of all the critical points. Scaling can modify the trajectories significantly and can potentially improve the speed of convergence. The characterization of the scaling factor required for various systems is a potential research topic.

Though proposed for a few global methods like smoothing and evolutionary algorithms, TRUST-TECH can be used to improve the performance of any hierarchical stochastic global optimization method. For example, multi-level optimization methods are popular tools that are powerful in solving problems in various domains of engineering. The basic idea is to reduce the dimensionality of the problem and obtain a solution and carefully trace back this solution to the original problem. During this trace back procedure, local methods might be inefficient in tracing the solution accurately and the use of TRUST-TECH based tier-by-tier search can help in exploring the neighborhood and obtaining the global optimal solution accurately.

\bibliographystyle{plain}
\bibliography{Thesis,EMThesis,SaddleThesis,NeuralThesis,smoothThesis,MotifThesis,EvolThesis}

\end{document}